\documentclass[preprint,5p,twocolumn]{elsarticle}

\journal{NeuroImage}
\usepackage[table, usenames, dvipsnames]{xcolor}
\definecolor{mybetter}{RGB}{159, 255, 180}
\usepackage{lineno,hyperref}
\usepackage{bm}
\usepackage{amsmath,graphicx}
\usepackage{caption}
\usepackage{multirow}
\usepackage{amssymb}
\usepackage{tikz} 
\usepackage{array}
\usepackage[font=footnotesize,labelfont=bf]{caption}
\usepackage{ulem}
\usepackage{adjustbox}
\usepackage{afterpage}
\usepackage{hhline}
\usepackage[ruled, linesnumbered]{algorithm2e}
\usetikzlibrary{calc,arrows,positioning,shapes,shadows,spy,snakes,plotmarks,matrix,fit,backgrounds}
\modulolinenumbers[5]

\usepackage[draft]{fixme}
\usepackage{float}

\newcommand{\chg}[1]{{\color{black}{#1}}}
\newcommand{\xhsp}[1]{{\color{black}{#1}}}

\begin{document}

\begin{frontmatter}

\title{Brain Extraction from Normal and Pathological Images: \\ A Joint PCA/Image-Reconstruction Approach}

\author[unc]{Xu Han\corref{mycorrespondingauthor}}
\cortext[mycorrespondingauthor]{Corresponding author}
\ead{xhs400@cs.unc.edu}

\author[uni-salzburg]{Roland Kwitt}
\author[kitware]{Stephen Aylward}
\author[upenn]{Spyridon Bakas}
\author[tum]{Bjoern Menze}
\author[usc]{Alexander Asturias}
\author[ucla]{Paul Vespa}
\author[usc]{John Van Horn}
\author[unc]{Marc Niethammer}

\address[unc]{Department of Computer Science, University of North Carolina at Chapel Hill, USA}
\address[uni-salzburg]{Department of Computer Science, University of Salzburg, Austria}
\address[kitware]{Kitware Inc., USA}
\address[upenn]{Center for Biomedical Image Computing and Analytics, Perelman School of Medicine, University of Pennsylvania, USA}
\address[tum]{Department of Computer Science, Technical University of Munich, Germany}
\address[usc]{Institute of Neuroimaging and Informatics, University of Southern California, USA}
\address[ucla]{David Geffen School of Medicine, UCLA Medical Center, USA}

\begin{abstract}
Brain extraction from 3D medical images is a common pre-processing step. A variety of approaches exist, but they are frequently only designed to perform brain extraction from images \textit{without} strong pathologies. Extracting the brain from images exhibiting strong pathologies, for example, the presence of a brain tumor or of a traumatic brain injury (TBI), is challenging. In such cases, tissue appearance may substantially deviate from normal tissue appearance and hence violates algorithmic assumptions for standard approaches to brain extraction; consequently, the brain may not be correctly extracted. 

This paper proposes a brain extraction approach which can explicitly account for pathologies by jointly modeling normal tissue appearance and pathologies. Specifically, our model uses a three-part image decomposition: (1) normal tissue appearance is captured by principal component analysis (PCA), (2) pathologies are captured via a total variation term, and (3) the skull and surrounding tissue is captured by a sparsity term. Due to its convexity, the resulting decomposition model allows for efficient optimization. Decomposition and image registration steps are alternated to allow statistical modeling of normal tissue appearance in a fixed atlas coordinate system. As a beneficial side effect, the decomposition model allows for the identification of potentially pathological areas and the reconstruction of a quasi-normal image in atlas space. 

We demonstrate the effectiveness of our approach on four datasets: the publicly available IBSR and LPBA40 datasets which show normal image appearance, the BRATS dataset containing images with brain tumors, and a dataset containing clinical TBI images. We compare the performance with other popular brain extraction models: ROBEX, BEaST, MASS, BET, BSE and a recently proposed deep learning approach. Our model performs better than these competing approaches on all four datasets. \chg{Specifically, our model achieves the best median (97.11) and mean (96.88) Dice scores over all datasets. The two best performing competitors, ROBEX and MASS, achieve scores of 96.23/95.62 and 96.67/94.25 respectively}. Hence, our approach is an effective method for high quality brain extraction for a wide variety of images.
\end{abstract}

\begin{keyword}
Brain Extraction, Image Registration, PCA, Total-Variation, Pathology
\end{keyword}

\end{frontmatter}

\section{Introduction}
\label{section:introduction}

Brain extraction\footnote{We avoid the commonly used term skull stripping. We are typically interested in removing more than the skull from an image and are instead interested only in retaining the parts of an image corresponding to the brain.} from volumetric magnetic resonance (MR) or computed tomography images~\cite{muschelli2015validated} is a common pre-processing step in neuroimaging as it allows to spatially focus further analyses on the areas of interest. The most straightforward approach to brain extraction is by manual expert delineation. Unfortunately, such expert segmentations are time consuming and very labor intensive and therefore not suitable for large-scale imaging studies. Moreover, brain extraction is complicated by differences in image acquisitions and the presence of tumors and other pathologies that add to inter-expert segmentation variations.

\vskip1ex
Many methods have been proposed to replace manual delineation by automatic brain extraction. In this paper, we focus on and compare with the following six widely-used or recently published brain extraction methods, which cover a wide range of existing approaches:

\begin{itemize}
\item {\it Brain Extraction Tool (BET):} BET~\cite{smith2002fast} is part of FMRIB Software Library (FSL) ~\cite{jenkinson2012fsl, FSL} and is a widely used method for brain extraction. BET first finds a rough threshold based on the image intensity histogram, which is then used to estimate the center-of-gravity (COG) of the brain. Subsequently, BET extracts the brain boundary via a surface evolution approach, starting from a sphere centered at the estimated COG.

\item {\it Brain Surface Extraction (BSE):} BSE~\cite{shattuck2001magnetic} is part of BrainSuite~\cite{shattuck2002brainsuite, BrainSuite}. BSE uses a sequence of low-level operations to isolate and classify brain tissue within T1-weighted MR images. Specifically, BSE uses a combination of diffusion filtering, edge detection and morphological operations to segment the brain. \chg{BrainSuite provides a user interface which allows for human interaction. Hence better performance may be obtained by interactive use of BSE. However, our objective was to test algorithm behavior for a fixed setting across a number of different datasets.}

\item {\it Robust Learning-based Brain Extraction System (ROBEX):} ROBEX~\cite{iglesias2011robust,ROBEX} is another widely used method which uses a random forest classifier as the discriminative model to detect the boundary between the brain and surrounding tissue. It then uses an active shape model to obtain a plausible result. \chg{While a modification of ROBEX for images with brain tumors has been proposed~\cite{speier2011robust}, its implementation is not available. Hence we use the standard ROBEX implementation for all our tests.}

\item {\it Deep Brain Extraction:} We additionally compare against a recently proposed deep learning approach for brain extraction~\cite{kleesiek2016deep,DeepBrainExtraction} which uses a 3D convolutional neural network (CNN) trained on normal images and images with mild pathologies. Specifically, it is trained on the IBSR v2.0\footnote{This is a different dataset than the IBSR dataset that we use in this paper.}~\cite{IBSRv2}, LPBA40~\cite{shattuck2008construction,LPBA40} and OASIS~\cite{marcus2007open,OASIS} datasets. We use this model as is without additional fine-tuning for other datasets.

\item {\it Brain Extraction Based on non-local Segmentation Technique (BEaST):} BEaST~\cite{eskildsen2012beast, BEAST} is another recently proposed method, which is inspired by patch-based segmentation. In particular, it identifies brain patches by assessing candidate patches based on their sum-of-squared-difference (SSD) distance to known brain patches. \xhsp{BEaST allows using different image libraries to guide the brain extraction.}

\item {\it Multi-Atlas Skull Stripping (MASS):} MASS~\cite{doshi2013multi}, uses multi-atlas registration and label fusion for brain extraction. It has shown excellent performance on normal (IBSR, LPBA40) and close to normal (OASIS) image datasets. One of its main disadvantages is its runtime. \chg{An advantage of MASS, responsible for its performance and robustness, is that one can easily make use of dataset-specific brain templates. However, this requires obtaining such brain masks via costly manual segmentation. For a fair comparison to all other methods, and to test the performance of a given algorithm across a wide variety of datasets, we select 15 anonymized templates for MASS's multi-atlas registration. These templates were obtained from various studies and are provided along with the MASS software package~\cite{MASS}, as well as through CBICA's Image Processing Portal}~\cite{CBICAIPP2015}.
\end{itemize}

\vskip1ex
In addition to these methods, many other approaches have been proposed. For example, Segonne et al.~\cite{segonne2004hybrid} proposed a hybrid approach which combines watershed segmentation with a deformable surface model. Watershed segmentation is used to obtain an initial estimate of the brain region which is then refined via a surface evolution process. 3dSkullStrip is part of the AFNI (Analysis of Functional Neuro Images) package~\cite{cox1996afni, AFNI}. It is a modified version of BET. In contrast to BET, it uses image data inside and outside the brain during the surface evolution to avoid segmenting the eyes and the ventricles. 

\vskip1ex
Even though all these brain extraction methods exist and are regularly used, a number of challenges for automatic brain extraction remain:
\begin{itemize}
\item Many methods show varying performances on different datasets due to differences in image acquisition (e.g., slightly different sequences or differing voxel sizes). Hence, a method which can reliably extract the brain from images acquired with a variety of different imaging protocols would be desirable.
\item Most methods only work for images which appear normal or show very minor pathologies. Strong pathologies, however, may induce strong brain deformations or strong localized changes in image appearance, which can impact brain extraction. For example, for methods based on registration, the accuracy of brain extraction will depend on the accuracy of the registration, which can be severely affected in the presence of pathologies. Hence, a brain extraction method which works reliably even in the presence of pathologies (such as brain tumors or traumatic brain injuries) would be desirable. 
\end{itemize}

Inspired by the low-rank + sparse (LRS) image registration framework proposed by Liu et al.~\cite{liu2014low} and our prior work on image registration in the presence of pathologies~\cite{han2017efficient}, we propose a brain extraction approach which can tolerate image pathologies (by explicitly modeling them) while retaining excellent brain extraction performance in the absence of pathologies.

The contributions of our work are as follows:
\begin{itemize}
\item {\it (Robust) brain extraction:} Our method can reliably extract the brain from a wide variety of images. We achieve state-of-the-art results on images with normal appearance, slight, and strong pathologies. Hence our method is a generic brain extraction approach.
\item {\it Pathology identification:} Our method captures pathologies via a total variation term in the decomposition model. 
\item {\it Quasi-normal estimation:} Our model allows the reconstruction of a quasi-normal image, which has the appearance of a corresponding pathology-free or pathology-reduced image. This quasi-normal image also allows for accurate registrations to, e.g., a normal atlas. 
\item {\it Extensive validation:} We extensively validate our approach on four different datasets, two of which exhibit strong pathologies. We demonstrate that our method achieves state-of-the-art results on all these datasets using a \textit{single} fixed parameter setting.
  \item {\it Open source:} Our approach is available as open-source software.
\end{itemize}

The remainder of the paper is organized as follows. Section \ref{section:mothodology} introduces the datasets that we use and discusses our proposed model, including the pre-processing, the decomposition and registration, and the post-processing procedures. Section \ref{section: experiments} presents experimental results on 3D MRI datasets demonstrating that our method consistently performs better than BET, BSE, ROBEX, BEaST, MASS and the deep learning approach for all four datasets. Section \ref{section:discussion} concludes the paper with a discussion and an outlook on possible future work.

\section{Materials and Methods}
\label{section:mothodology}
\subsection{Datasets}

We use the ICBM 152 non-linear atlas (2009a)~\cite{fonov2009unbiased} as our normal control atlas. ICBM 152 is a 1x1x1 mm template with 197$\times$233$\times$189 voxels, obtained from T1-weighted MRIs. Importantly, it also includes the brain mask. As the ICBM 152 atlas image itself contains the skull, we can obtain a \textit{brain-only} atlas simply by applying the provided brain mask. 

\vskip1ex
We use five different datasets for our experiments. Specifically, we use one (OASIS, see below) 
of the datasets to build our PCA model and the remaining four to test our brain extraction approach.

\vskip1ex
\noindent {\bf OASIS.} We use images from the Open Access Series of Imaging Studies (OASIS)~\cite{marcus2007open,OASIS} to build the PCA model for our brain extraction approach. The OASIS cross-sectional MRI dataset consists of 416 sagittal T1-weighted MRI scans from subjects between 18 and 96 years of age. In this data corpus, 100 of the subjects over 60 years old have been diagnosed with very mild to mild Alzheimer's disease (AD). The original scans were obtained with in-plane resolution $1\times 1$ mm ($256\times 256$), slice thickness = 1.25 mm and slice number = 128. For each subject, a gain-field corrected atlas-registered image and its corresponding masked image in which all non-brain voxels have been assigned an intensity of zero are available. Each image is resampled to $1\times 1\times 1$ mm isotropic voxels and is of size $176\times 208 \times 176$.

\vskip1ex
We \textit{evaluate} our approach on four datasets, which all provide brain masks. Although in our study, we focus on T1-weighted images only, our model can be applied to other modalities as long as the PCA model is also built from data acquired by the same modality. The datasets we use for validation are described below.

\vskip1ex
\noindent {\bf IBSR.} The Internet Brain Segmentation Repository (IBSR)~\cite{IBSR} contains MR images from 20 healthy subjects of age 29.1$\pm$4.8 years including their manual brain segmentations, provided by the Center for Morphometric Analysis at Massachusetts General Hospital. All coronal 3D T1-weighted spoiled gradient echo MRI scans were acquired using two different MR systems: ten scans (4 males and 6 females) were performed on a 1.5T Siemens Magnetom MR system (with in-plane resolution of $1\times 1$ mm and slice thickness of 3.1 mm); another ten scans (6 males and 4 females) were acquired from a 1.5T General Electric Signa MR system (with in-plane resolution of $1\times 1$ mm and slice thickness of 3 mm). 

\vskip1ex
\noindent {\bf LPBA40.} The LONI Probabilistic Brain Atlas (LPBA40) dataset of the Laboratory of Neuro Imaging (LONI)~\cite{shattuck2008construction,LPBA40} consists of 40 normal human brain volumes. LPBA40 contains images of 20 males and 20 females of age $29.20 \pm 6.30$ years. Coronal T1-weighted images with slice thickness 1.5 mm were acquired using a 1.5T GE system. Images for 38 of the subjects have in-plane resolution of $0.86\times 0.86$ mm; the images for the remaining two subjects have a resolution of $0.78\times 0.78$ mm. A manually segmented brain mask is available for each image.

\vskip1ex
\noindent {\bf BRATS:} We use twenty T1-weighted image volumes of low and high grade glioma patients from the Brain Tumor Segmentation (BRATS 2016) dataset~\cite{menze2015multimodal} that include cases with large tumors, deformations, or resection cavities. We do not use the BRATS images available as part of the BRATS challenge as these have already been pre-processed (i.e., brain-extracted and co-registered). Instead, we obtain a subset of twenty of the originally acquired images. The BRATS dataset is challenging as the images were acquired with different clinical protocols and various different scanners from multiple ($n=19$) institutions~\cite{bakas2017advancing}. \chg{Our subset of twenty images is from six different institutions.} Furthermore, the BRATS images have comparatively low resolution and some of them contain as few as 25 axial slices (with slice thickness as large as $7$mm). The in-plane resolutions vary from 0.47$\times$0.47 mm to 0.94$\times$0.94 mm with image grid sizes between 256$\times$256 and 512$\times$512 pixels. We manually segment the brain in these images to obtain an accurate brain mask for validation.

\vskip1ex
\noindent {\bf TBI.} Finally, we use our own Traumatic Brain Injury (TBI) dataset which contains 8 TBI images as well as manual brain segmentations. \chg{These are standard MPRAGE~\cite{brant1992mp} T1-weighted images with no contrast enhancement.} They have been resampled to 1$\times$1$\times$1 mm isotropic voxel size with image size between $192\times 228\times 170$ and $256\times 256\times 176$. Segmentations are available for healthy brain, hemorrhage, edema and necrosis. To generate the brain masks, we always use the union of healthy tissue and necrosis. We also include hemorrhage and edema if they are contained within healthy brain tissue.

\begin{figure}[t!]
\includegraphics[width=1\columnwidth]{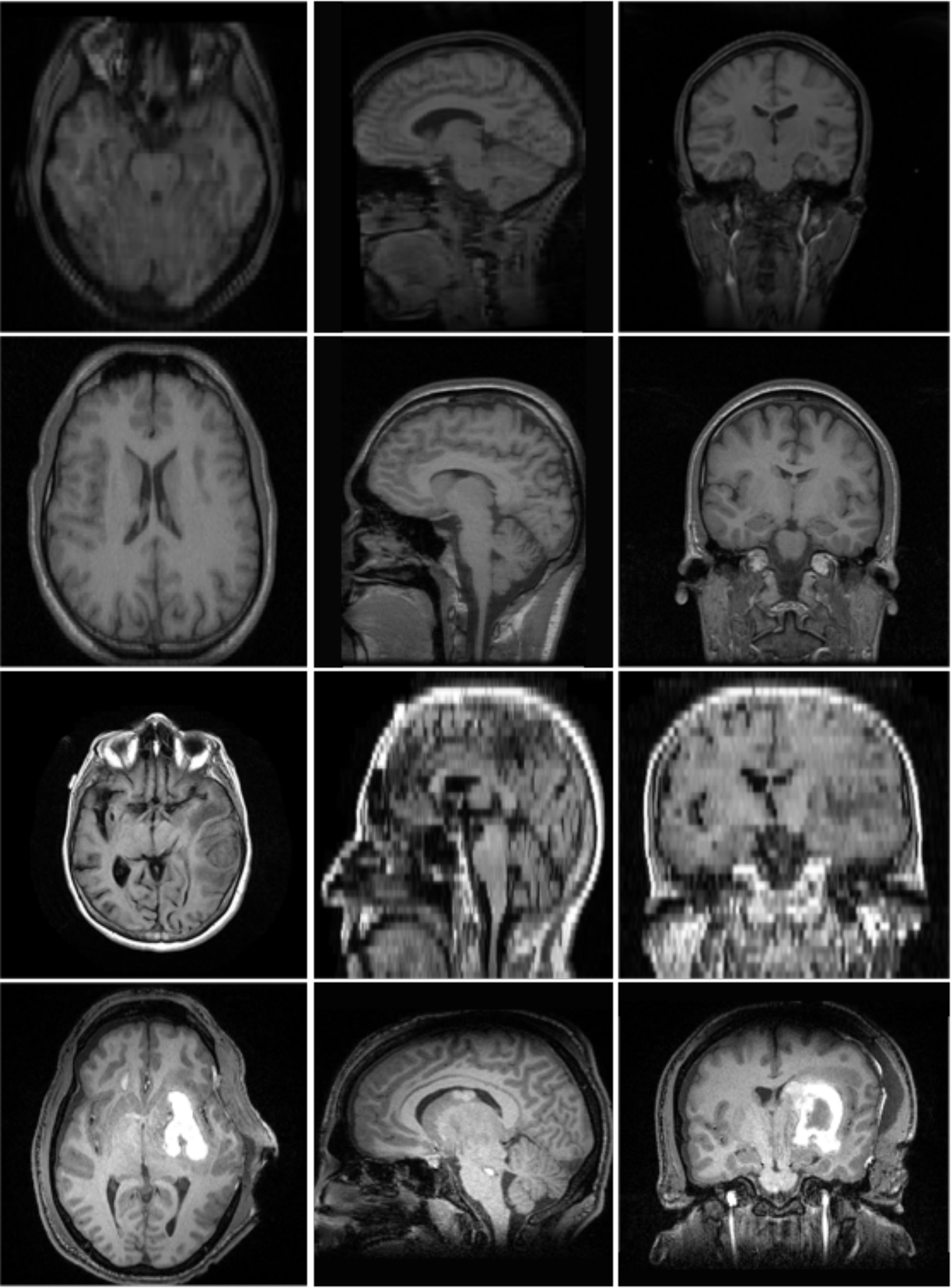}
\caption{\label{fig:example_images} Illustration of image appearance variability on a selection of images from each (evaluation) database. 
From \textit{top} to \textit{bottom}: IBSR, LPBA40, BRATS and TBI.}
\end{figure}

\vskip1ex
Fig.~\ref{fig:example_images} shows example images from each dataset to illustrate image variability. IBSR and LPBA40 contain images from normal subjects and include large portions of the neck; BRATS has very low out-of-plane resolution; and the TBI dataset contains large pathologies and abnormal skulls.

\subsection{Dataset processing}
 
\subsubsection{PCA model}
\label{subsection:pca-model}
We randomly pick 100 images and their brain masks to build our PCA model of the brain. Specifically, we register the brain-masked images to the brain-masked ICBM atlas using a B-spline registration. We use \texttt{NiftyReg}~\cite{modat2010fast} to perform the B-spline registration with local normalized cross-correlation (LNCC) as similarity measure. To normalize image intensities, we apply an affine transform to the image intensities of the warped images so that the 1st percentile is mapped to 0.01 and 99th percentile is mapped to 0.99 and then clamp the image intensities to be within $[0, 1]$. We then perform PCA on the now registered and normalized images and retain the top 50 PCA modes, \chg{which preserve 63\% of the variance}, for our statistical appearance model. This is similar to an active appearance model~\cite{cootes2001active}.

\subsubsection{IBSR refined segmentation}
For IBSR, segmentations of the brain images into white matter, gray matter and cerebrospinal fluid (CSF) are provided. While, in principle, the union of the segmentations of white matter, gray matter and CSF should represent the desired brain mask, this is not exactly the case (see Fig.~\ref{fig:ibsr_seg_refine}). To alleviate this issue for each segmentation, we use morphological closing to fill in remaining gaps and holes inside the brain mask and, in particular, to disconnect the background inside the brain mask from the surrounding image background. \chg{The structuring element for closing is a voxel and its 18 neighborhood}\footnote{The 18-voxel connectivity is also used for other morphological operations in this manuscript.}. We then find the connected component for the background and consider its complement the brain mask. Fig.~\ref{fig:ibsr_seg_refine} shows the pre-processing result after these refinement steps, compared to the original IBSR segmentation (i.e., the union of white matter, gray matter, and the CSF).

\begin{table}[!htb]
	\centering
	\begin{tabular}{@{}c@{}c@{}c@{}}
		\includegraphics[width=0.33\columnwidth]{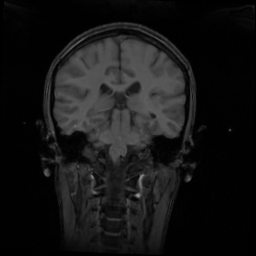} &
		\includegraphics[width=0.33\columnwidth]{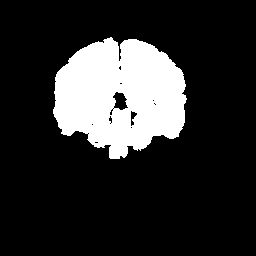} &
		\includegraphics[width=0.33\columnwidth]{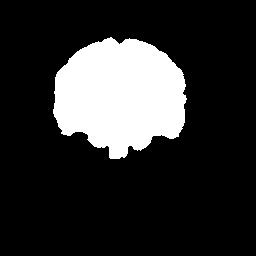}\\
		\small(a) & \small(b) & \small(c)
	\end{tabular}
	\captionof{figure}{Example coronal slice of (a) an IBSR MR brain image, (b) the corresponding original IBSR brain segmentation (i.e., union of white matter, gray matter and CSF) and (c) the refined brain segmentation result.}
	\label{fig:ibsr_seg_refine}
\end{table}

\subsection{Review of related models}

As mentioned previously, brain extraction is challenging because it requires the identification of all non-brain tissue which can be highly variable (cf. Fig.~\ref{fig:example_images}). Our brain extraction approach is based on image alignment to an atlas space where a brain mask is available. However, this requires a reliable registration approach which can tolerate variable image appearance as well as pathologies (i.e., brain tumors, traumatic brain injuries, or general head injuries resulting in skull deformations and fractures). In both cases, no one-to-one mapping between image and atlas space may be available and a direct application of standard image similarity measures for image registration may be inappropriate.

A variety of approaches have been proposed to address the registration of pathological images. For example, cost function masking~\cite{brett2001spatial} and geometric metamorphosis~\cite{niethammer2011geometric} exclude the pathological regions when measuring image similarities. However, these approaches require prior segmentations of the pathologies, which can be non-trivial and/or labor intensive. A conceptually different approach is to learn the normal image appearance from population data and to estimate a quasi-normal image from a pathological image. Then, the quasi-normal image can be used for registration~\cite{yang2016registration}. The low-rank + sparse (LRS) image registration framework, proposed by Liu et al.~\cite{liu2014low}, follows this idea by iteratively registering the low-rank components from the input images to the atlas and then re-computes the low-rank components. After convergence, the image is well-aligned with the atlas.

Our proposed brain extraction model builds upon our previous PCA-based approach for pathological image registration~\cite{han2017efficient} which, in turn, builds upon and removes many shortcomings of the low-rank + sparse approach of Liu et al.~\cite{liu2014low}. We therefore briefly review the low-rank + sparse technique in Sec.~\ref{sec:lrs} and the PCA approach for pathological image registration in Sec.~\ref{sec:pca_tv}. We discuss our proposed model for brain extraction in Sec.~\ref{sec:proposed_method}. 

\subsubsection{Low-Rank + Sparse (LRS)}
\label{sec:lrs}

\vskip0.5ex
\chg{An LRS decomposition aims at minimizing~\cite{wright2009}
\begin{equation}
E(L,S)=\text{rank}(L) + \lambda \|S\|_0\quad\text{s.t.}\quad D = L+S.
\end{equation}
I.e., the goal is to find an additive decomposition of a data matrix $D=L+S$ such that $L$ is low-rank and $S$ is sparse. Here, $\|S\|_0$ denotes the number of non-zero elements in $S$ and $\lambda>0$ weighs the contribution of the sparse part, $S$, in relation to the low-rank part $L$. Neither rank nor sparsity are convex functions. Hence, to simplify the solution of this optimization problem it is relaxed:  the rank is replaced by the nuclear norm and the sparsity term is replaced by the one-norm. As both of these norms are convex and $D=L+S$ is a linear constraint one obtains the convex approximation to LRS decomposition by minimizing the energy}
\begin{equation}
E(L,S) = \|L\|_* + \lambda\|S\|_1,\quad\text{s.t.}\quad D = L + S\enspace, 
\label{eq:lrs}
\end{equation}
where $\|\cdot\|_*$ is the nuclear norm (i.e., a convex approximation for the matrix rank). In imaging applications, $D$ contains all the (vectorized) images: each image is represented as a column of $D$. The low-rank term captures common information across columns. The sparse term, on the other hand, captures uncommon/unusual information. As Eq.~\eqref{eq:lrs} is convex, minimization results in a global minimum.

In practice, applying the LRS model requires forming the matrix $D$ from all the images. $D$ is of size $m\times n$, where $m$ is the number of voxels, and $n$ is the number of images. For 3D images, $m \gg n$ (typically). Assuming all images are spatially well-aligned, $L$ captures the quasi-normal appearance of the images whereas $S$ contains pathologies which are not shared across the images. Of course, in practice, the objective is image alignment and hence the images in $D$ cannot be assumed to be aligned a-priori. Hence, Liu et al.~\cite{liu2014low} alternate LRS decomposition steps with image registration steps. Here the registrations are between all the low-rank images (which are assumed to be approximately pathology-free) and an atlas image. This approach is effective in practice, but can be computationally costly, may require large amounts of memory, and has the tendency to lose fine image detail in the quasi-normal image reconstructions, $L$. In detail, the matrix $D$ has a large number of rows for typical 3D images, hence it can be costly to store. Furthermore, optimizing the LRS decomposition involves a singular value decomposition (SVD) at each iteration with a complexity of $\mathcal{O}(min\{mn^2,m^2n\})$ \cite{holmes2007fast} for an $m \times n$ matrix. While large datasets are beneficial to capturing data variation, the quadratic complexity renders LRS computationally challenging in these situations.

\vskip0.5ex
However, it is possible to overcome many of these shortcomings while staying close to the initial motivation of the original LRS approach. The following Section~\ref{sec:pca_tv} discusses how this can be accomplished.

\subsubsection{Joint PCA-TV model}
\label{sec:pca_tv}

To avoid the memory and computational issues of the low-rank + sparse decomposition discussed above, we previously proposed a joint PCA/Image-Reconstruction model~\cite{han2017efficient} for improved and more efficient registration of images with pathologies. In this model, we have a collection of normal images and register all the normal images to the atlas \textit{once}, using a standard image similarity measure. These normal images do not need to be re-registered during the iterative approach. We mimic the low-rank part of the LRS by a PCA decomposition of the atlas-aligned normal images from which we obtain the PCA basis and the mean image. Let us consider the case when we are now given a \textit{single} pathological image $I$. Let $\hat{I}$ denote the pathological image after subtracting the mean image $M$ and $B$ the PCA basis matrix. $\hat{L}$ and $T$ are images of the same size as $I$\footnote{Images are vectorized for computational purposes, but the spatial gradient $\nabla$ denotes the gradient in the spatial domain.}. Specifically, we minimize
\begin{equation}
\begin{split}
E(T, \hat{L}, \bm{\alpha}) = \frac{1}{2}\|\hat{L}- B\bm{\alpha}\|_2^2 + \gamma \|\nabla T\|_{2,1}, \\
~\text{s.t.}~\hat{I} = \hat{L} + T
\end{split}
\end{equation}
where $\|\nabla T\|_{2,1}=\sum_i \|\nabla T_i\|_2$ and $i$ denotes spatial location. This model is similar to the Rudin-Osher-Fatemi (ROF) image denoising model~\cite{rudin1992nonlinear}. It results in a total variation (TV) term, $T$, which captures the parts of $\hat{I}$ that are (i) relatively large, (ii) spatially contiguous, and (iii) cannot be explained by the PCA basis, e.g., pathological regions. The quasi-low-rank part $\hat{L}$ remains close to the PCA space but retains fine image detail. The quasi-normal image $L$ can then be reconstructed as $L=M+\hat{L}$. We refer to this model as our joint PCA-TV model.

As in the LRS approach, we can register the quasi-normal image $L$ to atlas space and alternate decomposition and registration steps. However, in contrast to the LRS model, the PCA-TV model registers only \textit{one} image ($L$) in each registration step and consequently requires less time and memory to compute. Furthermore, the reconstructed quasi-normal image, $L$, retains fine image detail as pathologies are captured via the total variation term in the PCA-TV model.

\subsection{Proposed brain extraction approach}
\label{sec:proposed_method}

The following sections describe how our proposed brain extraction approach builds upon the principles of the PCA-TV model (Section~\ref{subsection:pca-s-tv}), and discusses image pre-processing (Section~\ref{subsection:preprocessing}), the overall registration framework (Section~\ref{subsection:reg-framework}), and post-processing steps (Section~\ref{subsection:postprocessing}). 

\subsubsection{Joint PCA-Sparse-TV model}
\label{subsection:pca-s-tv}

\vskip0.2ex
The PCA-TV model captures the pathological information well, but it does not model non-brain regions (such as the skull) appropriately. The skull is, for example, usually a thin, shell-shape structure and other non-brain tissue may be irregularly shaped with various intensities. The only commonality is that all these structures surround the brain. Specifically, if a test image is aligned to the atlas well, these non-brain tissues should {\it all} be located outside the atlas' brain mask. Hence, we reject these non-brain regions via a spatially distributed sparse term. We penalize sparsity heavily inside the brain and relatively little on the outside of the brain. This has the effect that it is very cheap to assign voxels outside the brain to the sparse term; hence, these are implicitly declared as brain outliers. Of course, if we would already have a reliable brain mask we would not need to go through any modeling. Instead, we assume that our initial affine registration provides a good {\it initial alignment} of the image, but that it will be inaccurate at the boundaries. We therefore add a constant penalty close to the boundary of the atlas brain mask. Specifically, we create two masks: a two-voxel-eroded brain mask, which we are confident is within the brain and a one-voxel-dilated brain mask, which we are confident includes the entire brain. We then obtain the following model: 

\begin{equation}
\begin{split}
E(S,T,\hat{L},\bm{\alpha}) =  \frac{1}{2}\|\hat{L} - B\bm{\alpha}\|_2^2 & + \gamma\|\nabla T\|_{2,1} \\
& + \|\bm{\Lambda}\odot S\|_1,\\
 ~\text{s.t.}~\hat{I} = \hat{L} + S + T \label{eq:pca-sparse-tv}
\end{split}
\end{equation}
where $\bm{\Lambda} = \Lambda(\bm{x})\geq 0$ is a spatially varying weight 
\begin{equation}
\Lambda(\bm{x}) = \begin{cases}
\infty, &\quad \bm{x} \in \text{Eroded Mask (inside)}\\
\lambda, &\quad \bm{x} \in \text{Dilated Mask and } \\
&\quad \bm{x} \notin \text{Eroded Mask (at boundary)}\\
0, &\quad \bm{x} \notin \text{Dilated Mask (outside)}
\end{cases}
\end{equation}
with $\bm{x}$ denoting the spatial location. Further, in Eq.~\eqref{eq:pca-sparse-tv}, $\odot$ indicates an element-wise product and $\gamma\geq 0$ weighs the total variation term. 

\vskip0.5ex
We refer to this model as our joint PCA-Sparse-TV model. It decomposes the image into three parts. Similar to the PCA-TV model, the quasi-low-rank part $\hat{L}$ remains close to the PCA space and the TV term, $T$, captures pathological regions. Here, the PCA basis is generated from normal images that have been already brain-extracted. Therefore $\hat{L}$ only contains the brain tissue. Different from the previous model, we add a spatially distributed sparse term, $S$, which captures tissue outside the brain, e.g., the skull. In effect, since $\Lambda$ is very large inside the eroded mask, none of the image inside the eroded mask will be assigned to the sparse part. Conversely, all of the image outside the dilated mask will be assigned to the sparse part. We then integrate this PCA-Sparse-TV model into the low-rank registration framework. This includes three parts: pre-processing, iterative registration and decomposition, and post-processing as we will discuss in the following. 

\tikzstyle{block} = [rectangle, draw=none, text centered, rounded corners, minimum height=4em]
\tikzstyle{line} = [draw, -latex']
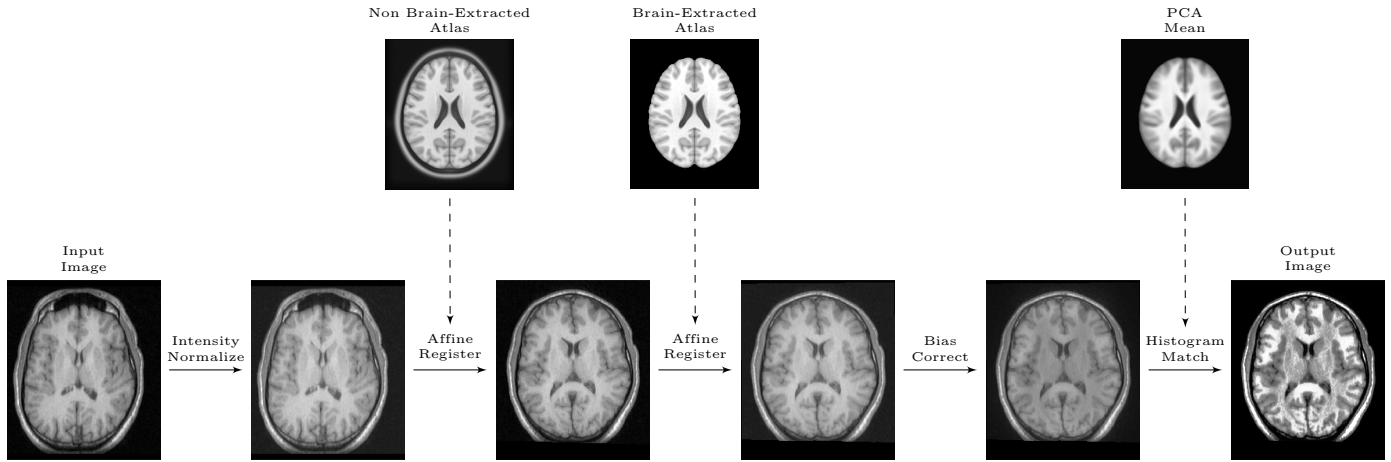
\begin{figure*}
	\centering
	\begin{adjustbox}{width=\textwidth}
		\begin{tikzpicture}[align=center, font=\tiny, node distance = 1cm, every label/.append style={font=\tiny,shift={(0,-0.15)}} ]
		\node[block, label={Input\\Image}] (input) {\includegraphics[scale=0.2]{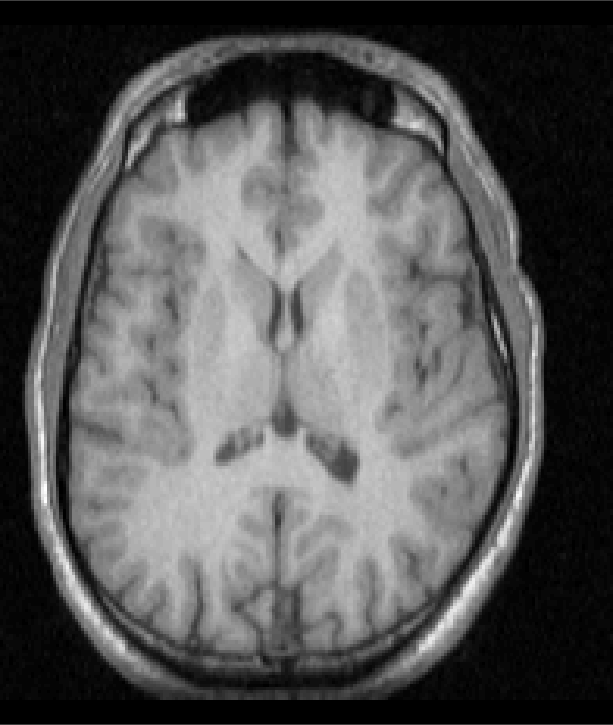}};
		\node[block, right =1cm of input] (normalized) {\includegraphics[scale=0.2]{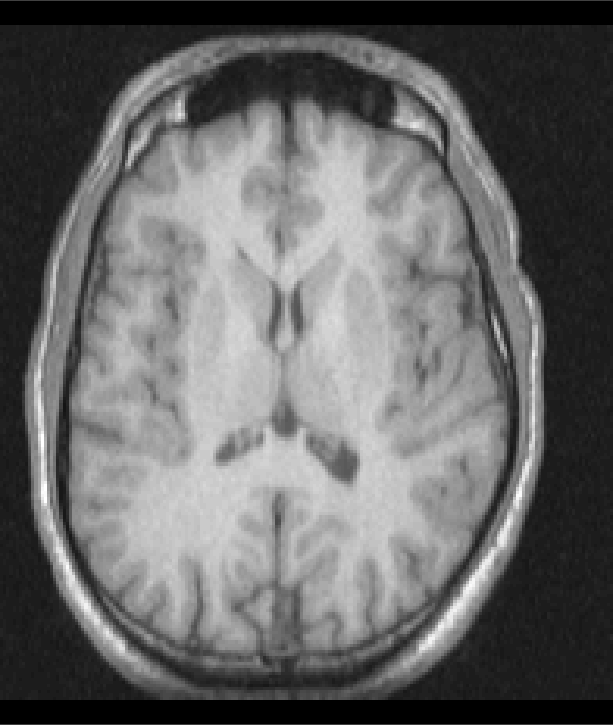}};
		\node[block, label={Non Brain-Extracted\\Atlas}] (atlas1) [above right =1cm and -0.5cm of normalized] {\includegraphics[scale=0.25]{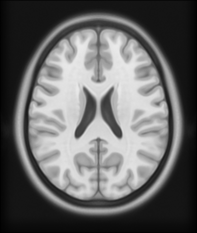}};
		\node[block] (affine1) [right=1cm of normalized] {\includegraphics[scale=0.2]{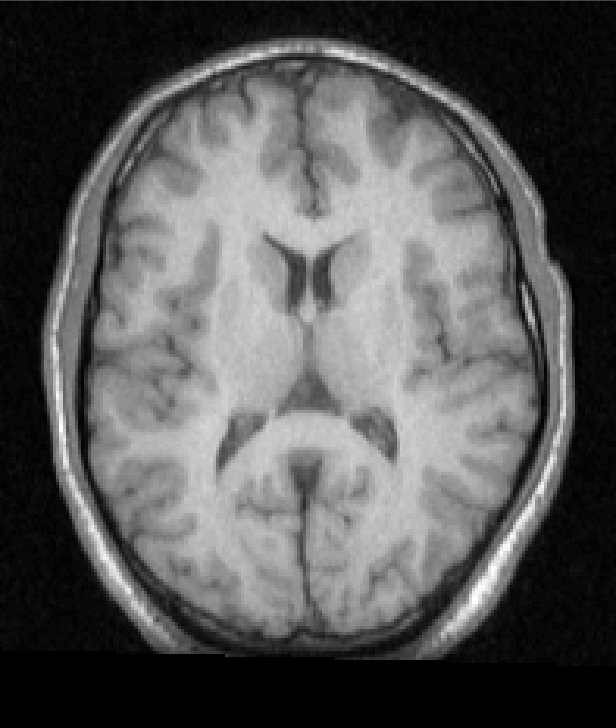}};
		\node[block, label={Brain-Extracted\\Atlas}, above right = 1cm and -0.5cm of affine1] (atlas2) {\includegraphics[scale=0.25]{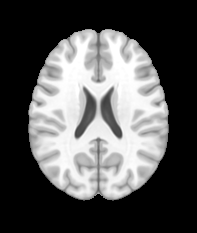}};
		\node[block, right = 1cm of affine1] (affine2) {\includegraphics[scale=0.2]{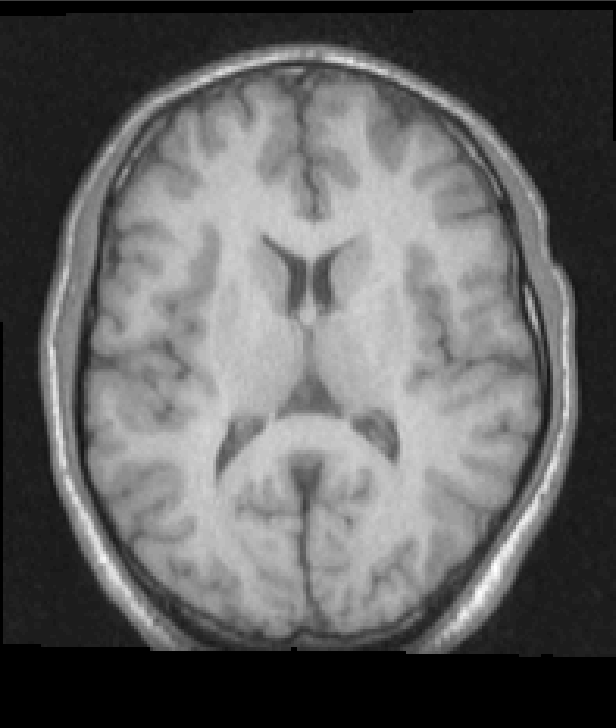}};
		\node[block, right = 1cm of affine2] (bias) {\includegraphics[scale=0.2]{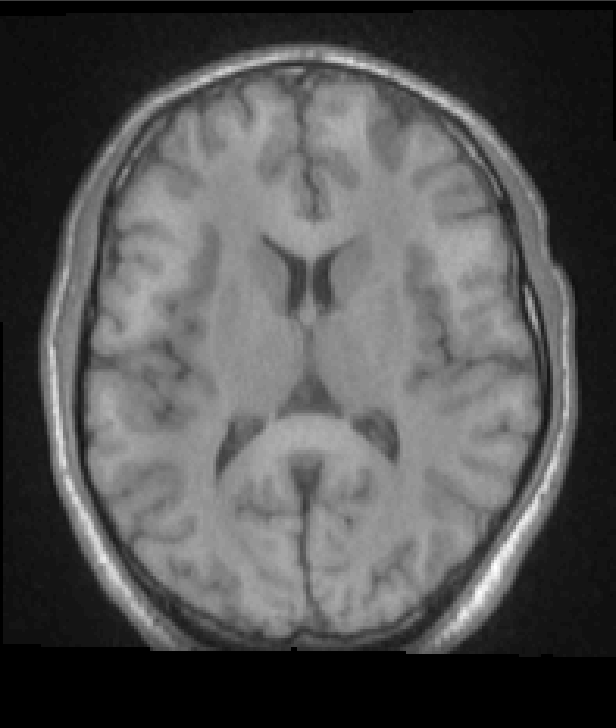}};
		\node[block, label={PCA\\Mean}, above right = 1cm and -0.5cm of bias] (mean) {\includegraphics[scale=0.25]{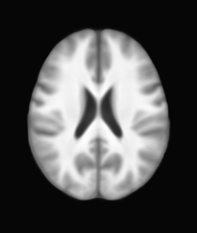}};    
		\node[block, label={Output\\Image}, right = 1cm of bias] (matched) {\includegraphics[scale=0.2]{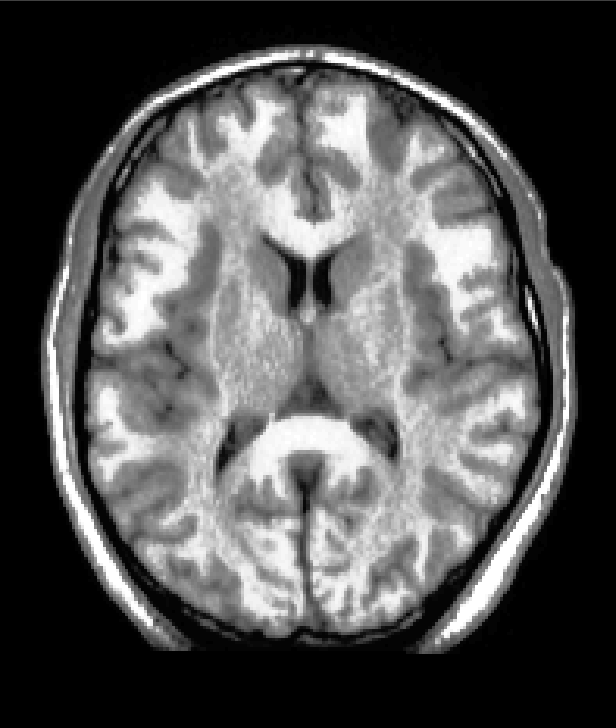}};
		
		\path [line] (input) -- node(i) [midway,above] {Intensity\\Normalize}(normalized);		
		\path [line] (normalized) -- node(a1) [midway,above] {Affine\\Register}(affine1);
		\path [line] (affine1) -- node(a2) [midway,above] {Affine\\Register}(affine2);
		\path [line] (affine2) -- node(b) [midway,above] {Bias\\Correct}(bias);
		\path [line] (bias) -- node(h) [midway,above] {Histogram\\Match}(matched);
		\path [line, dashed](atlas1) -- (a1);
		\path [line, dashed](atlas2) -- (a2);
		\path [line, dashed](mean) -- (h);
		\end{tikzpicture}
	\end{adjustbox}
	\caption{Preprocessing flow chart: Input image is the original image. Eventually, the output image will be fed into the registration/decomposition framework.}
        \label{fig:preprocessing_flowchart}
\end{figure*}

\subsubsection{Pre-processing}
\label{subsection:preprocessing}
\vskip0.5ex

Fig.~\ref{fig:preprocessing_flowchart} shows a flowchart of our pre-processing approach as discussed in the following paragraphs.

\vskip1ex
\noindent
\textbf{Intensity normalization.} Given a test image from which we want to extract the brain, we first affinely transform the image intensities to standardize the intensity range to $[0,1000]$. \chg{Note that our PCA model of section~\ref{subsection:pca-model} is build based on images with intensities standardized to $[0,1]$. The different standardization is necessary here as the bias field correction algorithm removes negative and small intensity values ($<1$) followed by a log transform of the intensities.} Specifically, we first compute the 1st and the 99th percentile of the voxel intensities. We then affinely transform the image intensities of the entire image such that the intensity of the 1st percentile is mapped to 100 and of the 99th percentile to 900. As this may result in intensities smaller than zero or larger than 1000 for the extreme ends of the intensity distribution, we \chg{clamp} the intensities to be within $[0,1000]$.   

\vskip1ex
\noindent
\textbf{Atlas registration.} Next, we first align the intensity-normalized input image to the non brain-extracted atlas. Then, we affinely register the result from the first step to the brain-extracted atlas, but this time using a one-voxel-dilated brain mask in atlas space; this step has the effect of ignoring parts of the image which are not close to the brain in the registration and it gives us a better alignment in the brain region. For both steps we use \texttt{reg\_aladin} of \texttt{NiftyReg}~\cite{modat2014global} disabling symmetric registration (\texttt{-noSym}). The first registration initializes the transformation using the center of gravity (CoG) of the image. \chg{Note that the differing intensity range of the atlas and the image is immaterial in this step as the registration uses local normalized cross-correlation as the similarity measure.}

\vskip1ex
\noindent
\textbf{Bias field correction.} Next, we use N4ITK~\cite{tustison2010n4itk}, a variant of the popular non-parametric non-uniform intensity normalization (N3) algorithm~\cite{sled1998nonparametric}, to perform bias field correction. As the image has been affinely aligned to the atlas in the previous step, we use our two-voxel-eroded brain mask as the region for bias field estimation. Specifically, we use the \texttt{N4BiasFieldCorrection} function in \texttt{SimpleITK}~\cite{lowekamp2013design}, with its default settings.

\vskip1ex
\noindent
\textbf{Histogram matching:} The final step of the pre-processing is histogram matching. We match the histograms of the bias corrected image with the histogram of the mean image of the population data only within the two-voxel-eroded brain mask. This histogram matched image is then the starting point for our brain extraction algorithm \chg{and it is now in an intensity range comparable to the PCA model.}

\subsubsection{Registration framework}
\label{subsection:reg-framework}

Similar to the PCA-TV model, we alternate between \textit{image decomposition} steps using the PCA-Sparse-TV model and \textit{registration to the brain-extracted atlas}. We use a total of six iterations in our framework. In the first iteration ($k=1$), the images are in the original space. We decompose the input image $I_1=I$, into the quasi-normal ($L_1=\hat{L}_1+M$), sparse ($S_1$), and total variation ($T_1$) images by minimizing the energy from Eq. \eqref{eq:pca-sparse-tv}. We then obtain a pathology-free or pathology-reduced image, $R_1$, by adding the sparse and the quasi-normal images of the decomposition: $R_1 = L_1+S_1$. 
	
For the next two iterations ($k = \{2,3\}$), we first find the affine transform $\Phi_k^{-1}$ by affinely registering the pathology-reduced images from the previous iteration, $R_{k-1}$ (i.e., $R_{k-1}=L_{k-1}+S_{k-1}$), to the brain-extracted atlas\footnote{\chg{We follow standard image-registration notation. I.e., a map $\Phi^{-1}$ is defined in the space an image is deformed to. For us this is the space of the atlas image. Conversely, $\Phi$ maps an image from the atlas space back into the original image space and hence is defined in the original image coordinate space.}}. We use the one-voxel-dilated brain mask for cost-function masking which allows the registration to focus only on the brain tissue. \chg{This is important as the first few registrations will not be very precise as they are only based on an affine deformation model. The main objective is to reduce the pathology within the brain. Only after these initial steps, when a good initial alignment has already been obtained, we use the quasi-normal image (excluding the non-brain regions) to perform the registration.} We then apply the transform $\Phi_k^{-1}$ to transform the previous input images to atlas space and obtain new input images, $I_{k}$, (i.e., $I_k = I_{k-1}\circ\Phi_k^{-1} $). We minimize Eq.~\eqref{eq:pca-sparse-tv} again to obtain new decomposition results ($L_k, S_k, T_k$). These decomposition/affine-registration steps are repeated two times, which is empirically determined to be sufficient for convergence. These affine registration steps result in a substantially improved alignment in comparison to the initial affine registration by itself. 
	
The last three iterations ($k=\{4,5,6\}$) repeat the same process, but are different in the following aspects: (i) we now use a B-spline registration instead of the affine registration;  (ii) we use the pathology-reduced image and cost function masking only for the first B-spline registration step, as we did in the previous affine steps. For the remaining two steps, we use the quasi-normal images $L_{k:k=\{5,6\}}$ as the moving images and we do not use the mask during the registrations. The use of the mask is no longer necessary as registrations are now performed using the quasi-normal image; (iii) we use the non-greedy registration strategy of the original low-rank + sparse framework~\cite{liu2015low}, in which we deform the quasi-normal image back to the image space of the third iteration (after the affine steps) in order to avoid accumulating deformation errors.

These steps further refine the alignment, in particular, close to the boundary of the brain mask. After the last iteration, the image is well-aligned to the atlas and we have all the transforms from the original image space to atlas space. As a side effect, the algorithm also results in a quasi-normal reconstruction of the image, $L_6$, an estimate of the pathology, $T_6$, and an image of the non-brain tissue $S_6$, all in atlas space. 

\begin{algorithm*}
	\SetKwInOut{Input}{Input}
	\Input{Image $I$, Brain-Extracted Atlas $A$, Atlas Mask $A_M$}
	\SetKwInOut{Output}{Output}
	\Output{Brain-Extracted Image $I_B$ and mask $I_M$}
	$I_1$, $\Phi_1^{-1}$ = pre-processing($I$);\\
	\For{$k\gets1$ \KwTo $6$}{
		\uIf{$k \geq 2$}{
			\uIf{$k \leq 3$}{
			    find $\Phi_k^{-1}$, s.t., $R_{k-1} \circ \Phi_k^{-1}$ = $A$ and $\Phi_k^{-1}$ is affine;\\
		    }
		    \uElseIf{$k == 4$}{
		    	find $\Phi_k^{-1}$, s.t., $R_{k-1} \circ \Phi_k^{-1}$ = $A$ and $\Phi_k^{-1}$ is B-spline;\\
		    	}
		    \Else{
		    	find $\Phi_k^{-1}$, s.t., ($L_{k-1} \circ \Phi_{k-1}) \circ \Phi_k^{-1}$ = $A$ and $\Phi_k^{-1}$ is B-spline;\\
		    	}	
		    $I_k = I_{k-1}\circ\Phi_k^{-1}$;\\
		}
		Decompose $I_k$, s.t., $I_k = L_k + S_k + T_k$;\\
		\If{$k \leq 3$}{
		    $R_k = L_k + S_k$;\\
	    }
	}
	$I_B, I_M$ = post-processing($A_M$, $\{\Phi_k^{-1}\}$).
	\caption{\chg{Algorithm for Brain Extraction}}
	\label{alg: brain-extraction}
\end{algorithm*}

\subsubsection{Post-processing}
\label{subsection:postprocessing}

Post-processing consists of applying to the atlas mask the inverse transforms of the affine registrations in the pre-processing step and the inverse transforms of the registrations generated in the framework described in section~\ref{subsection:reg-framework}. The warped-back atlas mask is the brain mask for the original image. To extract the brain in the original image space, we simply apply the brain mask on the original input image. All subsequent validations are performed in the original image space.

\vskip1ex
\noindent
\chg{Algorithm~\ref{alg: brain-extraction} summarizes these steps as pseudo-code.}

\section{Experimental results}
\label{section: experiments}

\chg{The following experiments are for brain-extraction from T1-weighted MR images. However, our method can be easily adapted to images from other modalities, as long as the atlas image and the images from which the PCA basis is computed are from the same modality.}

\subsection{Experimental setup}

We evaluate our method on all four evaluation datasets. For comparison, we also assess the performance of BET, BSE, ROBEX, BEaST, MASS and CNN on these datasets. We use BET v2.1 as part of FSL 5.0, BSE v.17a from BrainSuite, ROBEX v1.2, BEaST (\texttt{mincbeast}) v1.90.00, and MASS v1.1.0. We solve our PCA model via a primal-dual hybrid gradient method~\cite{goldstein2013adaptive}. In addition, we implement the decomposition on the GPU and run it on an NVIDIA Titan X GPU~\cite{nickolls2008scalable}~\cite{givon_scikit-cuda_2015}.

\subsection{Evaluation Measures}

We evaluate the brain extraction approaches using the measures listed below.
\paragraph{Dice coefficient} 
Given two sets X and Y (containing the spatial voxel positions of a segmentation), the Dice coefficient $D(X,Y)$ is defined as 
\begin{equation}
D(X,Y) = \frac{2|X\cap Y|}{|X| + |Y|},
\end{equation}
where $X\cap Y$ denotes set intersection between $X$ and $Y$ and $|X|$ denotes the cardinality of set $X$.

\paragraph{Average, maximum and 95\% surface distance} We also measure the symmetric surface distances between the automatic brain segmentation and the gold-standard brain segmentation. This is defined as follows: the distance of a point $x$ to a set of points (or set of points of a triangulated surface $S_A$) is defined as
\begin{equation}
d(x,S_A) = \min_{y\in S_A} d(x,y),
\end{equation}
where $d(x,y)$ is the Euclidean distance between the point $x$ and $y$. The average symmetric surface distances between two surfaces $S_A$ and $S_B$ is then defined as
\begin{equation}
\begin{split}
ASD(S_A,S_B) &=\\ \frac{1}{|S_A| + |S_B|}&\times(\sum_{x\in S_A}d(x, S_B) + \sum_{y\in S_B}d(y, S_A)),
\end{split}
\end{equation}
where $|S_A|$ denotes the cardinality of $S_A$~\cite{yeghiazaryan2015overview} (i.e., number of elements if represented as a set or surface area if represented in the continuum). To assess behavior at the extremes, we also report the maximum symmetric surface distance as well as the 95th percentile symmetric surface distance, which is less prone to outliers. These are defined in analogy, i.e., by computing all distances from surface $S_A$ to $S_B$ and vice versa followed by the computation of the maximum and the 95th percentile of these distances.

\paragraph{Sensitivity and specificity} We also measure sensitivity, i.e., true positive (TP) rate and specificity, i.e., true negative (TN) rate. Here TP denotes the brain voxels which are correctly labeled as brain; TN denotes  the non-brain voxels correctly labeled as such. Furthermore, the false negatives (FN) are the brain voxels incorrectly labeled as non-brain and the false positives (FP) are the non-brain voxels which are incorrectly labeled as brain. Let $V$ be the set of all voxels of an image,  and $X$ and $Y$ the automatic brain segmentation and gold-standard brain segmentation, respectively. The sensitivity and specificity are then defined as follows~\cite{sonka2000handbook} :
\begin{align}
  sensitivity = \frac{TP}{TP+FN} &=& \frac{|X\cap Y|}{|Y|}\\
  specificity = \frac{TN}{TN+FP} &=& \frac{|V|-|X\cup Y|}{|V|-|Y|}
\end{align}

\subsection{Datasets of normal images: IBSR/LPBA40}

\begin{table*}[!htb]
	\centering
	\begin{tabular}{@{}c@{}c@{}c@{}}
		\includegraphics[width=0.7\columnwidth]{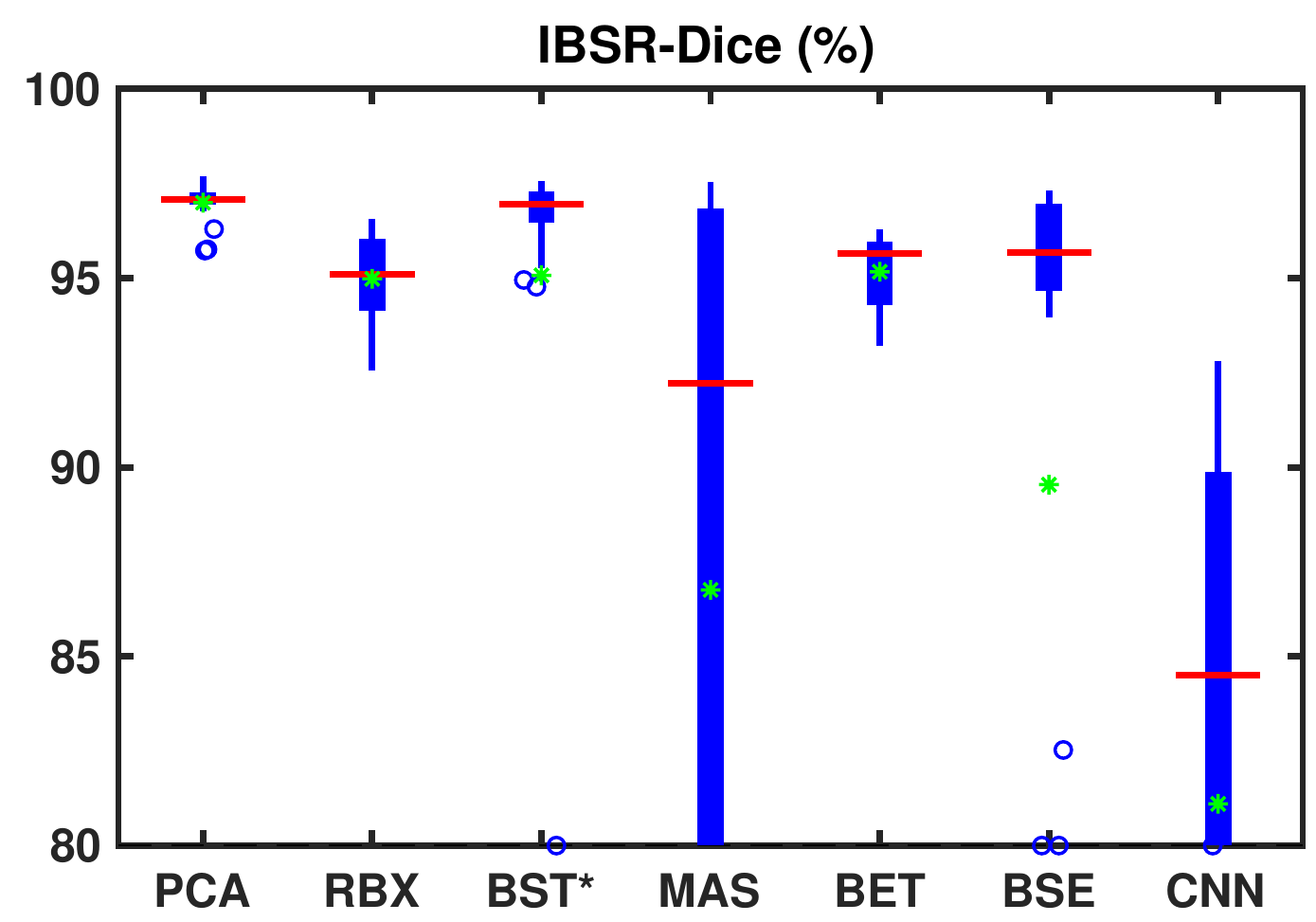} &
		\includegraphics[width=0.7\columnwidth]{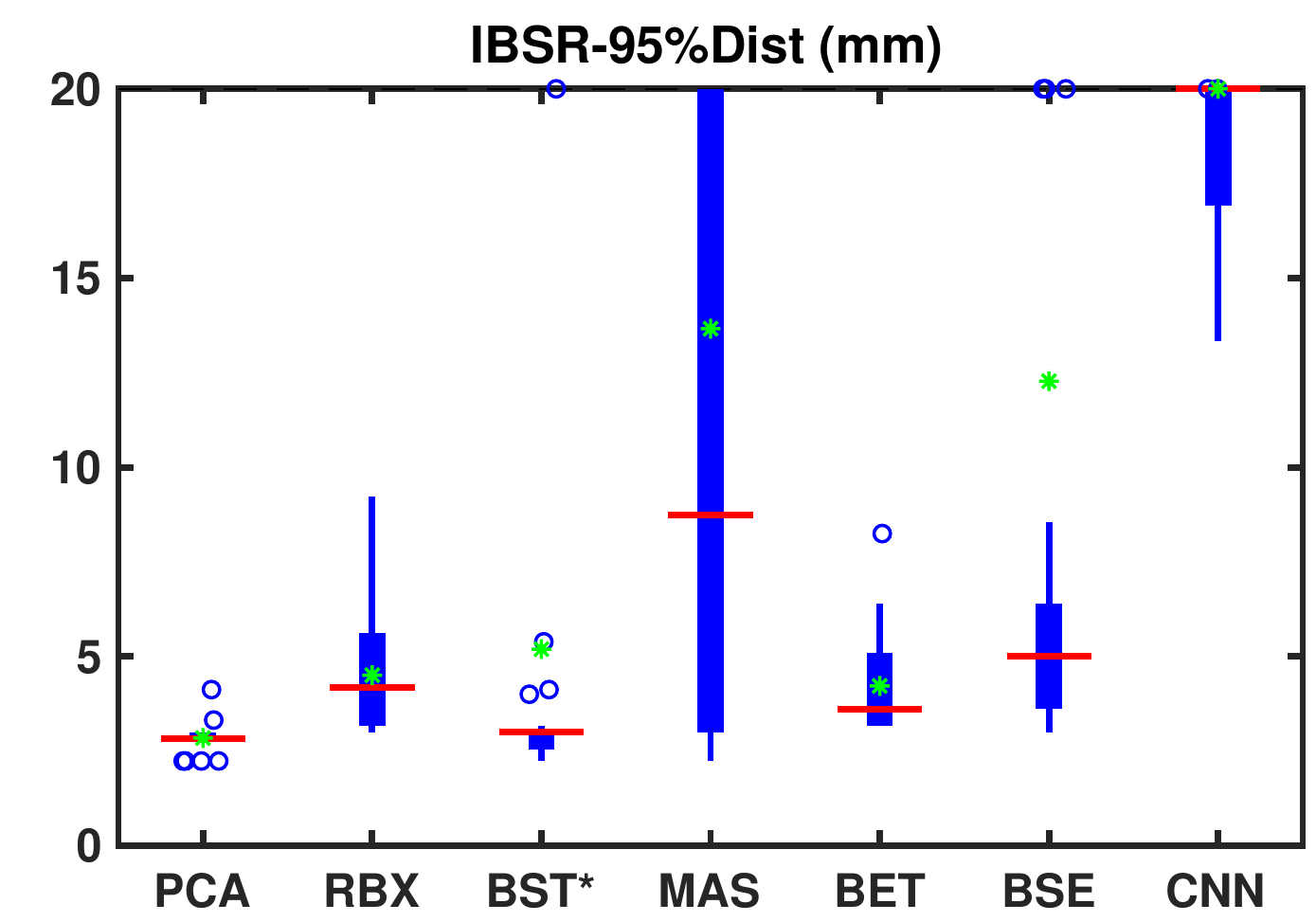} &
		\includegraphics[width=0.7\columnwidth]{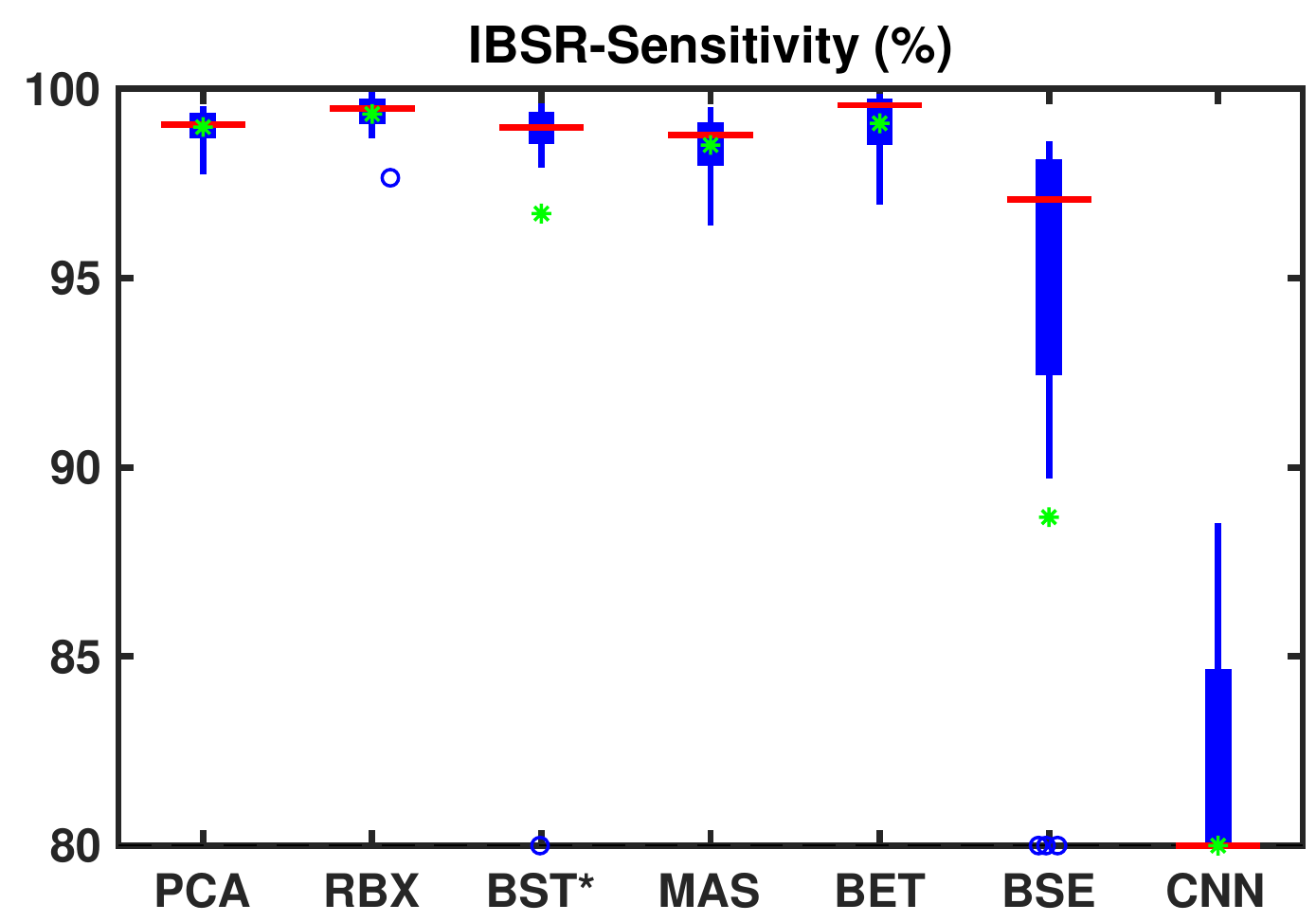} \\
		\includegraphics[width=0.7\columnwidth]{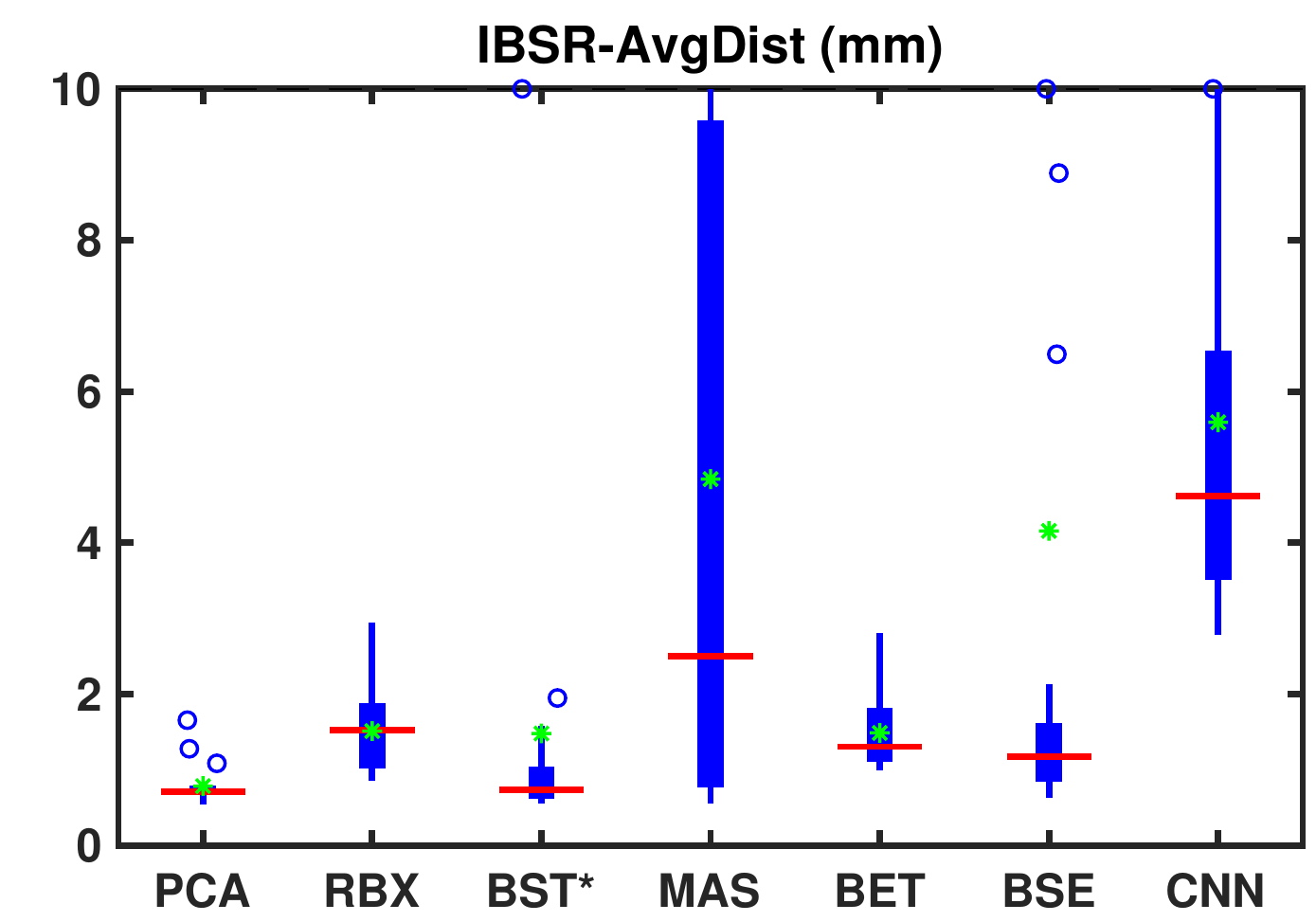} &
		\includegraphics[width=0.7\columnwidth]{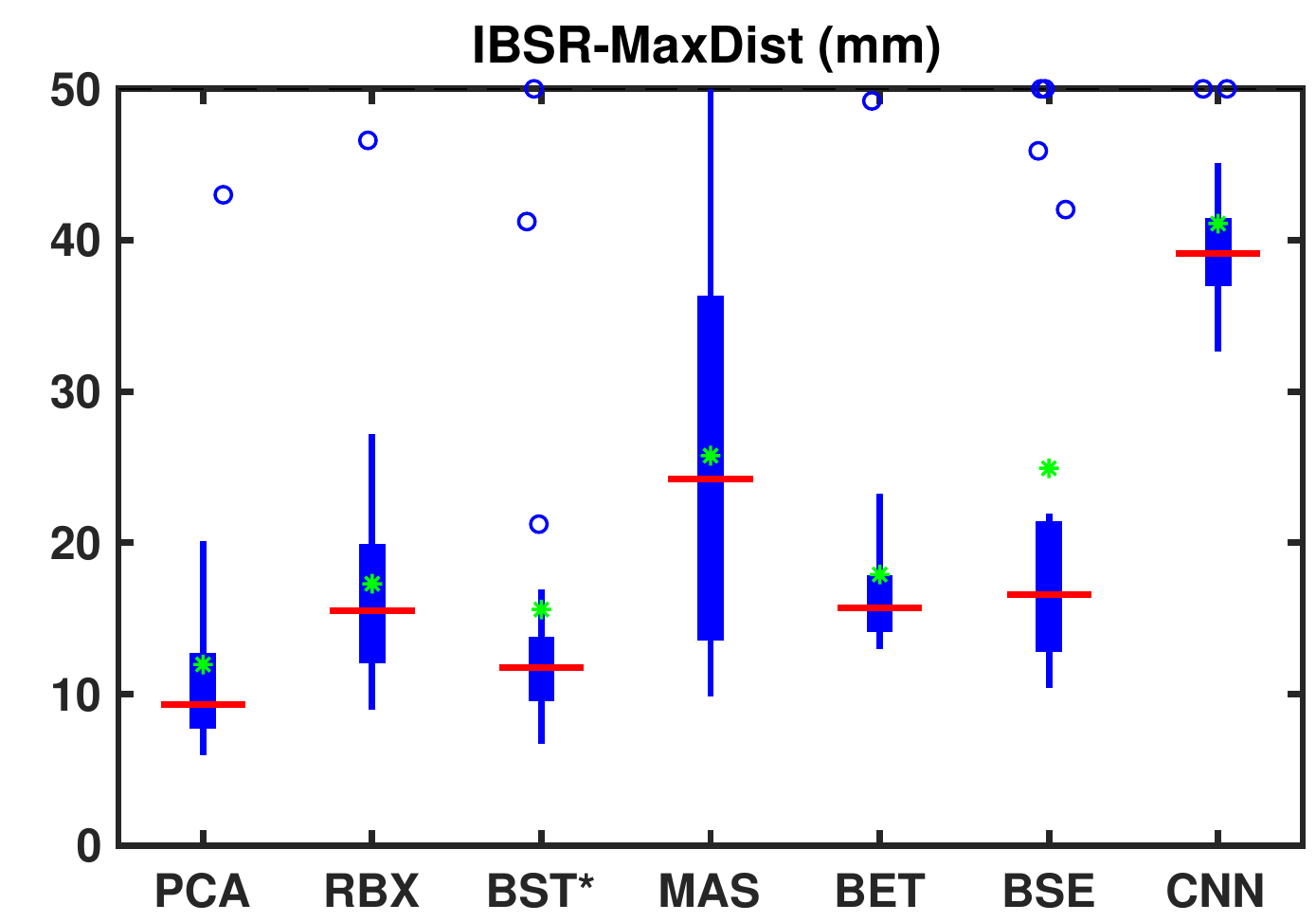} &
		\includegraphics[width=0.7\columnwidth]{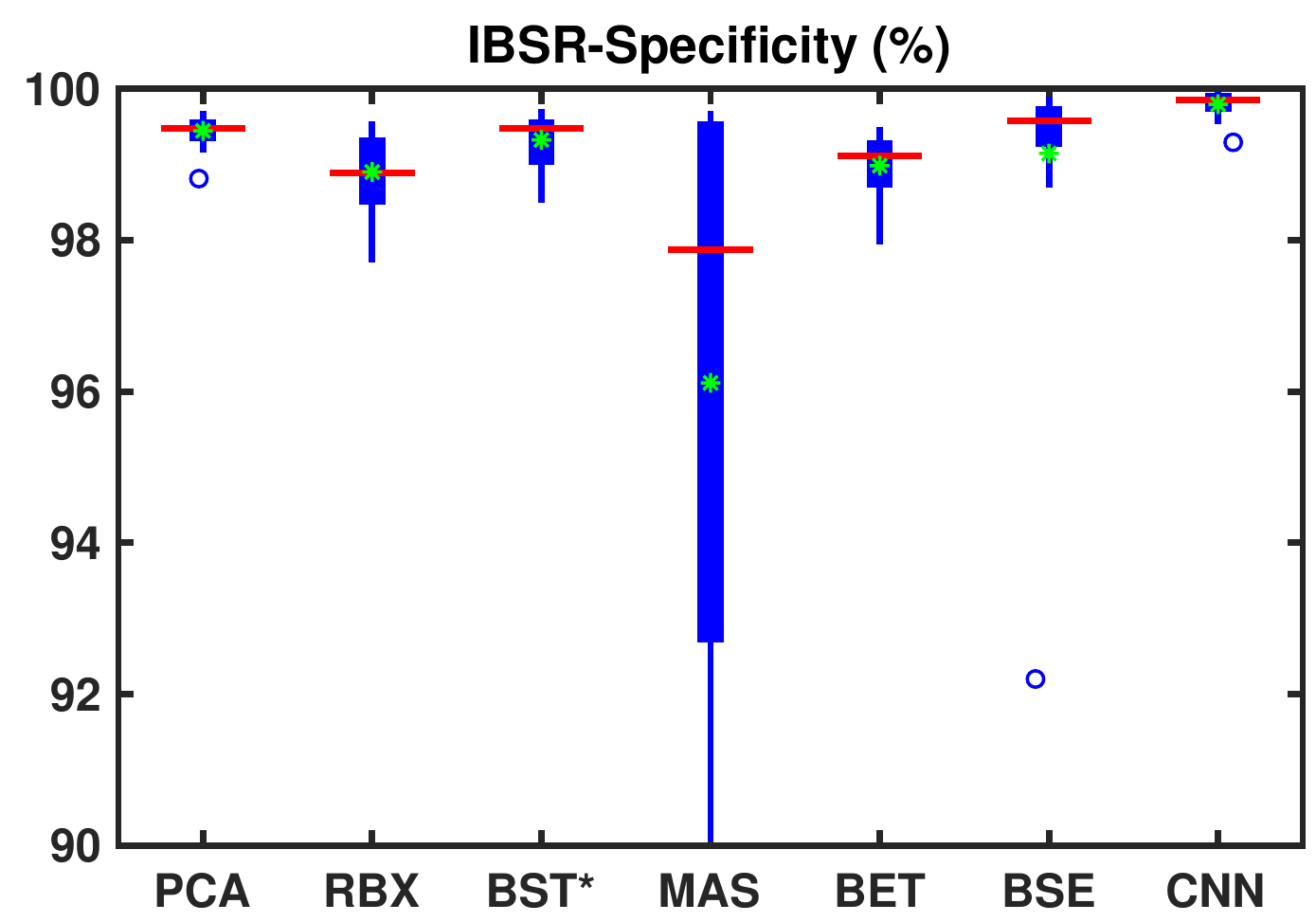}
	\end{tabular}
	\captionof{figure}{Box plot results for the IBSR normal dataset. \xhsp{We show the results from seven methods: PCA, RBX (ROBEX), BST* (BEaST*), MAS (MASS), BET, BSE and CNN.} Due to the poor results of MASS and CNN, and the outliers of BSE on this dataset, we limit the range of the plots for better visibility. On each box, the center line denotes the median, and the top and the bottom edge denote the 75th and 25th percentile, respectively. The whiskers extend to the most extreme points that are not considered outliers. The outliers are marked with `+' signs. \xhsp{In addition, we mark the mean with green `*' signs.} ROBEX, BET, and BSE show similar performance, but BSE exhibits two outliers. MASS works well on most images, but fails on many cases. \xhsp{BEaST fails on the original images. We therefore show the BEaST* results using the initial affine registration of our PCA model. BEaST* performs well with high Dice scores and low surface distances, but with low mean values.} CNN performs poorly on this dataset. \xhsp{Our PCA model has similar performance to BEaST* but with higher mean values. Both methods perform better than other methods on the Dice scores and surface distances.}}
	\label{fig:ibsr_box_plot}
	\vskip0.5ex
	\centering
	\begin{tabular}{@{}c@{}c@{}c@{}}
		\includegraphics[width=0.7\columnwidth]{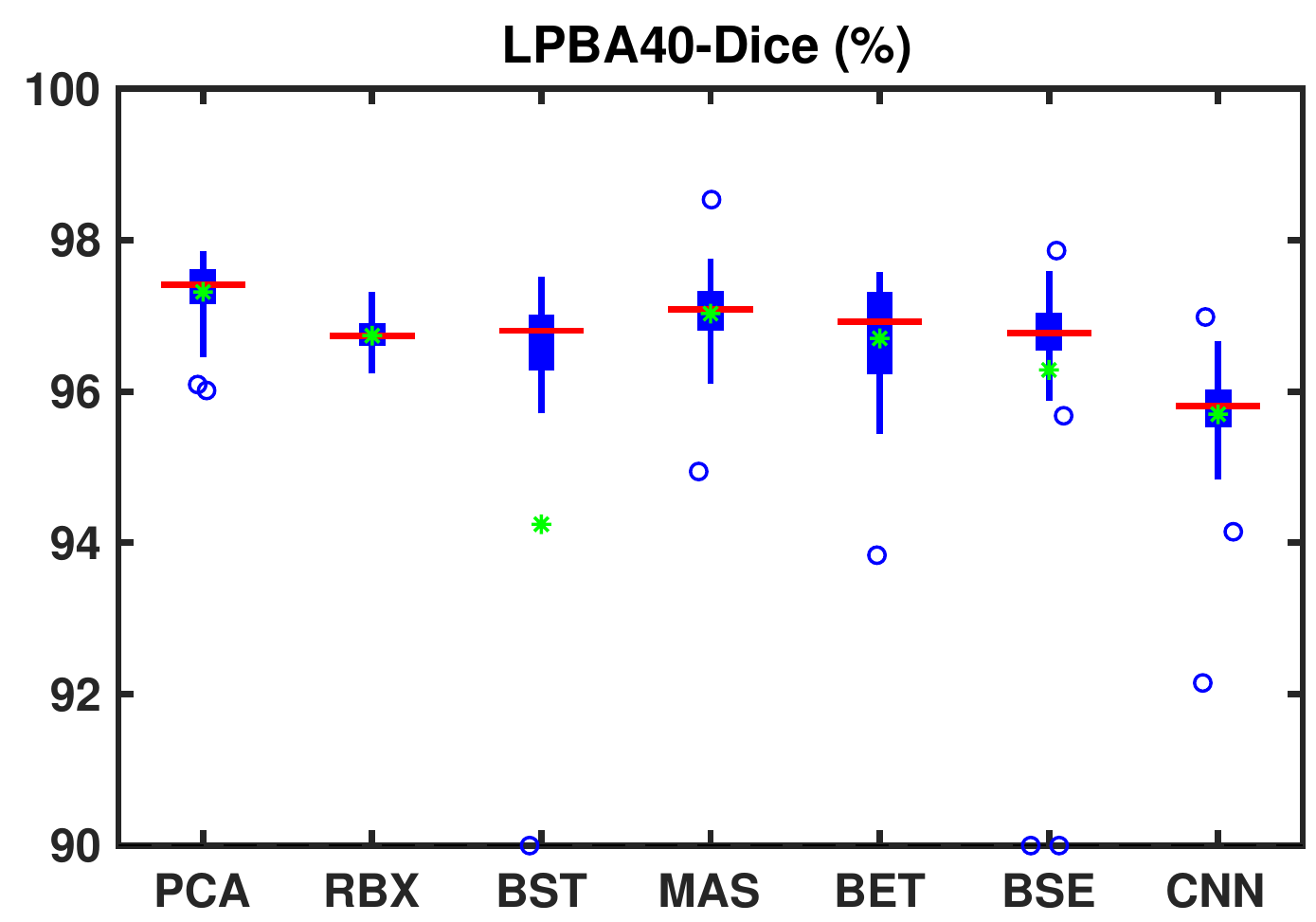} &
		\includegraphics[width=0.7\columnwidth]{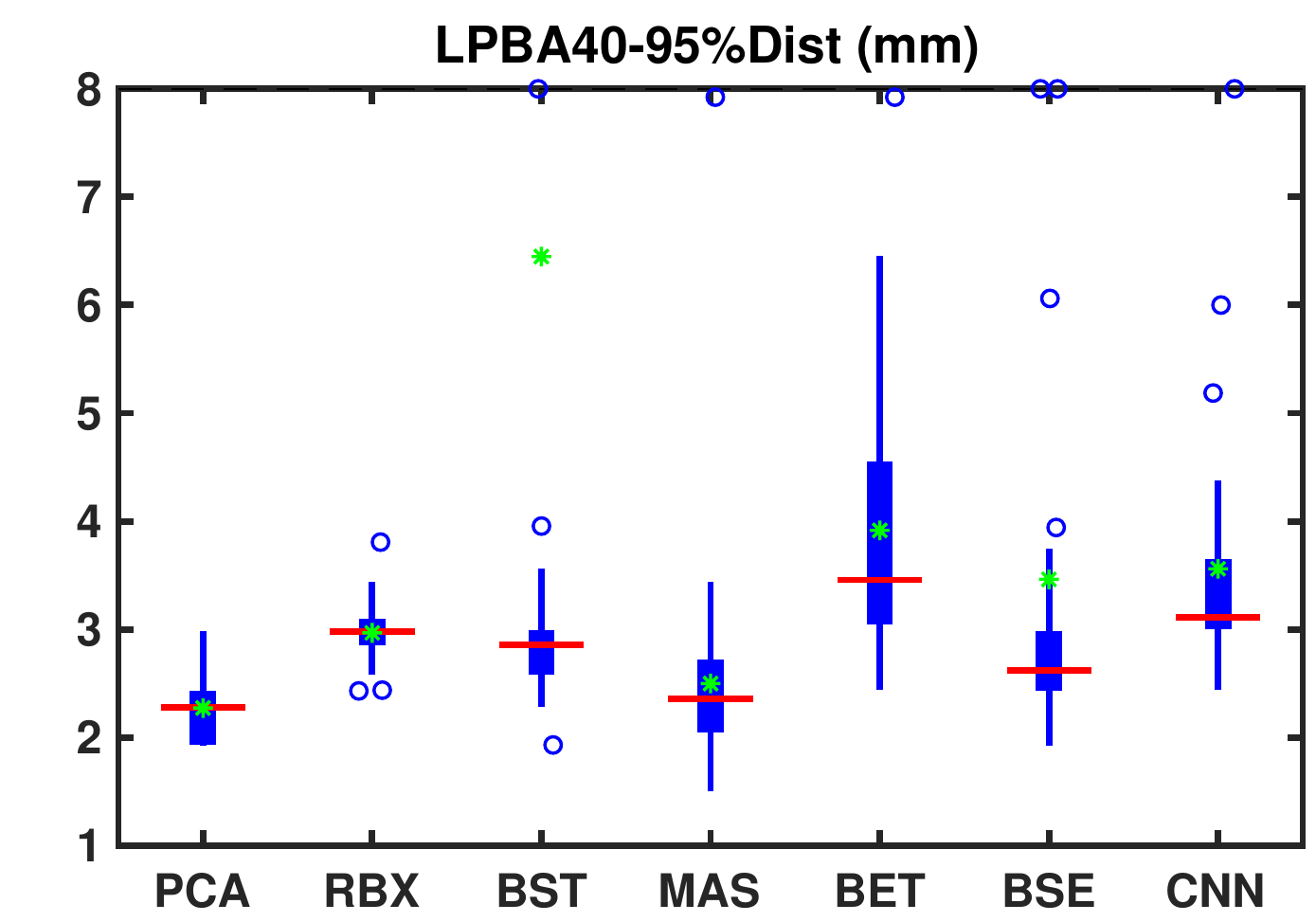} &
		\includegraphics[width=0.7\columnwidth]{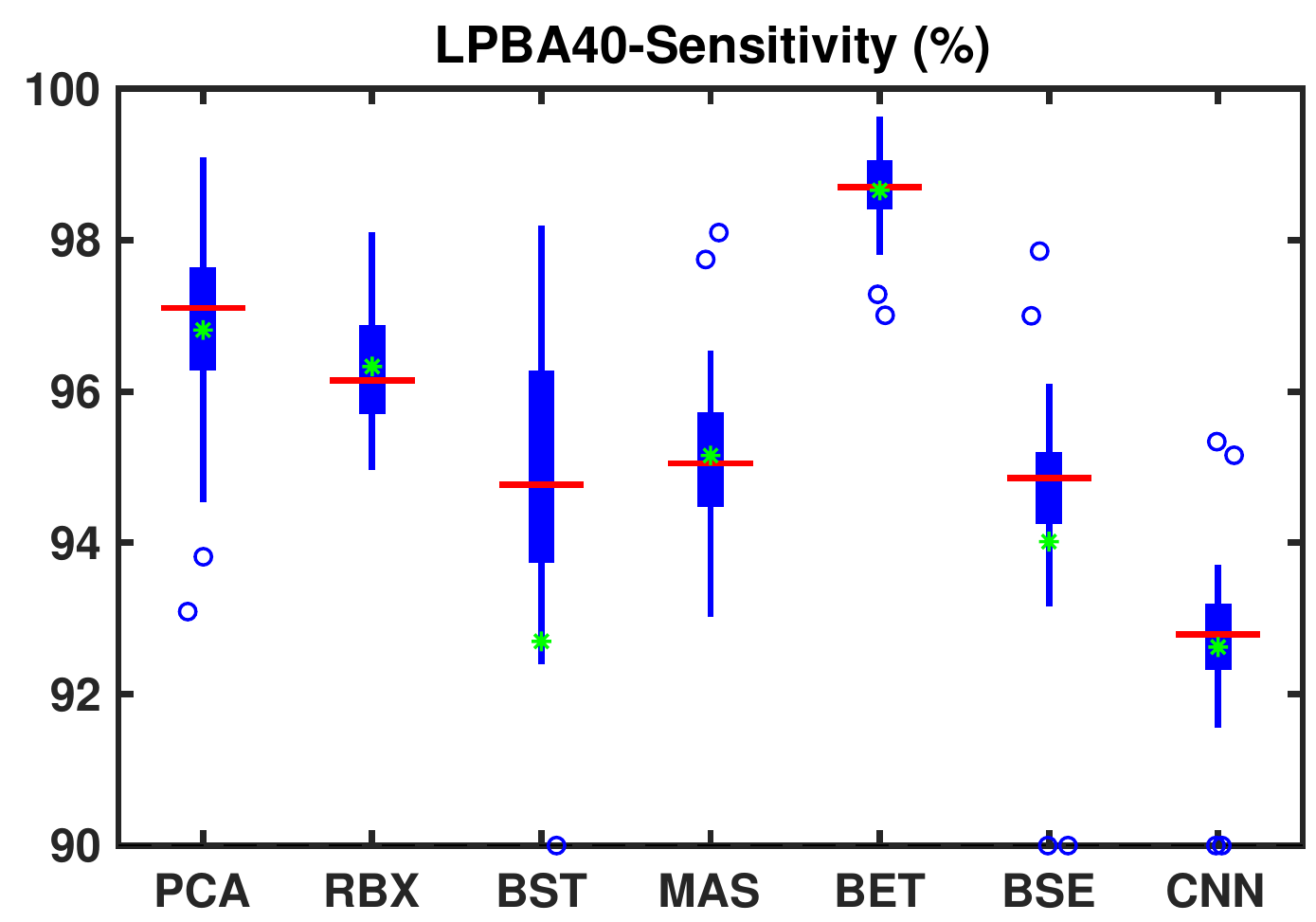} \\
		\includegraphics[width=0.7\columnwidth]{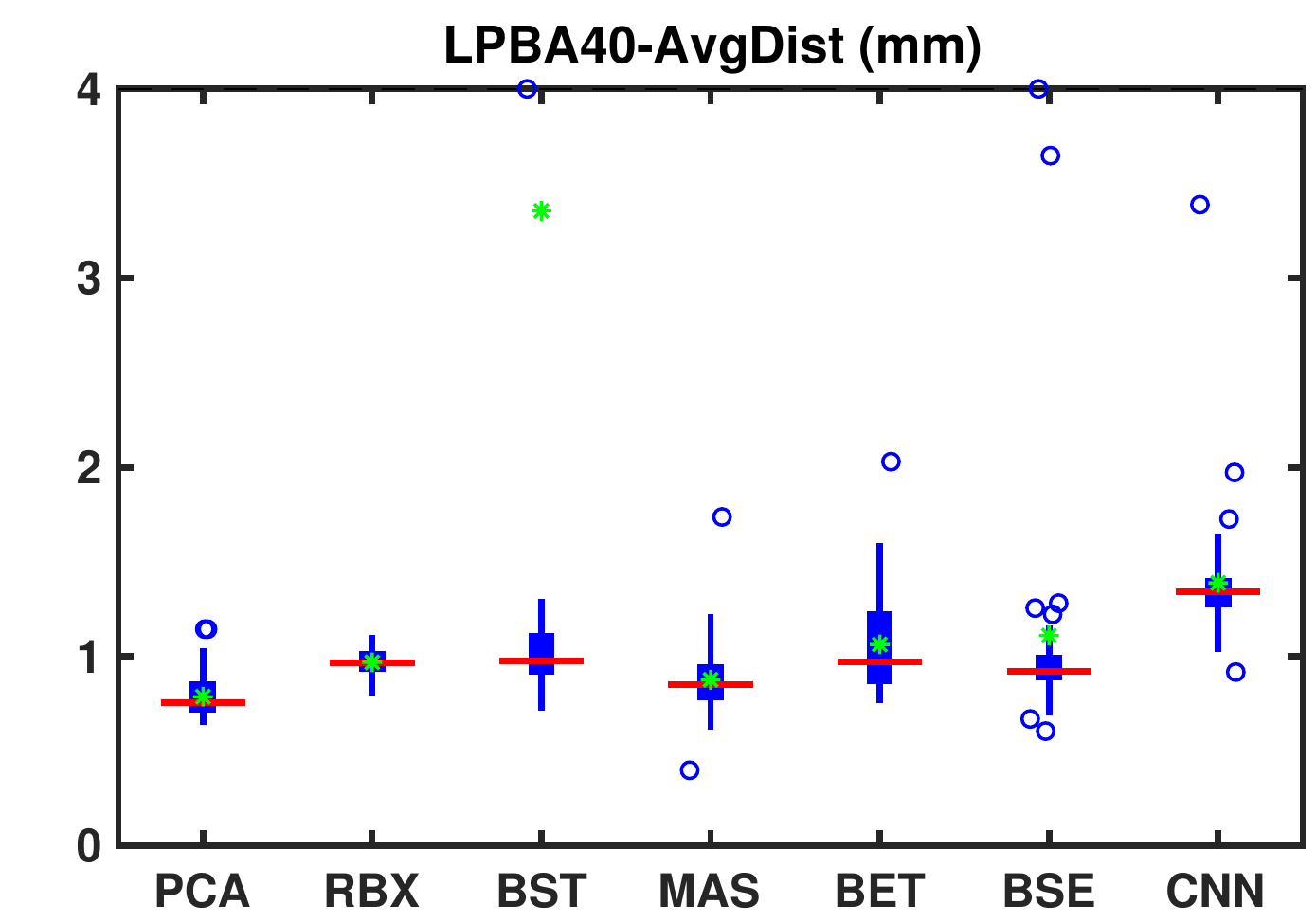} &
		\includegraphics[width=0.7\columnwidth]{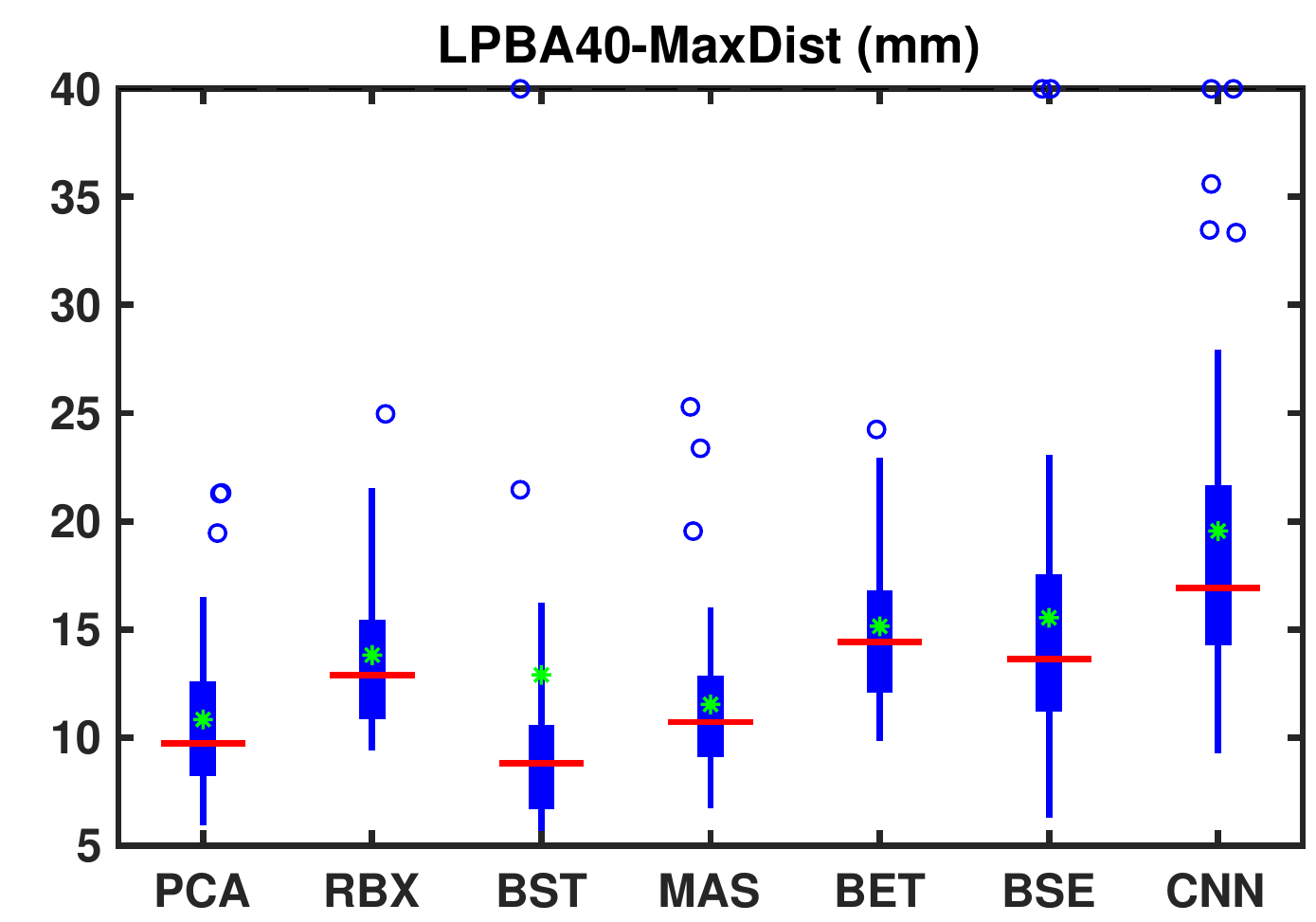} &
		\includegraphics[width=0.7\columnwidth]{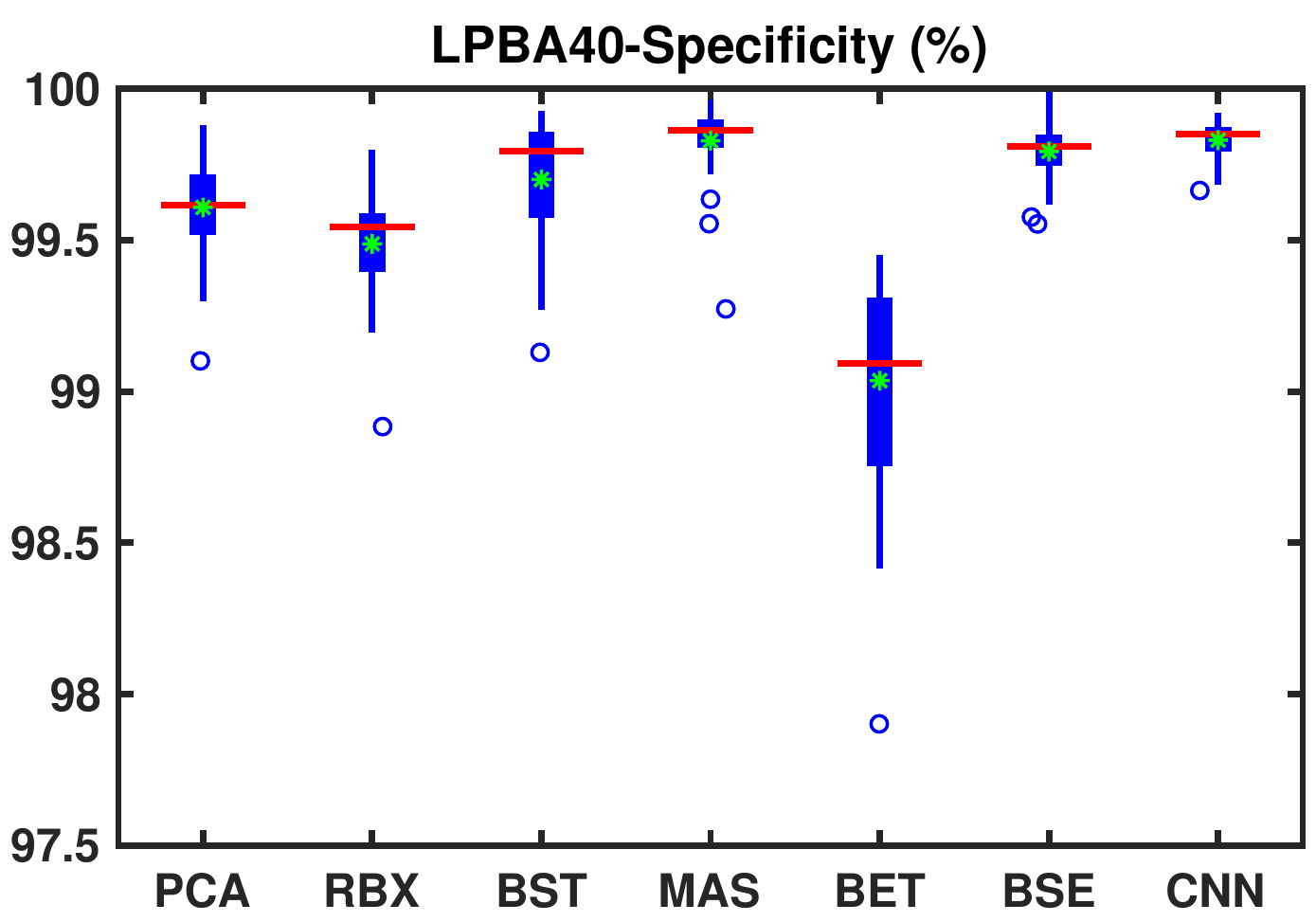}
	\end{tabular}
	\captionof{figure}{Box plot results for the LPBA40 normal dataset. All seven methods work well on this dataset. Our PCA model has the best Dice and surface distances. ROBEX, BEaST, MASS, BET and BSE show similar performance, but BET exhibits larger variance and BSE exhibits two outliers indicating failure. The CNN model shows overall slightly worse performance than the other methods.}
	\label{fig:lpba40_box_plot}
\end{table*}

\vskip0.5ex
\noindent
{\bf IBSR results:} Fig. \ref{fig:ibsr_box_plot} shows the box-plots summarizing the results for the IBSR dataset. Overall, ROBEX, BEaST*, BSE, BET and our model perform well on this dataset, with a median Dice coefficient above 0.95. \xhsp{BEaST does not work well when applied directly on the IBSR images. This is due to failures with the initial spatial normalization (in 5 cases the computations themselves fail and in 10 cases the results are poor). Therefore, in our experiment, we first applied the same affine registration to atlas space as in the pre-processing step for our PCA model for all images. This affine transformation corresponds to a composition of the two affine transformations in Fig.~\ref{fig:preprocessing_flowchart}. BEaST is then applied to the affinely aligned images. We use the same strategy for BRATS. We refer to the resulting approach as BEaST*. BEaST* performs well on most cases with high Dice scores and low surface distances.} MASS works well on some cases, but performs poorly on many cases. CNN does not perform satisfactorily, with low Dice scores, low sensitivity, large distance errors, and overall high variance. \xhsp{Our PCA model has similar performance to BEaST*, but does not result in extreme outliers and hence results in higher mean Dice scores than BEaST*. Both methods outperform all others with respect to Dice scores (median close to 0.97) and distance measures in most cases}. BSE also works well on most cases, but it shows larger variability and exhibits two outliers which represent failure cases. ROBEX and BET show the highest sensitivity, but reduced specificity. Conversely, our PCA model, BEaST*, BSE, and CNN have high specificity but reduced sensitivity (the CNN model dramatically so). 

Table~\ref{tbl:m_std} (top) shows medians, means and standard deviations for the test results on this dataset. Our PCA model achieves the highest median and mean Dice overlap scores (both at 0.97) with the smallest standard deviation. \xhsp{BEaST* also shows high median Dice scores, but results in reduces mean scores due to the presence of outliers.} ROBEX and BET show slightly reduced Dice overlap measures (mean and median around 0.95). BSE also shows slightly reduced median Dice scores, but greatly reduced mean scores. MASS show reduced median Dice scores. CNN shows the lowest performance. Our PCA model also performs best for the surface distance measures; it has the lowest mean and median surfaces distances. Overall our PCA model performs best.

\xhsp{In addition, we perform a one-tailed paired Wilcoxon signed-rank test (to safeguard against deviations from normality) to compare results between methods. We test the null hypothesis that the paired differences for the results of our PCA model and of the compared method come from a distribution with zero median, against the alternative that the median of the paired differences is non-zero.}\footnote{\xhsp{We perform a one-tailed test, thus we test for greater than zero for the Dice overlap scores, sensitivity and specificity, and less than zero for the surface distances.}}. Table \ref{tbl:sign_test_p_value} (top) shows the corresponding results. We apply the Benjamini-Hochberg procedure~\cite{benjamini1995controlling} for all the tests, in order to reduce the false discovery rate for multiple comparisons. We select an overall false discovery rate of 0.05 which results in an effective significance level of $\alpha\approx0.0351$. \chg{Our model outperforms all other methods on Dice and surface distances except for BEaST* which is significant only in Dice and average surface distance. In addition, our approach performs better than MASS, BSE and CNN on sensitivity and better than ROBEX, BEaST*, MASS, and BET on specificity.} 

\vskip1ex
\noindent
{\bf LPBA40 results:} Fig.~\ref{fig:lpba40_box_plot} shows the box-plots summarizing the validation results for the LPBA40 dataset. All seven methods perform well. ROBEX, BEaST, BET and BSE all have a median Dice score between 0.96 and 0.97. MASS has a median Dice score slightly above 0.97. Our PCA model obtains the highest median Dice score (0.974). All methods except for the CNN approach have a median average surface distance smaller than 1~mm. Table~\ref{tbl:m_std} (second top) shows the medians, means and standard deviations for all validation measures for this dataset. Again, all methods have satisfactory median, mean Dice scores and surface distances with low variances. Compared with other methods, the PCA model achieves the best results.

\chg{Table \ref{tbl:sign_test_p_value} (second top) shows the one-sided paired Wilcoxon signed-rank test results. Again we use the Benjamini-Hochberg procedure, resulting in a significance level $\alpha\approx 0.0351$. All methods perform well on this dataset, but our PCA approach still shows statistically significant improvement. We outperform other methods on Dice and all surface distances with statistical significance except for BEaST on maximum surface distance and for MASS on 95\% surface distance. We perform better than all other methods except BET on sensitivity and better than BET and ROBEX on specificity.}

Fig.~\ref{fig:heat_map} (left) visualizes the average brain mask errors for IBSR and LPBA40. All images are first affinely registered to the atlas. Then we transform the gold-standard expert segmentations as well as the automatically obtained brain masks of the different methods to atlas space. We compare the segmentations by counting the average over- and under-segmentation errors over all cases at each voxel. This results in a visualization for areas of likely mis-segmentation. Our PCA model, ROBEX, BEaST (BEaST*) and BET perform well on these two datasets. Compareed to our model, ROBEX, BEaST (BEaST*) and BET show larger localized errors, e.g., at the boundary of the parietal lobe, the occipital lobe and the cerebellum. While MASS, BSE and CNN perform well on the LPBA40 dataset, they perform poorly on the IBSR dataset. This is in particular the case for the CNN approach.

\subsection{Datasets with strong pathologies: BRATS/TBI}

\begin{table*}[!htb]
	\centering
	\begin{tabular}{@{}c@{}c@{}c@{}}
		\includegraphics[width=0.7\columnwidth]{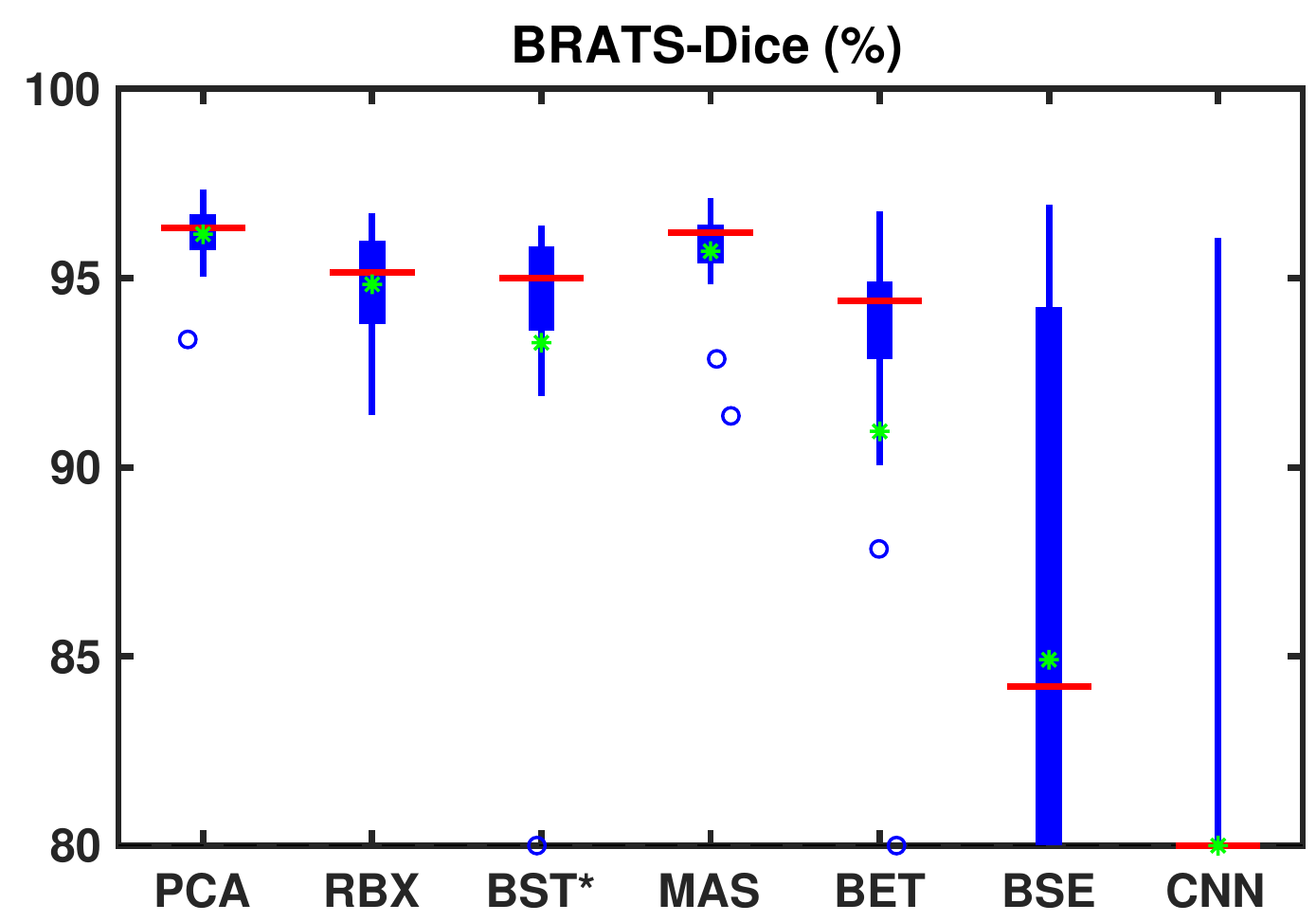} &
		\includegraphics[width=0.7\columnwidth]{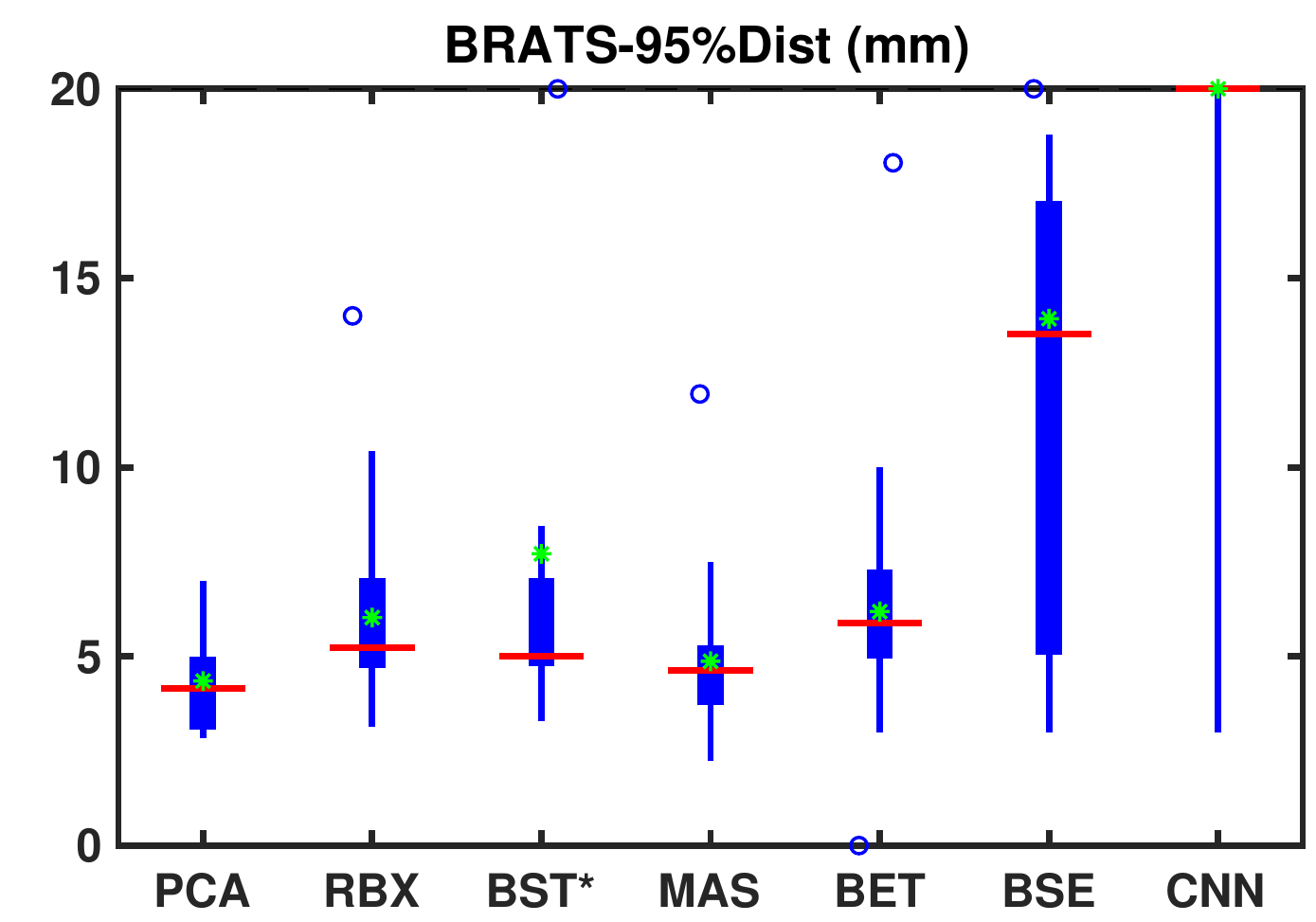} &
		\includegraphics[width=0.7\columnwidth]{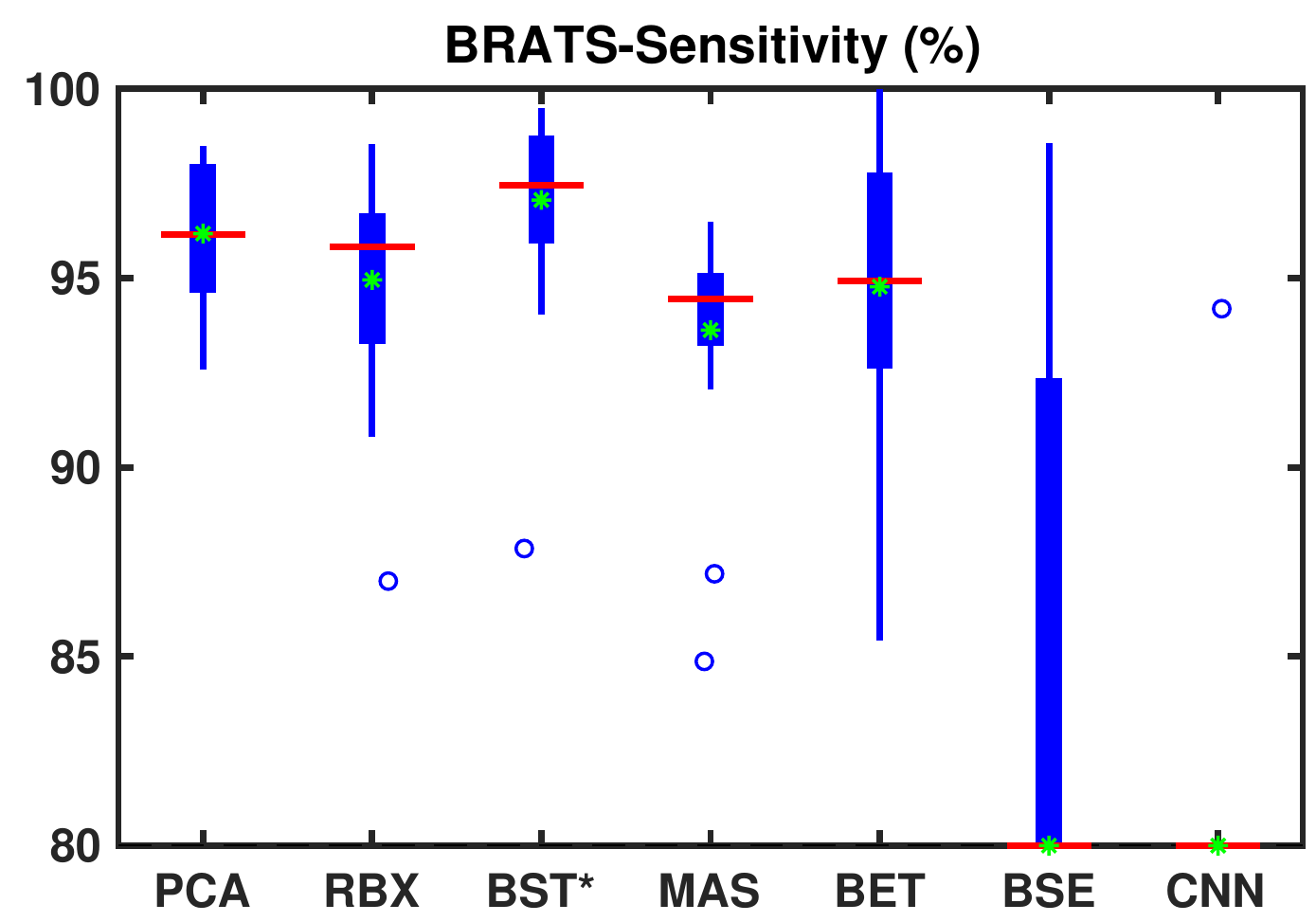} \\
		\includegraphics[width=0.7\columnwidth]{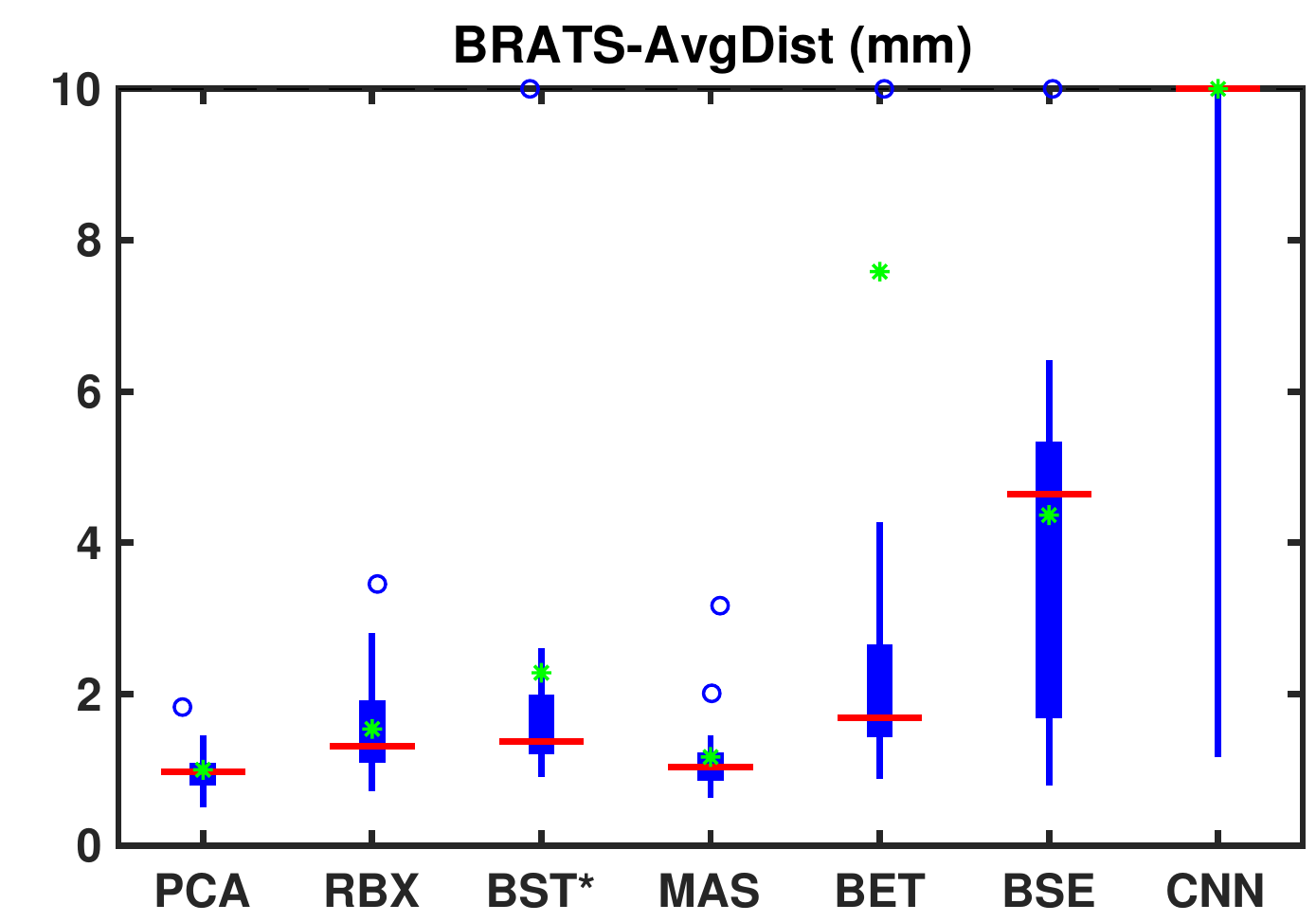} &
		\includegraphics[width=0.7\columnwidth]{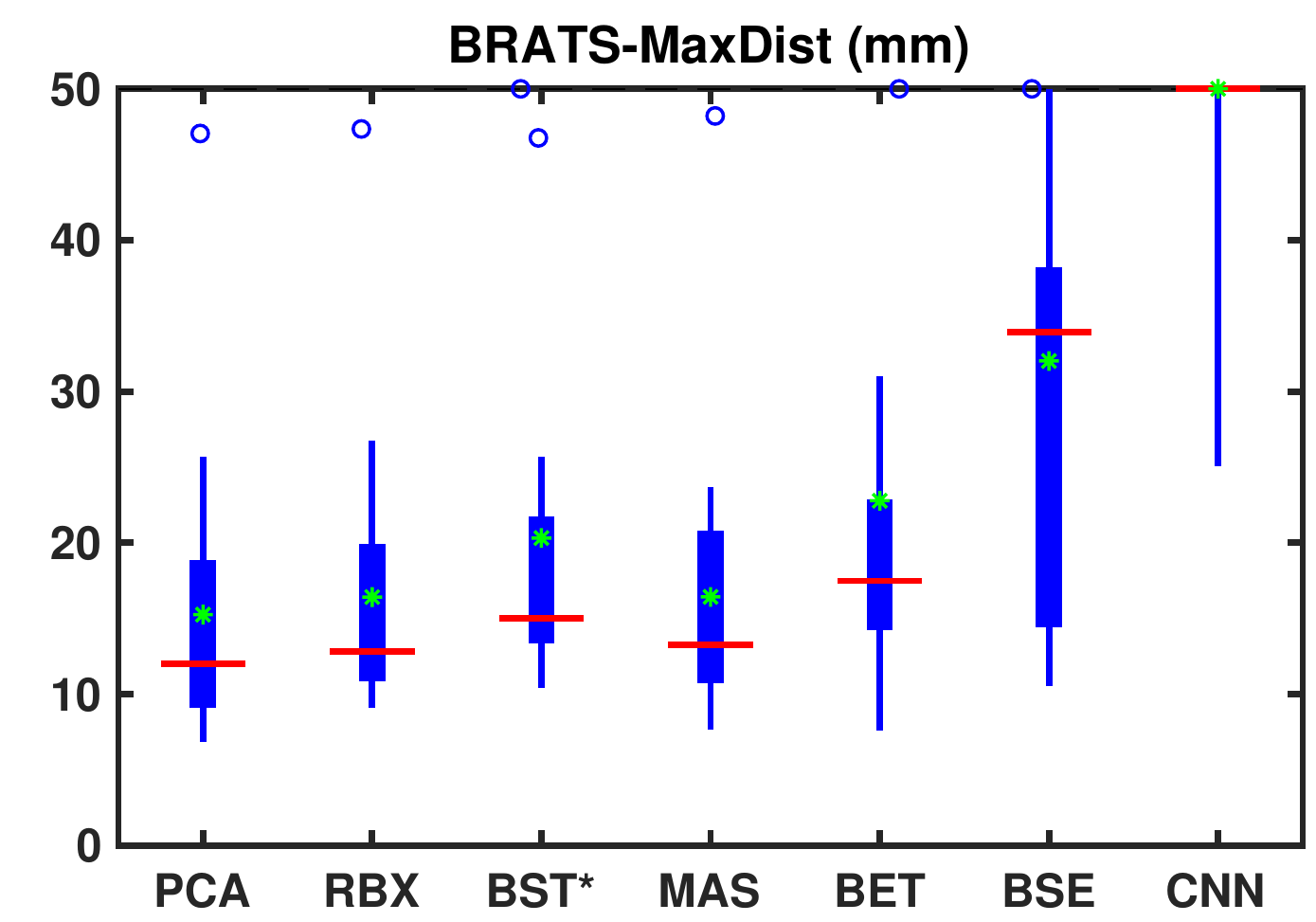} &
		\includegraphics[width=0.7\columnwidth]{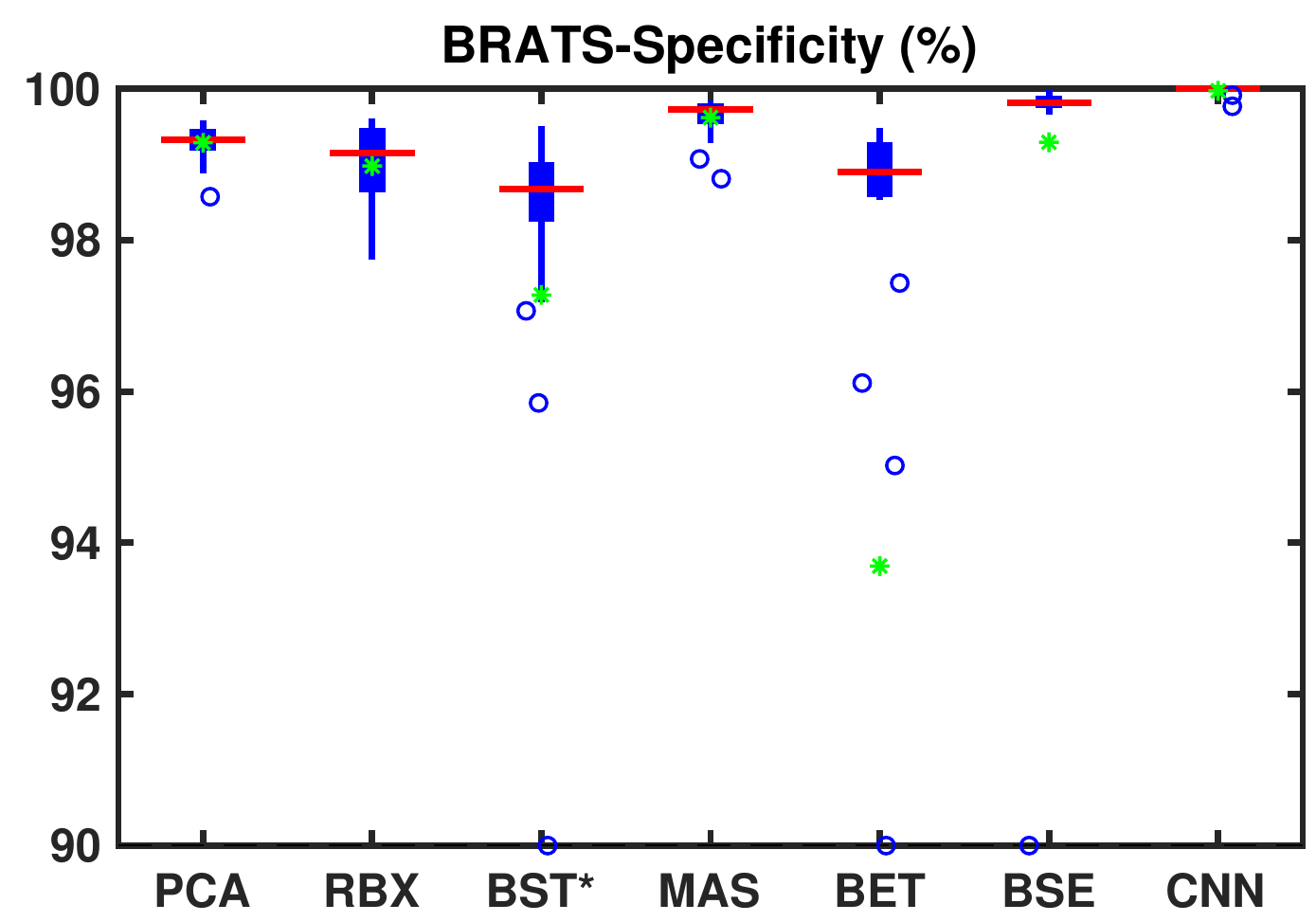}
	\end{tabular}
	\captionof{figure}{Box plot results for the BRATS tumor dataset. BSE and CNN fail on this dataset. \xhsp{BEaST also fails when applied directly to the BRATS dataset due to spatial normalization failures. We therefore show results for BEaST* here, our modification which uses the affine registration of the PCA model first.} BET shows better performance, but also exhibits outliers. ROBEX, BEaST*, MASS, and our PCA model work well on this dataset. Overall our model exhibits the best performance scores.}
	\label{fig:brats_box_plot}
	\centering
	\begin{tabular}{@{}c@{}c@{}c@{}}
		\includegraphics[width=0.7\columnwidth]{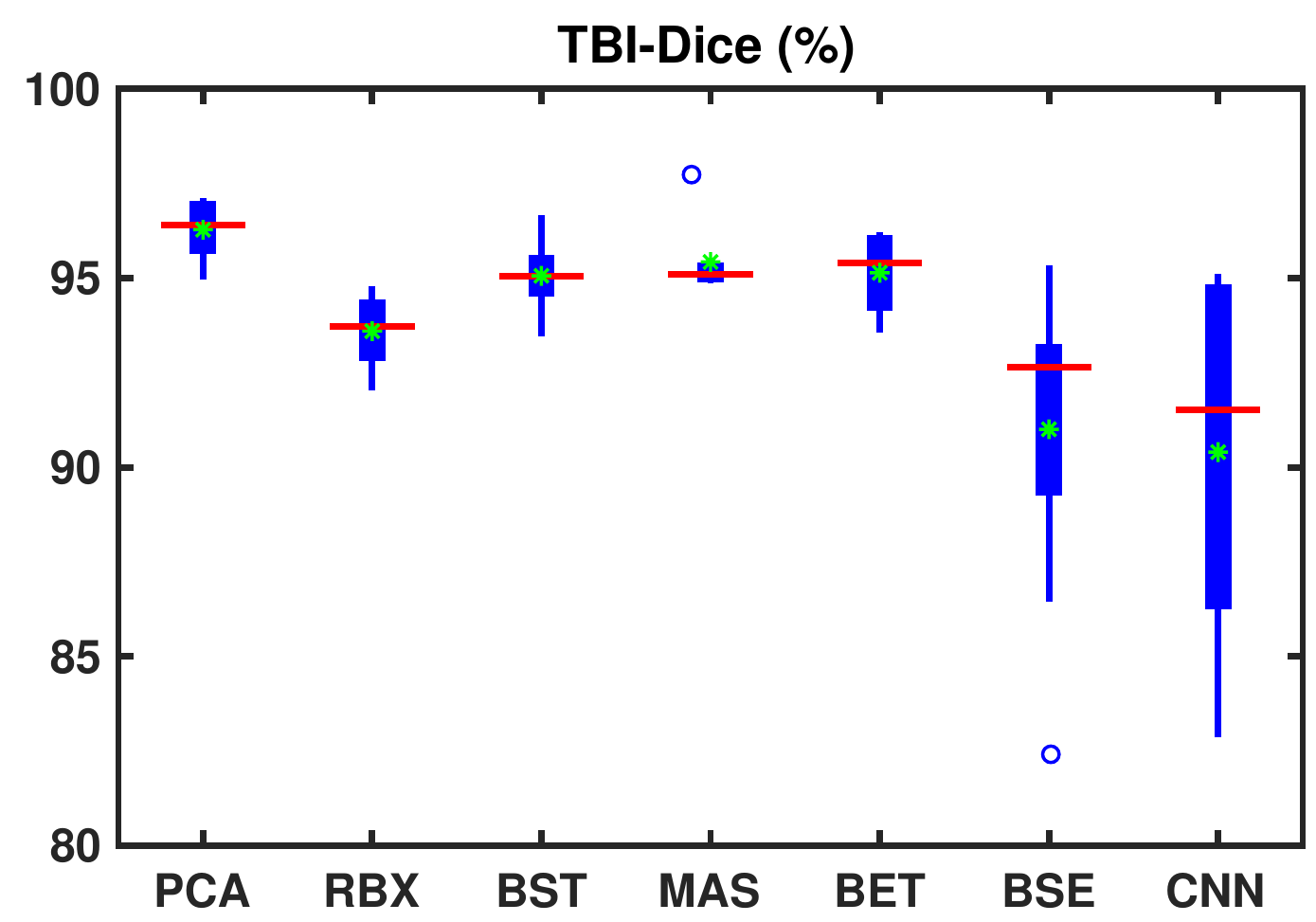} &
		\includegraphics[width=0.7\columnwidth]{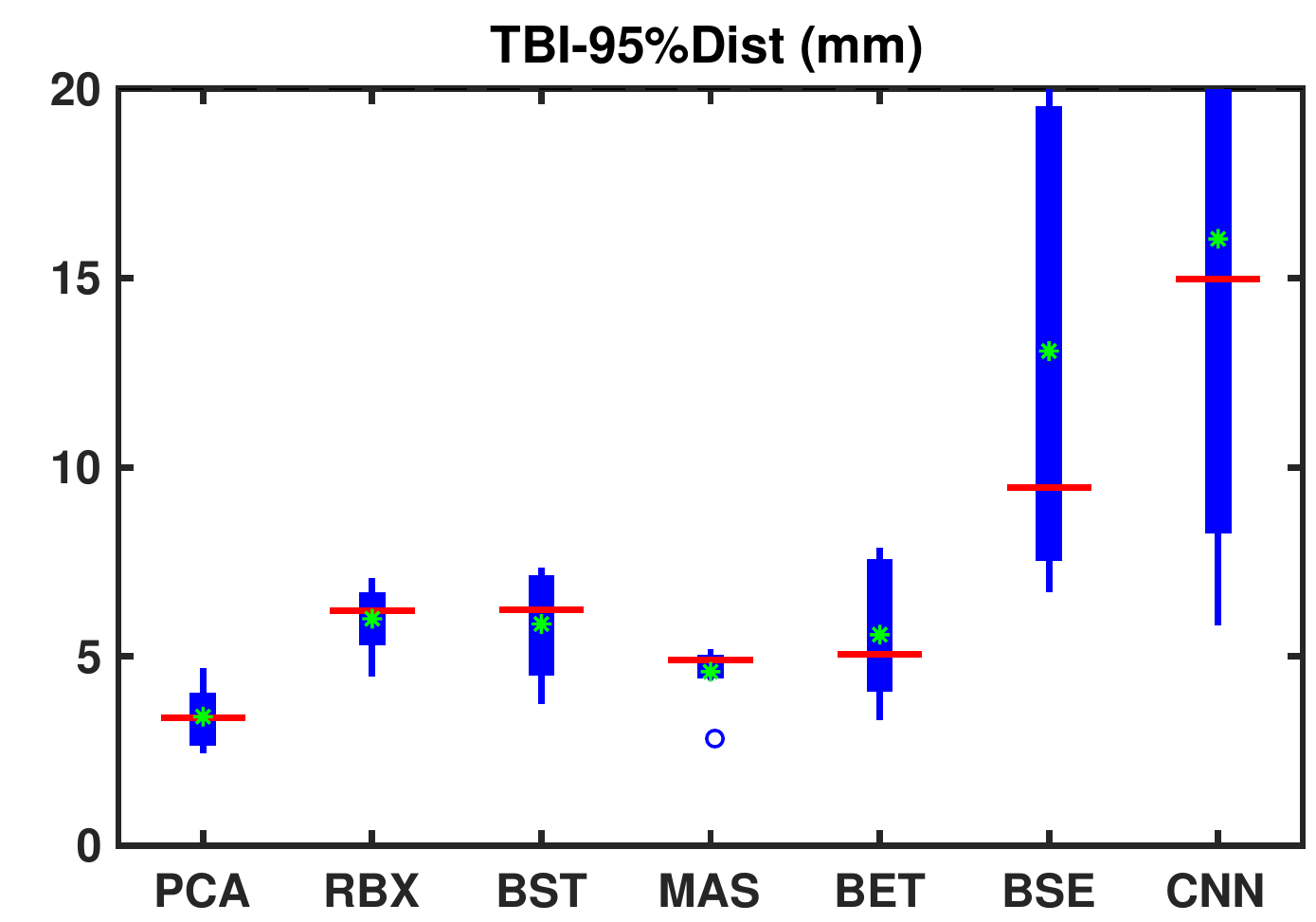} &
		\includegraphics[width=0.7\columnwidth]{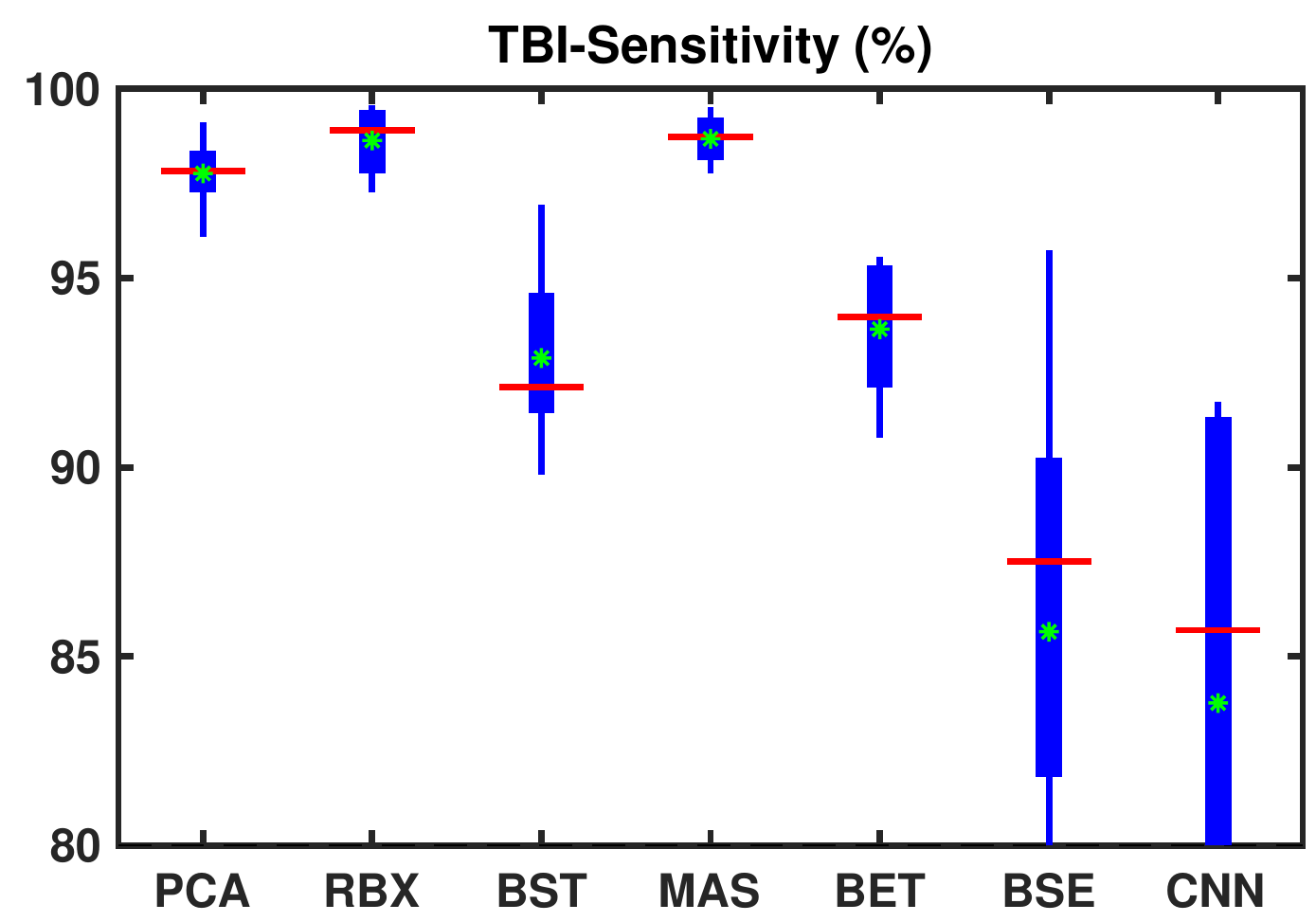} \\
		\includegraphics[width=0.7\columnwidth]{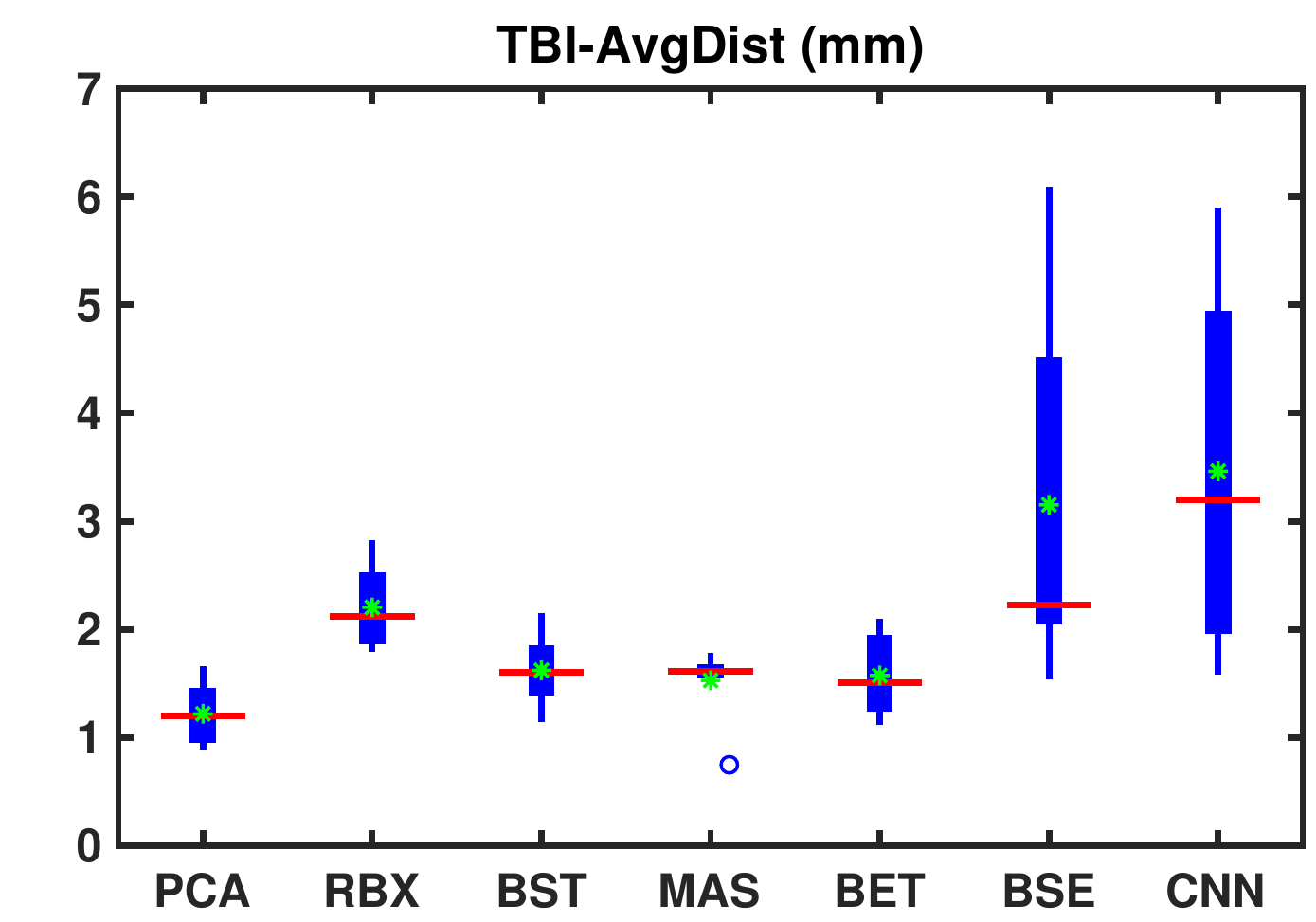} &
		\includegraphics[width=0.7\columnwidth]{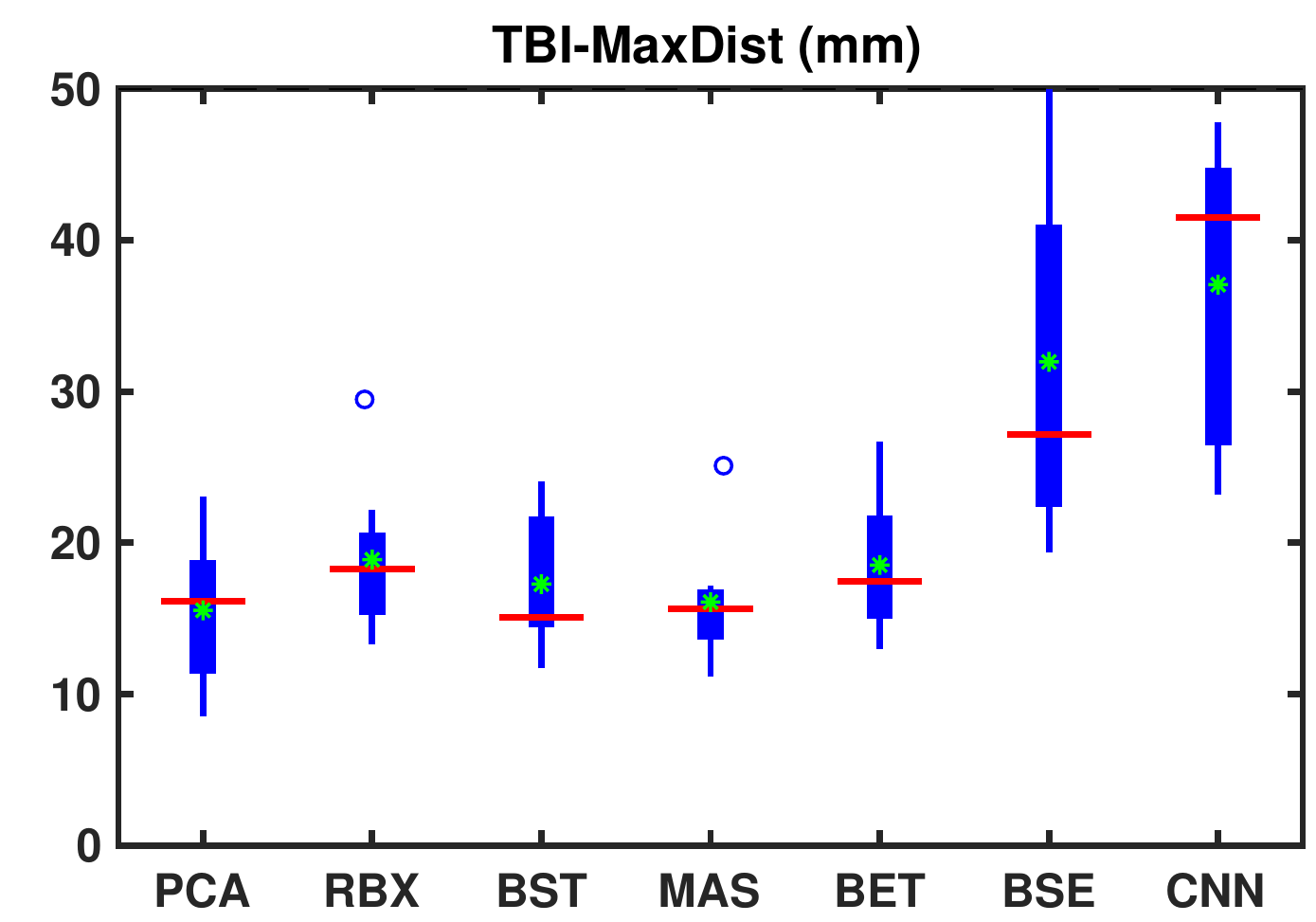} &
		\includegraphics[width=0.7\columnwidth]{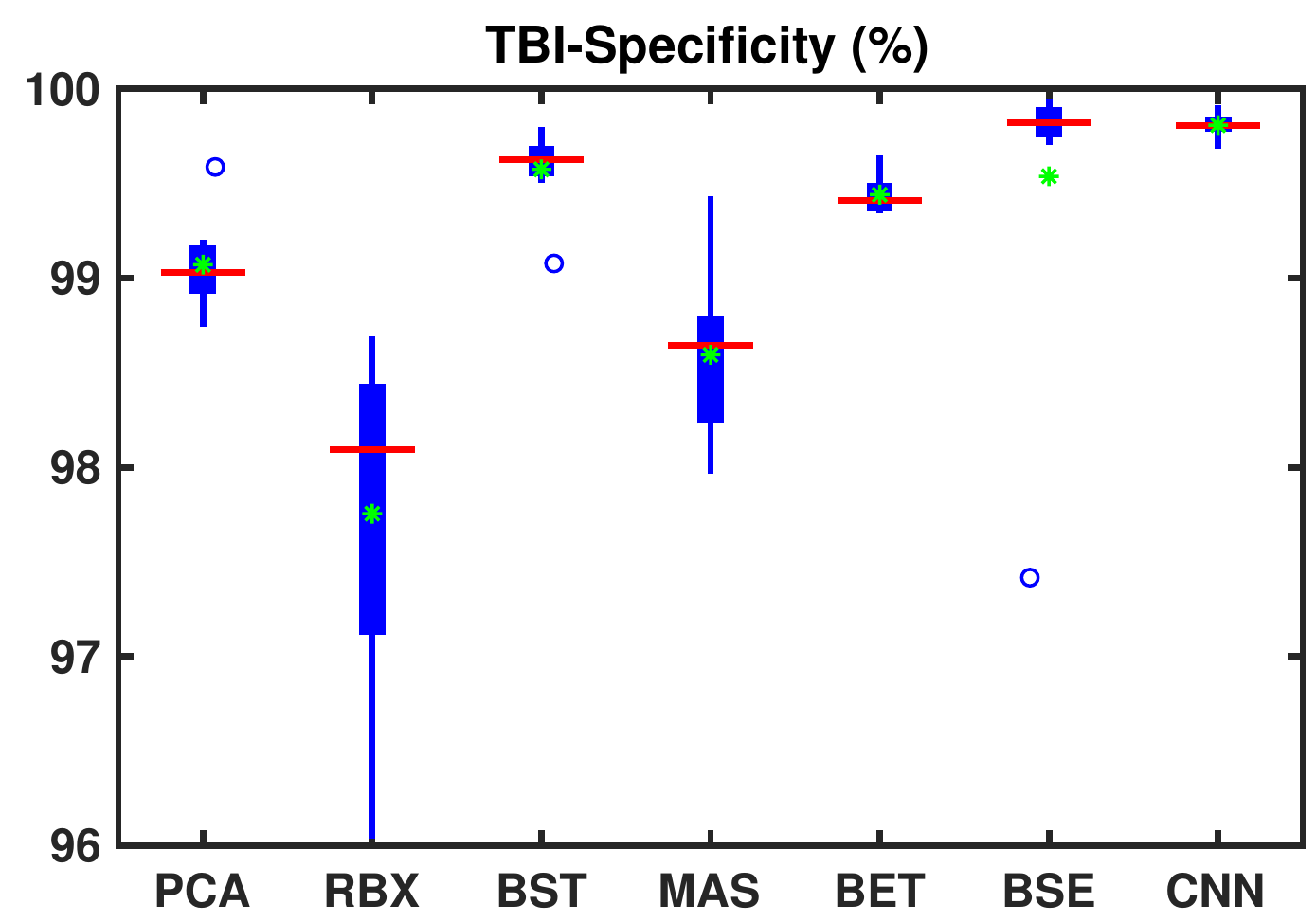}

	\end{tabular}
	\captionof{figure}{Box plot results for the TBI dataset. Our PCA model shows the best evaluation scores. BET, BEaST, MASS and ROBEX also perform reasonably well. BSE and CNN exhibit inferior performance on this dataset.}
	\label{fig:tbi_box_plot}
\end{table*}

\vskip0.5ex
\noindent
{\bf BRATS results:} Fig.~\ref{fig:brats_box_plot} shows the box-plots for the validation measures for the BRATS dataset. BSE and CNN, using their default settings, do not work well on the BRATS dataset. This may be because of the data quality of the BRATS data. Many of the BRATS images have relatively low out-of-plane resolutions. BSE results may be improved by a better parameter setting. However, as our goal is to evaluate all methods with the same parameter setting across all datasets, we do not explore dataset specific parameter tuning. \xhsp{BEaST also fails on the original BRATS images due to the spatial normalization. As for the IBSR dataset, we therefore use BEaST*, our adaptation of BEaST using the affine transformation of our PCA model.} BET shows good performance, but suffers from a few outliers. ROBEX and BEaST* work generally well, with a median Dice score around 0.95 and an average distance error of 1.3 mm. MASS also works well on most cases. However, as for IBSR and LPBA40, our PCA model performs generally the best with a median Dice score 0.96 and a 1 mm average distance error. The PCA model results also show lower variance, as shown in table ~\ref{tbl:m_std} (second bottom), underlining the very consistent behavior of our approach. 

\chg{Table ~\ref{tbl:sign_test_p_value} shows (via a one-sided paired Wilcoxon signed-rank test with a correction for multiple comparisons using a false discovery rate of $0.05$) that our model has statistically significantly better performance than ROBEX, BEaST*, BET, BSE, CNN on most measures. The improvement over MASS, however, is not statistically significant.}

\begin{table}[htb]
	\centering
	\begin{tabular}{@{}c@{}c@{}c@{}c@{}}
        		\includegraphics[width=0.25\columnwidth]{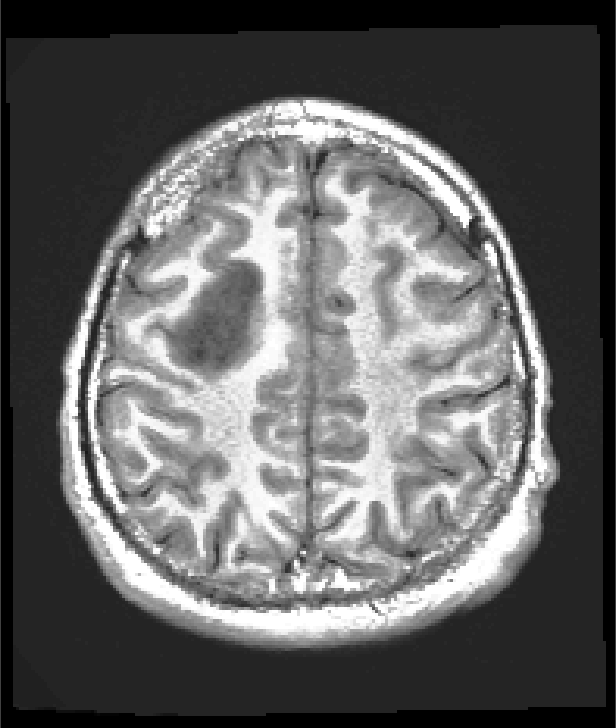} 	&
		\includegraphics[width=0.25\columnwidth]{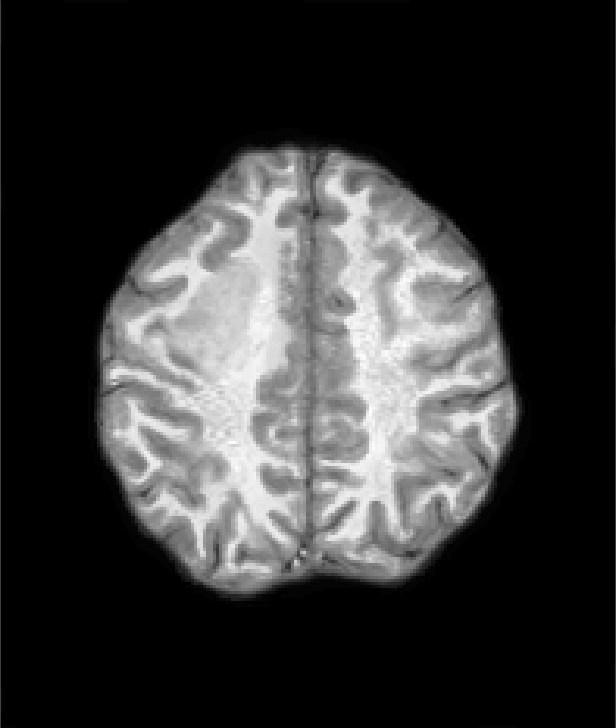} 	&
		\includegraphics[width=0.25\columnwidth]{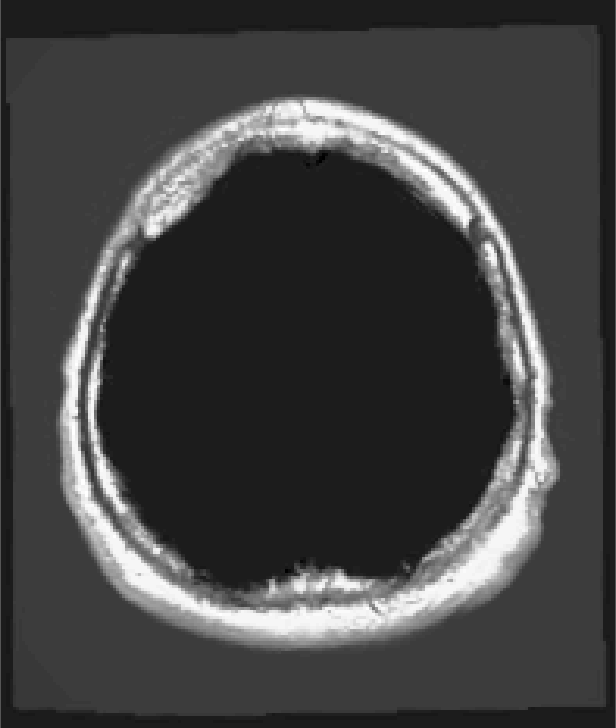} 	&
		\includegraphics[width=0.25\columnwidth]{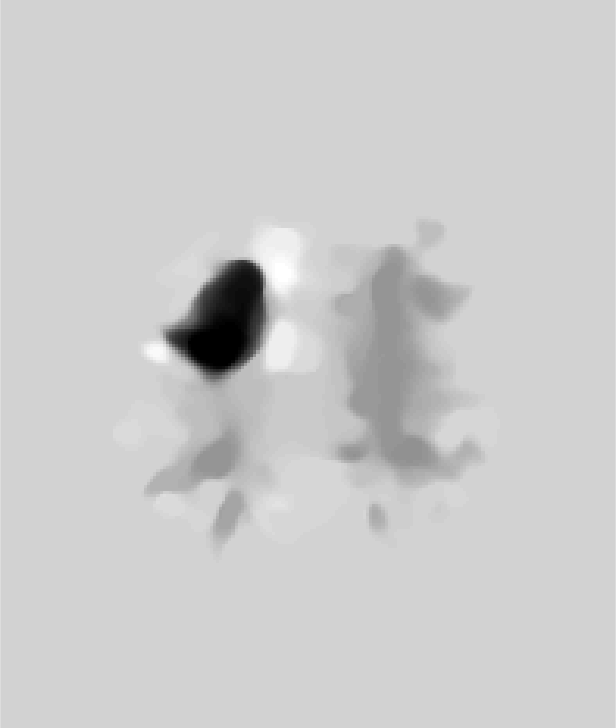}\\
		\small(a)&\small(b)&\small(c)&\small(d)\\[-3mm]
	\end{tabular}
	\captionof{figure}{Example BRATS image with its decomposition result in atlas space. (a) Input image after pre-processing; (b) quasi-normal image $L+M$; (c) non-brain image $S$; (d) pathology image $T$.}
	\label{fig:brats_result_1}
\end{table}

\vskip0.5ex
\noindent
{\bf TBI results:} Fig.~\ref{fig:tbi_box_plot} shows the box-plots for the results on our TBI dataset. Our PCA model still outperforms all other methods. Our method achieves the largest Dice scores, and the lowest surface distances among all methods with best mean and lowest variance as shown in table ~\ref{tbl:m_std} (bottom). \chg{Table \ref{tbl:sign_test_p_value} shows the one-sided paired Wilcoxon signed-rank test results with multiple comparisons correction (with a false discovery rate of $0.05$). Our model performs significantly better than ROBEX, BEaST, BET, BSE and CNN on most measures. The improvement over MASS is only statistically significant on Dice and 95\% surface distance.}

Finally, Fig.~\ref{fig:heat_map} (right) shows the average segmentation errors on the BRATS and TBI datasets: our PCA method shows fewer errors than most other methods in these two abnormal datasets. MASS also shows few errors, while ROBEX, BEaST (BEaST*) and BET exhibit slightly larger errors at the boundary of the brain. CNN and BSE particularly show large errors for the BRATS dataset presumably again due to the coarse resolution of the BRATS data. 

In addition to extracting the brain from pathological datasets, our method also allows for the estimation of a corresponding quasi-normal image in atlas space, although this is not the main goal of this paper. Fig.\ref{fig:brats_result_1} shows an example of the reconstructed quasi-normal image ($L$) for an image of the BRATS dataset, as well as an estimation of the pathology (pathology image $T$ and non-brain image $S$). Compared to the original image, the pathology shown in the quasi-normal image has been greatly reduced. Hence this image can be used for the registration with a normal image or a normal atlas. This has been shown to improve registration accuracy for the registration of pathological images~\cite{han2017efficient}. Furthermore, an estimate of the pathology (here a tumor) is also obtained which may be useful for further analysis. Note that in this example image the total variation term captures more than just the tumor. This may be due to inconsistencies in the image appearance between the normal images (obtained from OASIS data) and the test dataset. As our goal is atlas alignment rather than quasi-normal image reconstruction or pathology segmentation, such a decomposition is acceptable, although we could improve this by tuning the parameters or applying regularization steps as in \cite{han2017efficient}.

\begin{table*}[t]
	\centering
	\begin{tabular}{|c|c|c|c|c|c|c|}
		\hline
		\multicolumn{7}{|c|}{\textbf{Dataset: IBSR}}\\
		\hline
		&ROBEX &BEaST* &MASS &BET & BSE & CNN\\
		\hline
		Dice&\cellcolor{mybetter}4.78e-5&\cellcolor{mybetter}1.20e-2&\cellcolor{mybetter}2.77e-4&\cellcolor{mybetter}4.78e-5&\cellcolor{mybetter}7.55e-5&\cellcolor{mybetter}4.78e-5\\
		\hline
		Avg Dist&\cellcolor{mybetter}4.78e-5&\cellcolor{mybetter}2.73e-2&\cellcolor{mybetter}1.82e-4&\cellcolor{mybetter}4.78e-5&\cellcolor{mybetter}4.78e-5&\cellcolor{mybetter}4.78e-5\\
		\hline
		95\% Dist&\cellcolor{mybetter}4.74e-5&5.91e-2&\cellcolor{mybetter}1.05e-4&\cellcolor{mybetter}4.71e-5&\cellcolor{mybetter}4.74e-5&\cellcolor{mybetter}4.78e-5\\
		\hline
		Max Dist&\cellcolor{mybetter}4.78e-5&5.36e-2&\cellcolor{mybetter}4.78e-5&\cellcolor{mybetter}4.78e-5&\cellcolor{mybetter}1.58e-4&\cellcolor{mybetter}5.58e-5\\
		\hline
		Sensitivity&0.994&0.448&\cellcolor{mybetter}3.40e-3&0.829&\cellcolor{mybetter}4.78e-5&\cellcolor{mybetter}4.78e-5\\
		\hline
		Specificity&\cellcolor{mybetter}5.58e-5&\cellcolor{mybetter}2.97e-2&\cellcolor{mybetter}2.41e-3&\cellcolor{mybetter}4.78e-5&0.894&1.000\\
		\hline
		\noalign{\vskip 2em}
		\hline
		\multicolumn{7}{|c|}{\textbf{Dataset: LPBA40}}\\
		\hline
		&ROBEX &BEaST &MASS &BET & BSE & CNN\\
		\hline
		Dice&\cellcolor{mybetter}1.47e-7&\cellcolor{mybetter}2.51e-8&\cellcolor{mybetter}1.89e-3&\cellcolor{mybetter}9.58e-5&\cellcolor{mybetter}2.24e-7&\cellcolor{mybetter}1.85e-8\\
		\hline
		Avg Dist&\cellcolor{mybetter}1.36e-7&\cellcolor{mybetter}2.51e-8&\cellcolor{mybetter}2.75e-3&\cellcolor{mybetter}1.60e-6&\cellcolor{mybetter}6.31e-7&\cellcolor{mybetter}1.85e-8\\
		\hline
		95\% Dist&\cellcolor{mybetter}2.90e-8&\cellcolor{mybetter}3.29e-7&5.69e-2&\cellcolor{mybetter}2.71e-8&\cellcolor{mybetter}1.02e-5&\cellcolor{mybetter}1.85e-8\\
		\hline
		Max Dist&\cellcolor{mybetter}2.16e-8&1.000&\cellcolor{mybetter}2.58e-2&\cellcolor{mybetter}2.92e-8&\cellcolor{mybetter}3.01e-5&\cellcolor{mybetter}2.51e-8\\
		\hline
		Sensitivity&\cellcolor{mybetter}4.13e-3&\cellcolor{mybetter}1.27e-7&\cellcolor{mybetter}1.60e-6&1.000&\cellcolor{mybetter}6.14e-8&\cellcolor{mybetter}1.85e-8\\
		\hline
		Specificity&\cellcolor{mybetter}5.70e-6&0.998&1.000&\cellcolor{mybetter}2.00e-8&1.000&1.000\\
		\hline
		\noalign{\vskip 2em}
		\hline
		\multicolumn{7}{|c|}{\textbf{Dataset: BRATS}}\\
		\hline
		&ROBEX &BEaST* &MASS &BET & BSE & CNN\\
		\hline
		Dice&\cellcolor{mybetter}1.58e-4&\cellcolor{mybetter}3.18e-4&7.02e-2&\cellcolor{mybetter}4.78e-5&\cellcolor{mybetter}4.78e-5&\cellcolor{mybetter}4.78e-5\\
		\hline
		Avg Dist&\cellcolor{mybetter}1.36e-4&\cellcolor{mybetter}2.77e-4&9.89e-2&\cellcolor{mybetter}4.78e-5&\cellcolor{mybetter}4.78e-5&\cellcolor{mybetter}4.78e-5\\
		\hline
		95\% Dist&\cellcolor{mybetter}8.41e-5&\cellcolor{mybetter}4.17e-4&0.266&\cellcolor{mybetter}1.53e-3&\cellcolor{mybetter}4.78e-5&\cellcolor{mybetter}7.15e-5\\
		\hline
		Max Dist&\cellcolor{mybetter}1.91e-2&\cellcolor{mybetter}7.38e-4&0.222&\cellcolor{mybetter}2.41e-4&\cellcolor{mybetter}1.18e-4&\cellcolor{mybetter}4.78e-5\\
		\hline
		Sensitivity&\cellcolor{mybetter}3.51e-2&0.981&\cellcolor{mybetter}2.09e-4&8.08e-2&\cellcolor{mybetter}5.58e-5&\cellcolor{mybetter}4.78e-5\\
		\hline
		Specificity&6.53e-2&\cellcolor{mybetter}1.82e-4&0.999&\cellcolor{mybetter}4.73e-3&0.999&1.000\\
		\hline
		\noalign{\vskip 2em}
		\hline
		\multicolumn{7}{|c|}{\textbf{Dataset: TBI}}\\
		\hline
		&ROBEX &BEaST &MASS &BET & BSE & CNN\\
		\hline
		Dice&\cellcolor{mybetter}3.91e-3&\cellcolor{mybetter}1.95e-2&\cellcolor{mybetter}2.73e-2&\cellcolor{mybetter}7.81e-3&\cellcolor{mybetter}3.91e-3&\cellcolor{mybetter}3.91e-3\\
		\hline
		Avg Dist&\cellcolor{mybetter}3.91e-3&\cellcolor{mybetter}1.95e-2&3.91e-2&\cellcolor{mybetter}7.81e-3&\cellcolor{mybetter}3.91e-3&\cellcolor{mybetter}3.91e-3\\
		\hline
		95\% Dist&\cellcolor{mybetter}3.91e-3&\cellcolor{mybetter}7.81e-3&\cellcolor{mybetter}7.81e-3&\cellcolor{mybetter}3.91e-3&\cellcolor{mybetter}3.91e-3&\cellcolor{mybetter}3.91e-3\\
		\hline
		Max Dist&\cellcolor{mybetter}1.17e-2&9.77e-2&0.344&5.47e-2&\cellcolor{mybetter}3.91e-3&\cellcolor{mybetter}3.91e-3\\
		\hline
		Sensitivity&0.980&\cellcolor{mybetter}3.91e-3&0.961&\cellcolor{mybetter}3.91e-3&\cellcolor{mybetter}3.91e-3&\cellcolor{mybetter}3.91e-3\\
		\hline
		Specificity&\cellcolor{mybetter}3.91e-3&1.000&\cellcolor{mybetter}2.73e-2&1.000&0.926&1.000\\
		\hline
	\end{tabular}
	\caption{$p$-values for all datasets, computed by the signed-rank test. \xhsp{We perform a one-tailed paired Wilcoxon signed-rank test, where the null-hypothesis ($\mathcal{H}_0$) is that the paired differences for the results of our PCA model and of the compared method come from a distribution with zero median, against the alternative ($\mathcal{H}_1$) that the paired differences have a non-zero median (greater than zero for Dice, sensitivity and specificity, and less than zero for surface distances)}. In addition, we use the Benjamini-Hochberg procedure to reduce the false discovery rate (FDR). We highlight, in \textcolor{green}{green}, the results where our PCA model performs statistically significantly better. The results show that our PCA model outperforms other methods on most of the measures.}
	\label{tbl:sign_test_p_value}
\end{table*}

\begin{table*}[t]
	\footnotesize
	\centering
	\begin{tabular}{|c|c|c|c|c|c|c|c|}
		\hline
		\multicolumn{8}{|c|}{\textbf{Dataset: IBSR}}\\
		\hline
		&PCA &ROBEX &BEaST* &MASS &BET & BSE & CNN\\
		\hline
		Dice(\%)
		&\cellcolor{mybetter}\shortstack{\\97.07\\96.99$\pm$0.53}
		&\shortstack{\\95.09\\94.98$\pm$1.17}
		&\shortstack{\\96.94\\95.07$\pm$7.50}
		&\shortstack{\\92.23\\86.76$\pm$11.06}
		&\shortstack{\\95.66\\95.16$\pm$0.96}
		&\shortstack{\\95.68\\89.54$\pm$21.76}
		&\shortstack{\\84.50\\81.10$\pm$12.07}
		\\
		\hline
		Avg Dist(mm) 
		&\cellcolor{mybetter}\shortstack{\\0.71\\0.79$\pm$0.27}
		&\shortstack{\\1.52\\1.51$\pm$0.56}
		&\shortstack{\\0.74\\1.48$\pm$2.76}
		&\shortstack{\\2.50\\4.84$\pm$4.54}
		&\shortstack{\\1.31\\1.49$\pm$0.47}
		&\shortstack{\\1.18\\4.16$\pm$10.53}
		&\shortstack{\\4.62\\5.59$\pm$3.10}
		\\
		\hline
		95\% Dist(mm) 
		&\cellcolor{mybetter}\shortstack{\\2.83\\2.84$\pm$0.43}
		&\shortstack{\\4.18\\4.50$\pm$1.58}
		&\shortstack{\\3.00\\5.19$\pm$9.79}
		&\shortstack{\\8.73\\13.66$\pm$11.81}
		&\shortstack{\\3.61\\4.22$\pm$1.39}
		&\shortstack{\\5.00\\12.27$\pm$20.83}
		&\shortstack{\\20.05\\22.25$\pm$9.41}
		\\
		\hline
		Max Dist(mm) 
		&\cellcolor{mybetter}\shortstack{\\9.30\\11.97$\pm$8.14}
		&\shortstack{\\15.55\\17.30$\pm$8.40}
		&\shortstack{\\11.74\\15.60$\pm$13.41}
		&\shortstack{\\24.20\\25.76$\pm$12.94}
		&\shortstack{\\15.68\\17.91$\pm$7.85}
		&\shortstack{\\16.57\\24.93$\pm$23.32}
		&\shortstack{\\39.10\\41.10$\pm$8.14}
		\\
		\hline
		Sensitivity(\%)
		&\shortstack{\\99.06\\98.99$\pm$0.46}
		&\shortstack{\\99.47\\99.33$\pm$0.54}
		&\shortstack{\\98.98\\96.70$\pm$10.12}
		&\shortstack{\\98.77\\98.52$\pm$0.77}
		&\cellcolor{mybetter}\shortstack{\\99.57\\99.09$\pm$0.93}
		&\shortstack{\\97.08\\88.68$\pm$22.86}
		&\shortstack{\\74.87\\70.96$\pm$15.89}
		\\
		\hline
		Specificity(\%)
		&\shortstack{\\99.48\\99.44$\pm$0.21}
		&\shortstack{\\98.89\\98.90$\pm$0.51}
		&\shortstack{\\99.48\\99.32$\pm$0.37}
		&\shortstack{\\97.88\\96.11$\pm$3.81}
		&\shortstack{\\99.12\\98.98$\pm$0.46}
		&\shortstack{\\99.58\\99.15$\pm$1.67}
		&\cellcolor{mybetter}\shortstack{\\99.85\\99.80$\pm$0.19}
		\\
		\hline
		\multicolumn{8}{|c|}{\textbf{Dataset: LPBA40}}\\
		\hline
		&PCA &ROBEX &BEaST &MASS &BET & BSE & CNN\\
		\hline
		Dice(\%) 
		&\cellcolor{mybetter}\shortstack{\\97.41\\97.32$\pm$0.42}
		&\shortstack{\\96.74\\96.74$\pm$0.24}
		&\shortstack{\\96.80\\94.25$\pm$15.29}
		&\shortstack{\\97.08\\97.03$\pm$0.57}
		&\shortstack{\\96.92\\96.70$\pm$0.78}
		&\shortstack{\\96.77\\96.29$\pm$2.26}
		&\shortstack{\\95.80\\95.70$\pm$0.74}
		\\
		\hline
		Avg Dist(mm) 
		&\cellcolor{mybetter}\shortstack{\\0.76\\0.79$\pm$0.12}
		&\shortstack{\\0.97\\0.97$\pm$0.07}
		&\shortstack{\\0.97\\3.36$\pm$14.83}
		&\shortstack{\\0.85\\0.88$\pm$0.21}
		&\shortstack{\\0.97\\1.06$\pm$0.27}
		&\shortstack{\\0.92\\1.11$\pm$0.81}
		&\shortstack{\\1.34\\1.39$\pm$0.37}
		\\
		\hline
		95\% Dist(mm) 
		&\cellcolor{mybetter}\shortstack{\\2.28\\2.27$\pm$0.32}
		&\shortstack{\\2.98\\2.97$\pm$0.26}
		&\shortstack{\\2.86\\6.45$\pm$23.19}
		&\shortstack{\\2.36\\2.50$\pm$0.97}
		&\shortstack{\\3.46\\3.92$\pm$1.24}
		&\shortstack{\\2.62\\3.46$\pm$3.38}
		&\shortstack{\\3.11\\3.56$\pm$1.56}
		\\
		\hline
		Max Dist(mm) 
		&\shortstack{\\9.73\\10.83$\pm$3.76}
		&\shortstack{\\12.89\\13.81$\pm$3.47}
		&\cellcolor{mybetter}\shortstack{\\8.80\\12.89$\pm$24.44}
		&\shortstack{\\10.71\\11.53$\pm$4.04}
		&\shortstack{\\14.44\\15.14$\pm$3.75}
		&\shortstack{\\13.61\\15.54$\pm$7.74}
		&\shortstack{\\16.91\\19.55$\pm$8.17}
		\\
		\hline
		Sensitivity(\%)
		&\shortstack{\\97.10\\96.81$\pm$1.23}
		&\shortstack{\\96.15\\96.33$\pm$0.85}
		&\shortstack{\\94.76\\92.70$\pm$15.12}
		&\shortstack{\\95.05\\95.15$\pm$1.08}
		&\cellcolor{mybetter}\shortstack{\\98.70\\98.66$\pm$0.54}
		&\shortstack{\\94.85\\94.02$\pm$4.10}
		&\shortstack{\\92.79\\92.62$\pm$1.46}
		\\
		\hline
		Specificity(\%)
		&\shortstack{\\99.62\\99.61$\pm$0.16}
		&\shortstack{\\99.54\\99.49$\pm$0.16}
		&\shortstack{\\99.80\\99.70$\pm$0.21}
		&\cellcolor{mybetter}\shortstack{\\99.86\\99.83$\pm$0.12}
		&\shortstack{\\99.09\\99.04$\pm$0.34}
		&\shortstack{\\99.81\\99.79$\pm$0.09}
		&\shortstack{\\99.85\\99.83$\pm$0.07}
		\\
		\hline
		\multicolumn{8}{|c|}{\textbf{Dataset: BRATS}}\\
		\hline
		&PCA &ROBEX &BEaST* &MASS &BET & BSE & CNN\\
		\hline
		Dice(\%) 
		&\cellcolor{mybetter}\shortstack{\\96.34\\96.16$\pm$0.92}
		&\shortstack{\\95.15\\94.83$\pm$1.49}
		&\shortstack{\\94.99\\93.29$\pm$7.00}
		&\shortstack{\\96.20\\95.71$\pm$1.39}
		&\shortstack{\\94.40\\90.95$\pm$13.41}
		&\shortstack{\\84.21\\84.91$\pm$8.89}
		&\shortstack{\\1.75\\21.89$\pm$29.54}
		\\
		\hline
		Avg Dist(mm) 
		&\cellcolor{mybetter}\shortstack{\\0.97\\1.00$\pm$0.31}
		&\shortstack{\\1.31\\1.54$\pm$0.70}
		&\shortstack{\\1.38\\2.28$\pm$3.34}
		&\shortstack{\\1.03\\1.17$\pm$0.56}
		&\shortstack{\\1.68\\7.58$\pm$25.30}
		&\shortstack{\\4.65\\4.37$\pm$3.61}
		&\shortstack{\\55.88\\44.87$\pm$29.05}
		\\
		\hline
		95\% Dist(mm) 
		&\cellcolor{mybetter}\shortstack{\\4.15\\4.35$\pm$1.27}
		&\shortstack{\\5.23\\6.03$\pm$2.50}
		&\shortstack{\\5.01\\7.71$\pm$9.29}
		&\shortstack{\\4.62\\4.87$\pm$2.04}
		&\shortstack{\\5.87\\6.18$\pm$3.53}
		&\shortstack{\\13.52\\13.92$\pm$13.00}
		&\shortstack{\\78.45\\73.85$\pm$38.77}
		\\
		\hline
		Max Dist(mm)
		&\cellcolor{mybetter}\shortstack{\\12.03\\15.26$\pm$9.32}
		&\shortstack{\\12.83\\16.42$\pm$8.80}
		&\shortstack{\\15.03\\20.32$\pm$14.57}
		&\shortstack{\\13.29\\16.43$\pm$8.97}
		&\shortstack{\\17.49\\22.78$\pm$22.61}
		&\shortstack{\\33.95\\32.02$\pm$22.38}
		&\shortstack{\\87.73\\86.60$\pm$36.92}
		\\
		\hline
		Sensitivity(\%)
		&\shortstack{\\96.16\\96.17$\pm$1.84}
		&\shortstack{\\95.82\\94.95$\pm$2.88}
		&\cellcolor{mybetter}\shortstack{\\97.45\\97.06$\pm$2.66}
		&\shortstack{\\94.46\\93.62$\pm$2.87}
		&\shortstack{\\94.92\\94.77$\pm$3.82}
		&\shortstack{\\74.28\\77.80$\pm$13.43}
		&\shortstack{\\0.89\\16.17$\pm$24.73}
		\\
		\hline
		Specificity(\%)
		&\shortstack{\\99.33\\99.29$\pm$0.25}
		&\shortstack{\\99.16\\98.98$\pm$0.65}
		&\shortstack{\\98.68\\97.28$\pm$5.46}
		&\shortstack{\\99.73\\99.62$\pm$0.28}
		&\shortstack{\\98.90\\93.69$\pm$22.08}
		&\shortstack{\\99.82\\99.29$\pm$2.38}
		&\cellcolor{mybetter}\shortstack{\\100.00\\99.97$\pm$0.05}
		\\
		\hline
		
		\multicolumn{8}{|c|}{\textbf{Dataset: TBI}}\\
		\hline
		&PCA &ROBEX &BEaST &MASS &BET & BSE & CNN\\
		\hline
		Dice(\%) 
		&\cellcolor{mybetter}\shortstack{\\96.40\\96.28$\pm$0.85}
		&\shortstack{\\93.71\\93.60$\pm$1.00}
		&\shortstack{\\95.04\\95.06$\pm$0.96}
		&\shortstack{\\95.11\\95.42$\pm$0.96}
		&\shortstack{\\95.40\\95.14$\pm$1.12}
		&\shortstack{\\92.64\\91.00$\pm$4.31}
		&\shortstack{\\91.51\\90.40$\pm$5.07}
		\\
		\hline
		Avg Dist(mm) 
		&\cellcolor{mybetter}\shortstack{\\1.20\\1.22$\pm$0.30}
		&\shortstack{\\2.12\\2.20$\pm$0.40}
		&\shortstack{\\1.60\\1.62$\pm$0.33}
		&\shortstack{\\1.61\\1.53$\pm$0.33}
		&\shortstack{\\1.50\\1.57$\pm$0.40}
		&\shortstack{\\2.23\\3.15$\pm$1.66}
		&\shortstack{\\3.20\\3.46$\pm$1.75}
		\\
		\hline
		95\% Dist(mm) 
		&\cellcolor{mybetter}\shortstack{\\3.37\\3.41$\pm$0.85}
		&\shortstack{\\6.20\\5.99$\pm$0.97}
		&\shortstack{\\6.24\\5.86$\pm$1.44}
		&\shortstack{\\4.90\\4.59$\pm$0.77}
		&\shortstack{\\5.06\\5.57$\pm$1.91}
		&\shortstack{\\9.46\\13.07$\pm$7.11}
		&\shortstack{\\14.97\\16.04$\pm$8.72}
		\\
		\hline
		Max Dist(mm) 
		&\shortstack{\\16.13\\15.54$\pm$5.03}
		&\shortstack{\\18.25\\18.89$\pm$5.12}
		&\cellcolor{mybetter}\shortstack{\\15.09\\17.27$\pm$4.60}
		&\shortstack{\\15.65\\16.08$\pm$4.16}
		&\shortstack{\\17.46\\18.53$\pm$4.59}
		&\shortstack{\\27.16\\31.96$\pm$12.71}
		&\shortstack{\\41.49\\37.06$\pm$10.09}
		\\
		\hline
		Sensitivity(\%)
		&\shortstack{\\97.82\\97.76$\pm$0.92}
		&\cellcolor{mybetter}\shortstack{\\98.91\\98.64$\pm$0.93}
		&\shortstack{\\92.12\\92.89$\pm$2.44}
		&\shortstack{\\98.74\\98.68$\pm$0.66}
		&\shortstack{\\93.98\\93.65$\pm$1.87}
		&\shortstack{\\87.51\\85.65$\pm$8.17}
		&\shortstack{\\85.69\\83.77$\pm$8.58}
		\\
		\hline
		Specificity(\%)
		&\shortstack{\\99.03\\99.07$\pm$0.26}
		&\shortstack{\\98.09\\97.75$\pm$0.93}
		&\shortstack{\\99.63\\99.57$\pm$0.22}
		&\shortstack{\\98.64\\98.59$\pm$0.47}
		&\shortstack{\\99.41\\99.44$\pm$0.11}
		&\cellcolor{mybetter}\shortstack{\\99.82\\99.54$\pm$0.86}
		&\shortstack{\\99.80\\99.81$\pm$0.07}
		\\
		\hline
	\end{tabular}
	\caption{Medians (top), and means with standard deviations (bottom) for validation measures for all the methods and all the datasets. We highlight the best results in \textcolor{green}{green} based on the \textit{median} values. Among all datasets, our PCA model has the best median on Dice overlap scores and generally on surface distances. \xhsp{Exception is BEaST which achieves a lower maximum surface distances on the LPBA40 and the TBI datasets}. In addition, our model also has the best mean and variance for the Dice overlap scores and the surface distances on most of these datasets.}
	\label{tbl:m_std}
\end{table*}

\begin{table*}[!ht]
	\def\arraystretch{0}
	\centering
	\begin{tabular}{@{}c@{}c@{}c@{}c@{ }c@{ }c@{}c@{}c@{}c@{}}
		& \multicolumn{3}{c}{\textbf{IBSR}} && \multicolumn{3}{c}{\textbf{BRATS}}&\multirow{12}{*}{\includegraphics[scale=0.95]{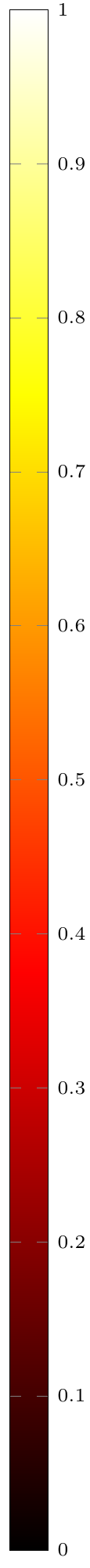}} \\[0.5ex]
		\textbf{PCA} &
		\includegraphics[scale=0.1]{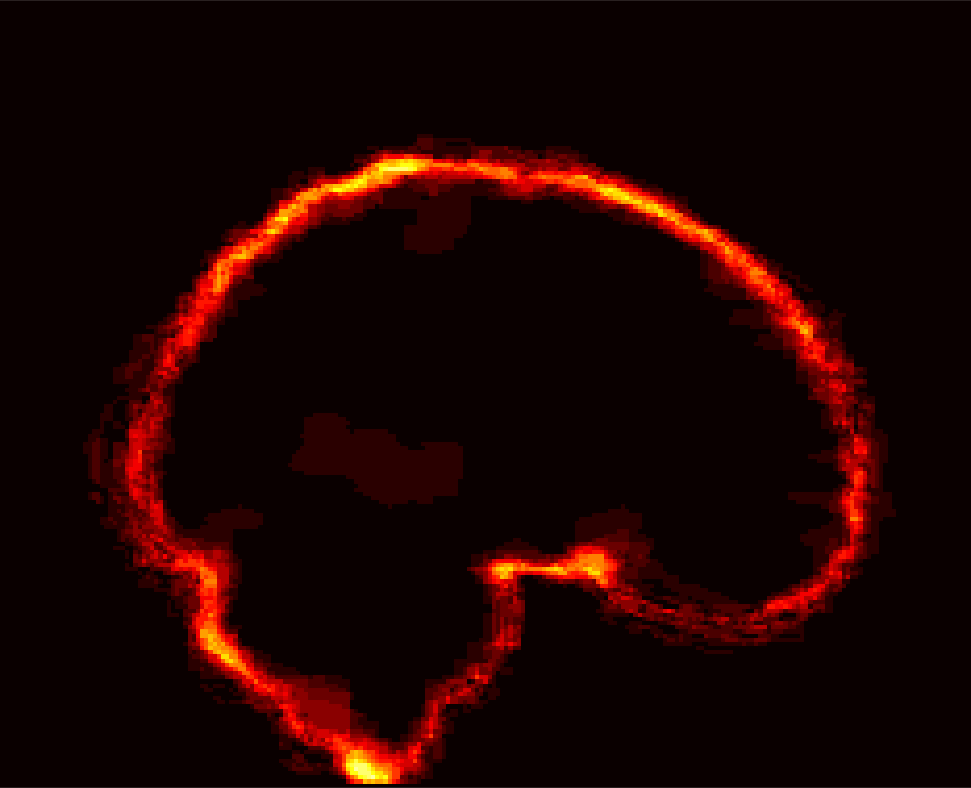} &
		\includegraphics[scale=0.1]{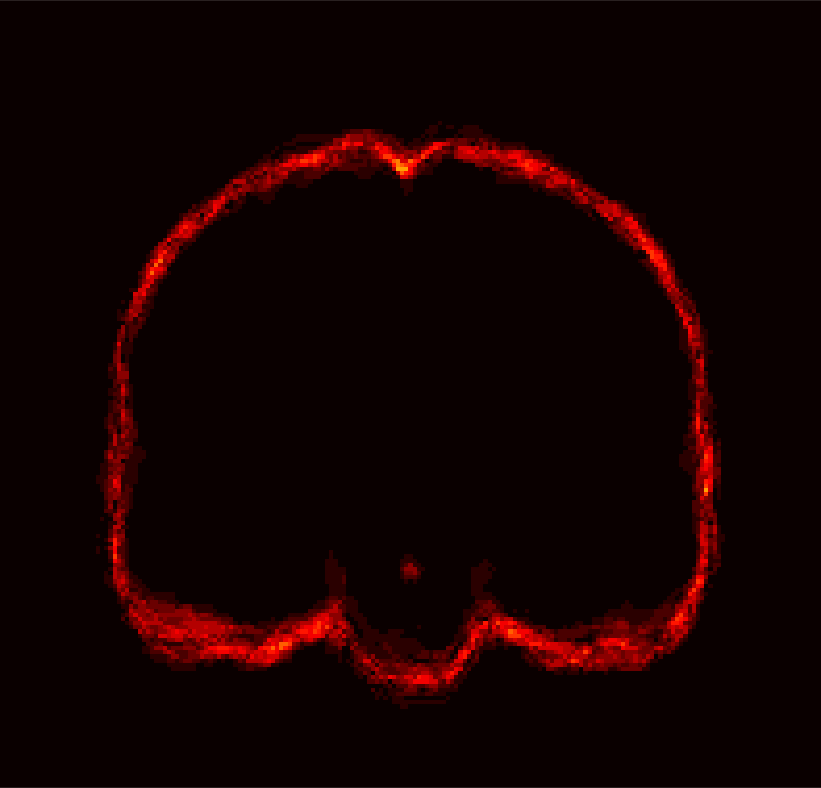} &
		\includegraphics[scale=0.08128]{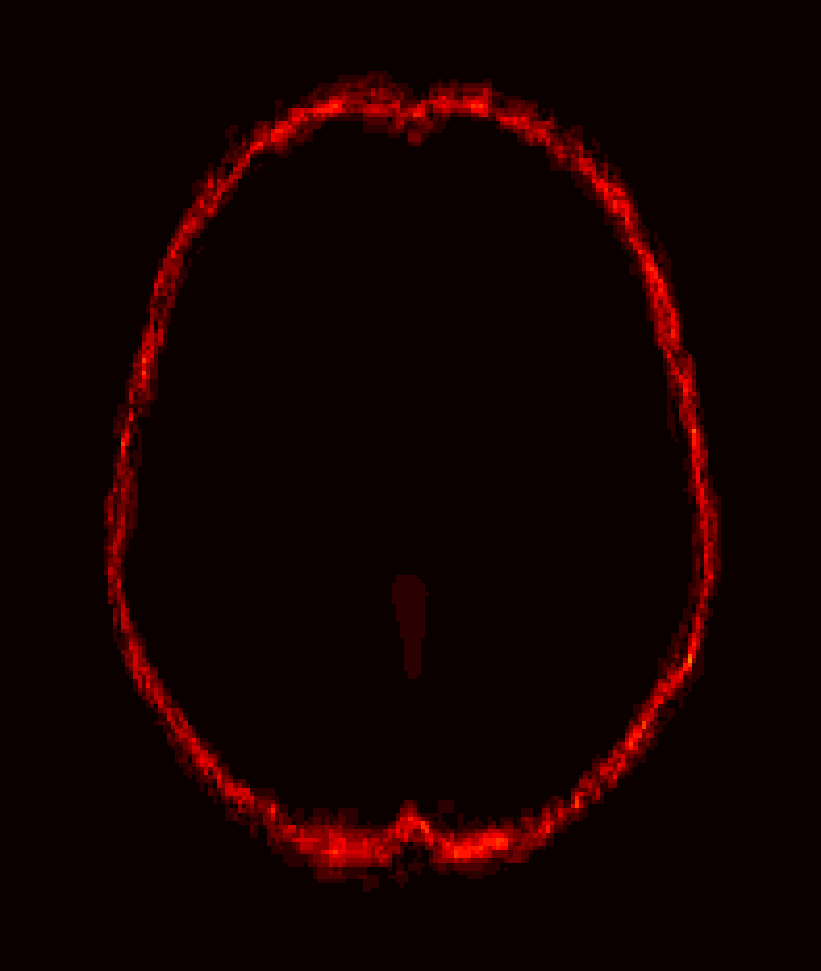} &&
		\includegraphics[scale=0.1]{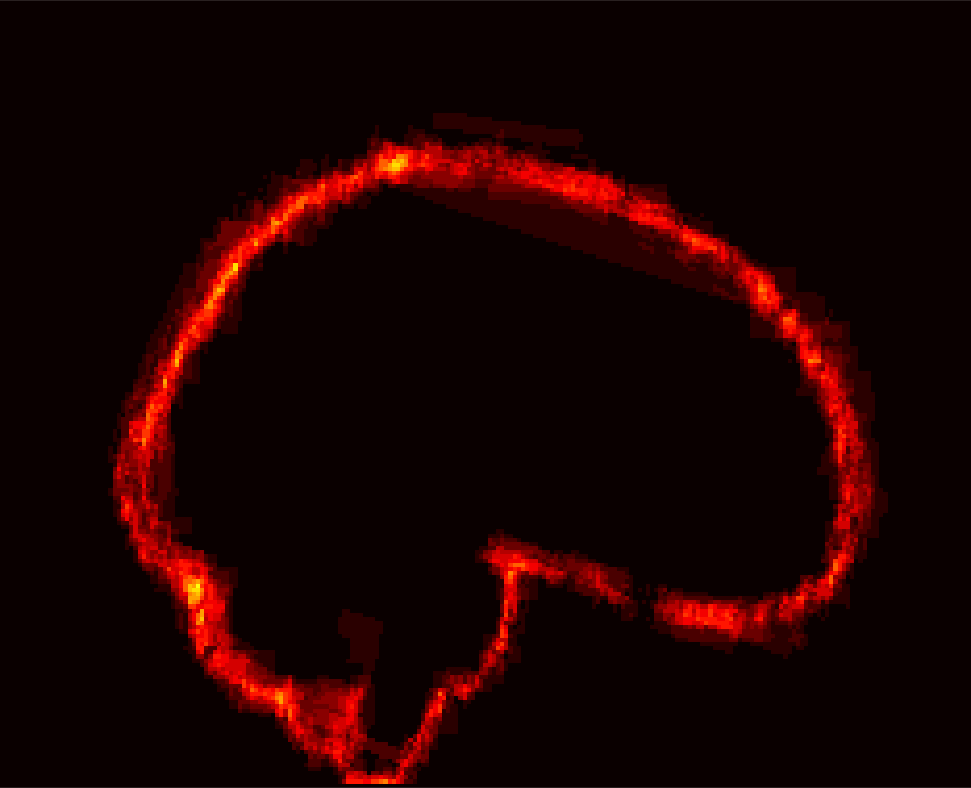} &
		\includegraphics[scale=0.1]{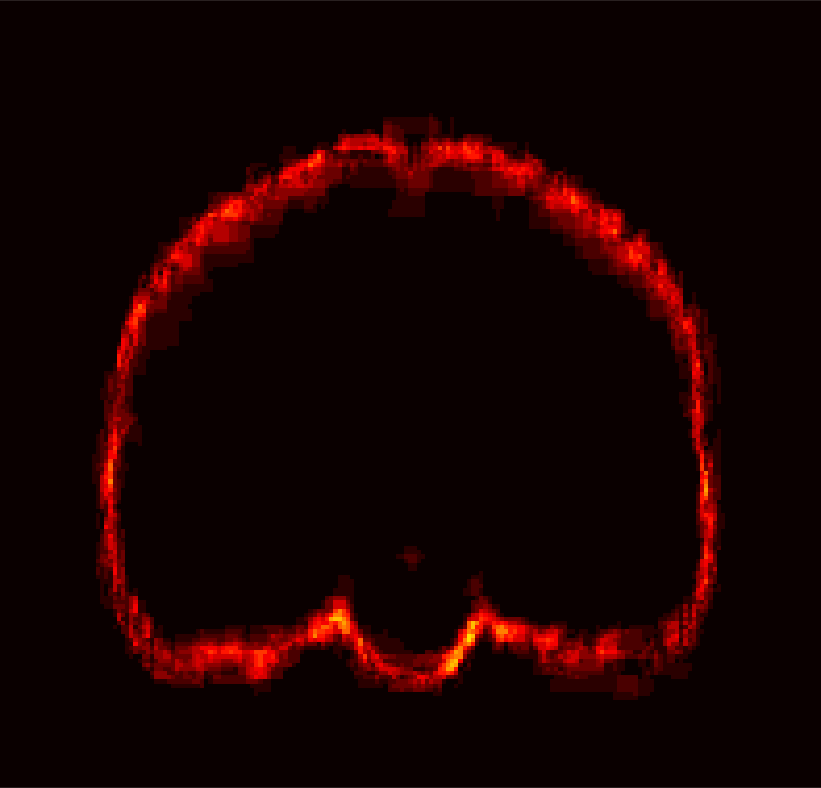} &
		\includegraphics[scale=0.08128]{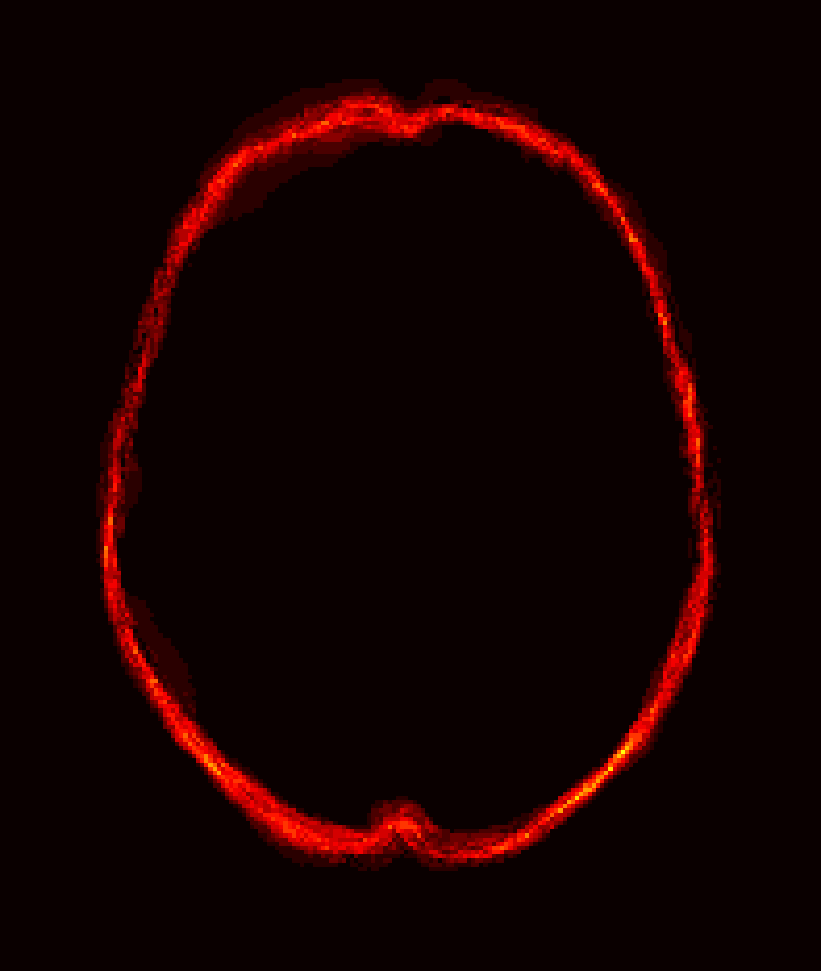}&
		\\
		\textbf{ROBEX} &
		\includegraphics[scale=0.1]{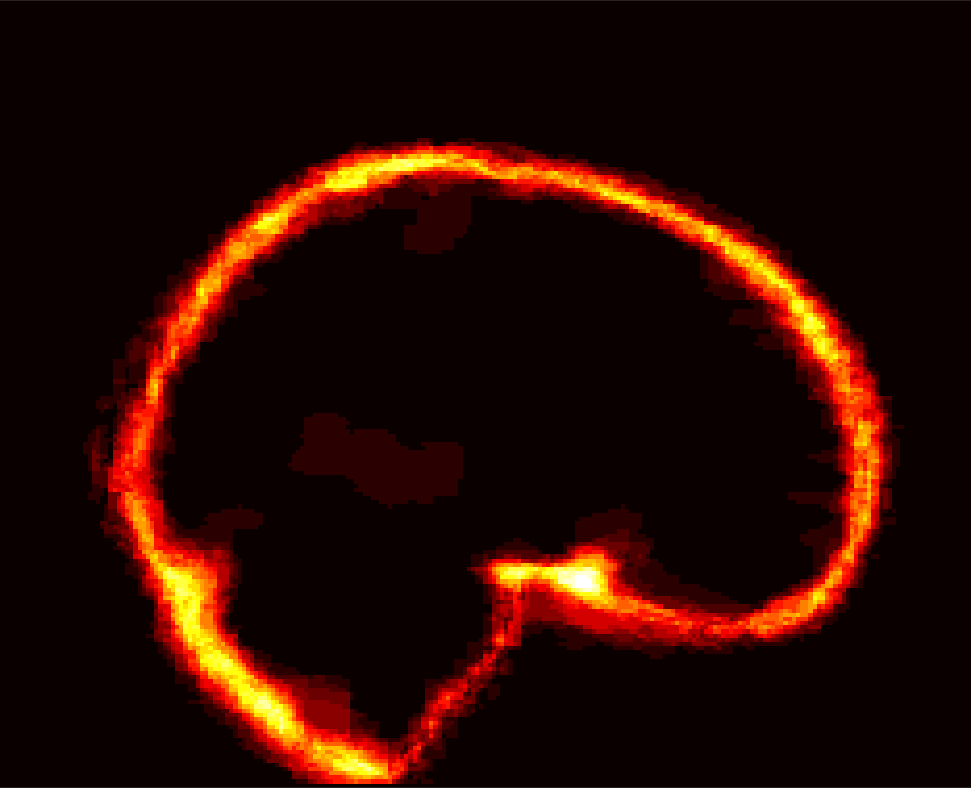} &
		\includegraphics[scale=0.1]{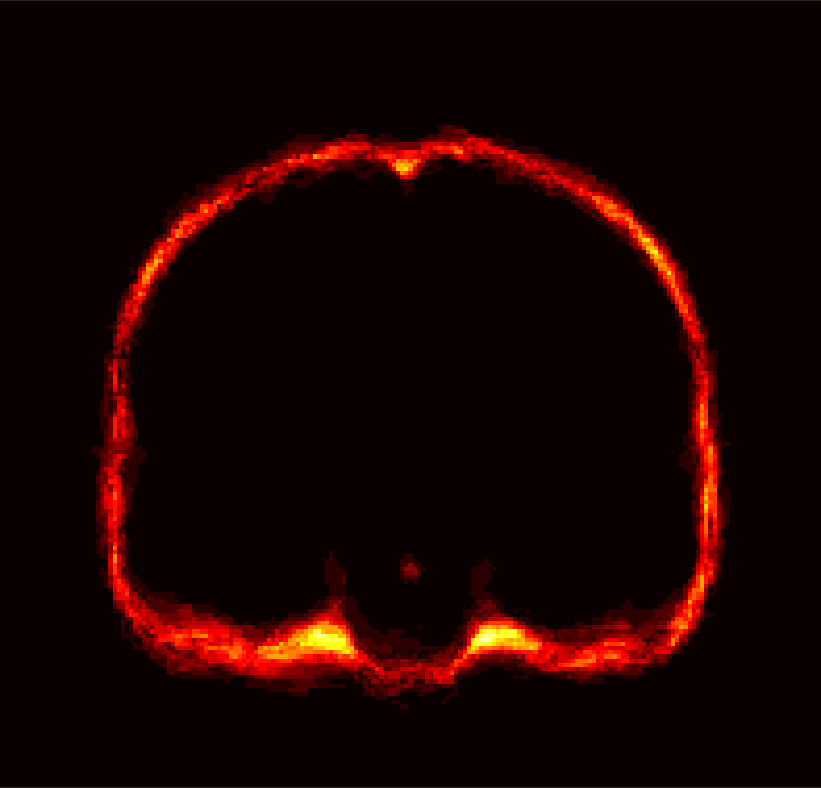} &
		\includegraphics[scale=0.08128]{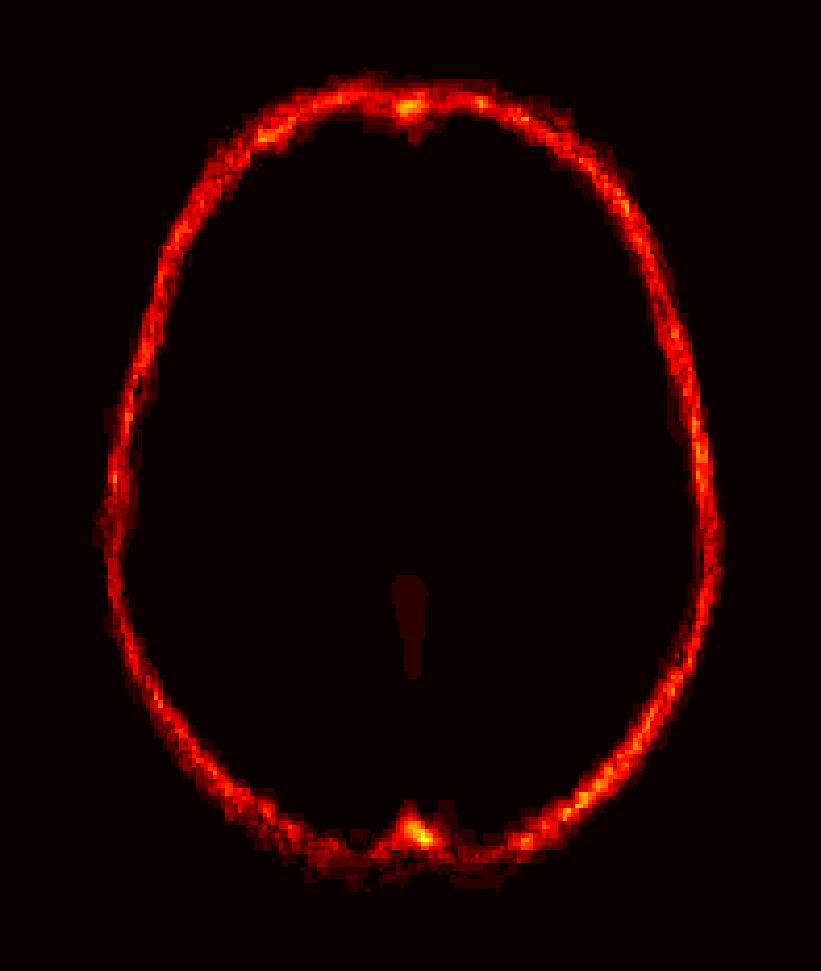} &&
		\includegraphics[scale=0.1]{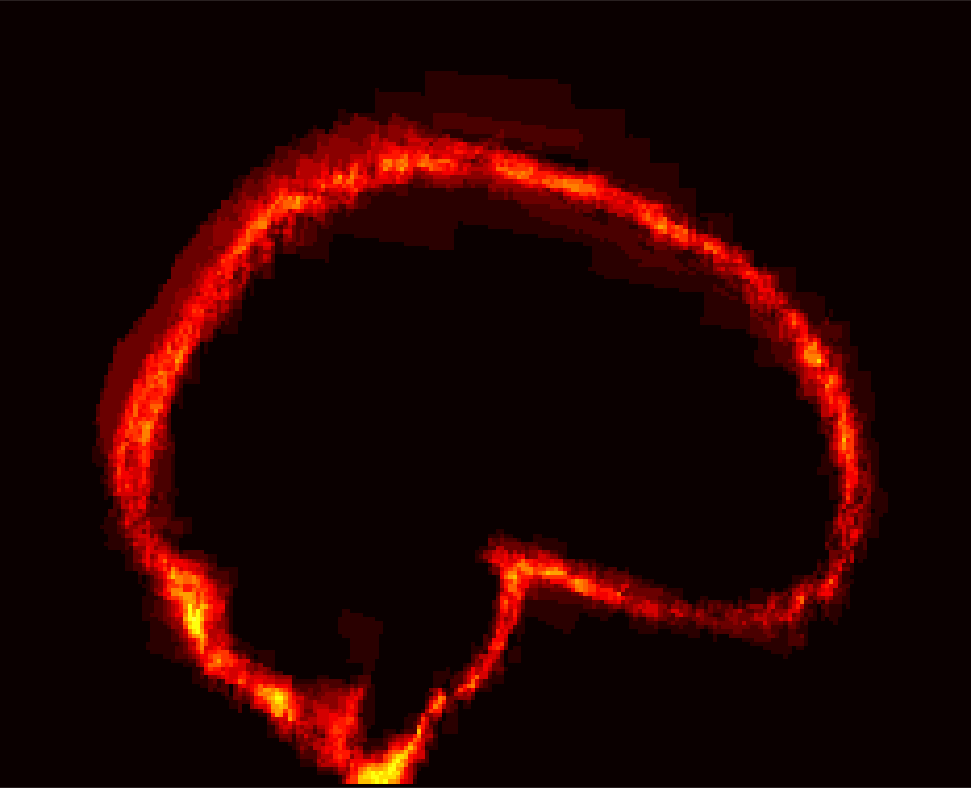} &
		\includegraphics[scale=0.1]{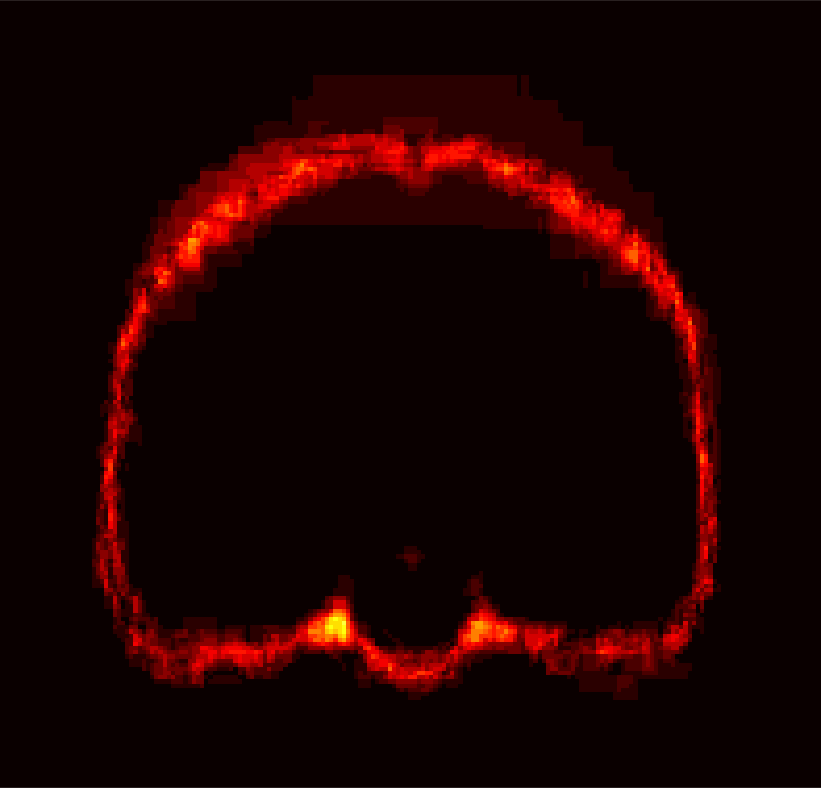} &
		\includegraphics[scale=0.08128]{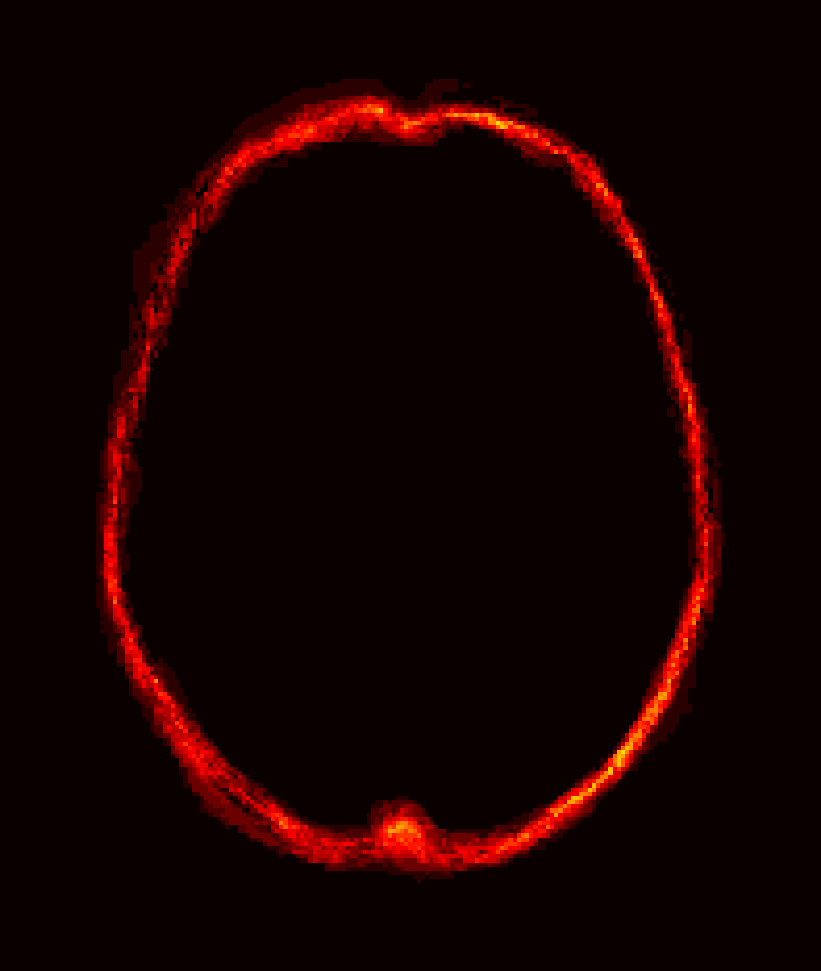} &\\
		\textbf{BEaST*}&
		\includegraphics[scale=0.1]{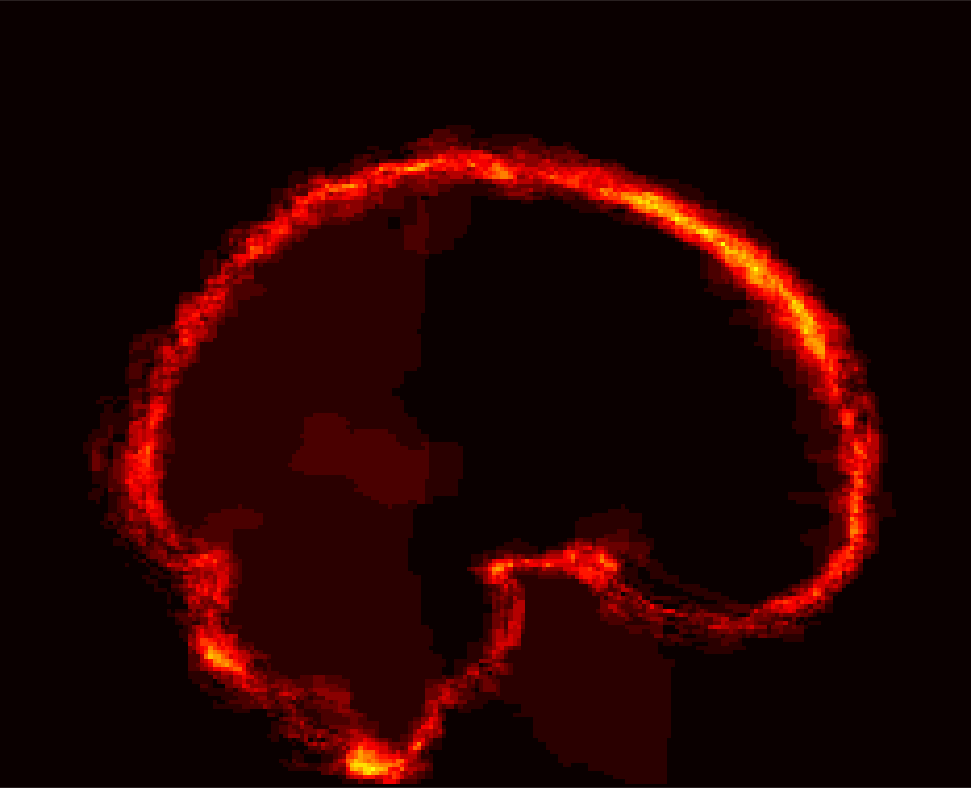} &
		\includegraphics[scale=0.1]{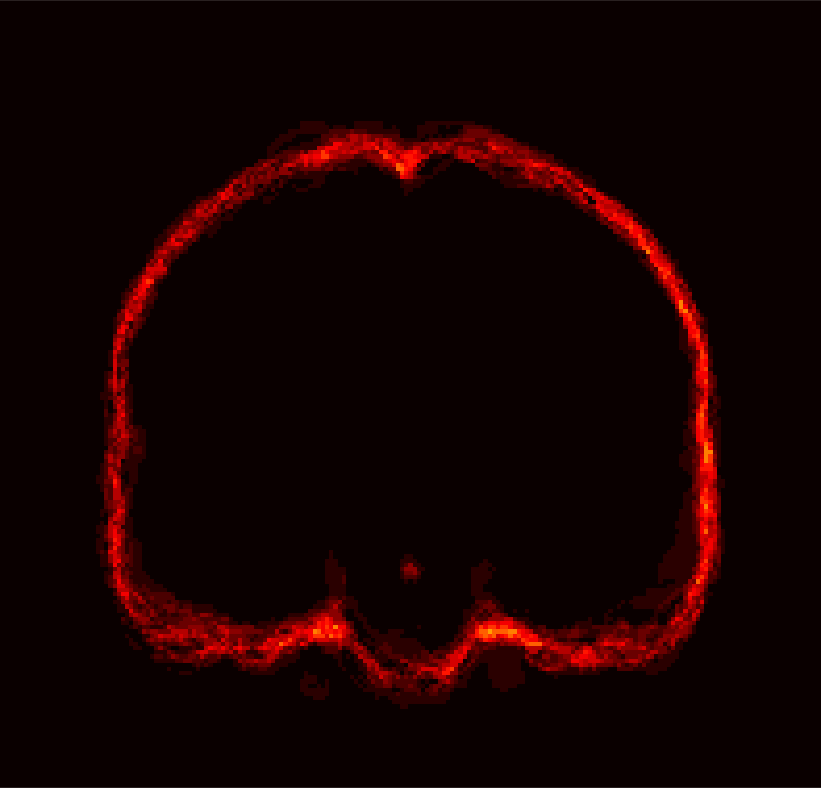} &
		\includegraphics[scale=0.08128]{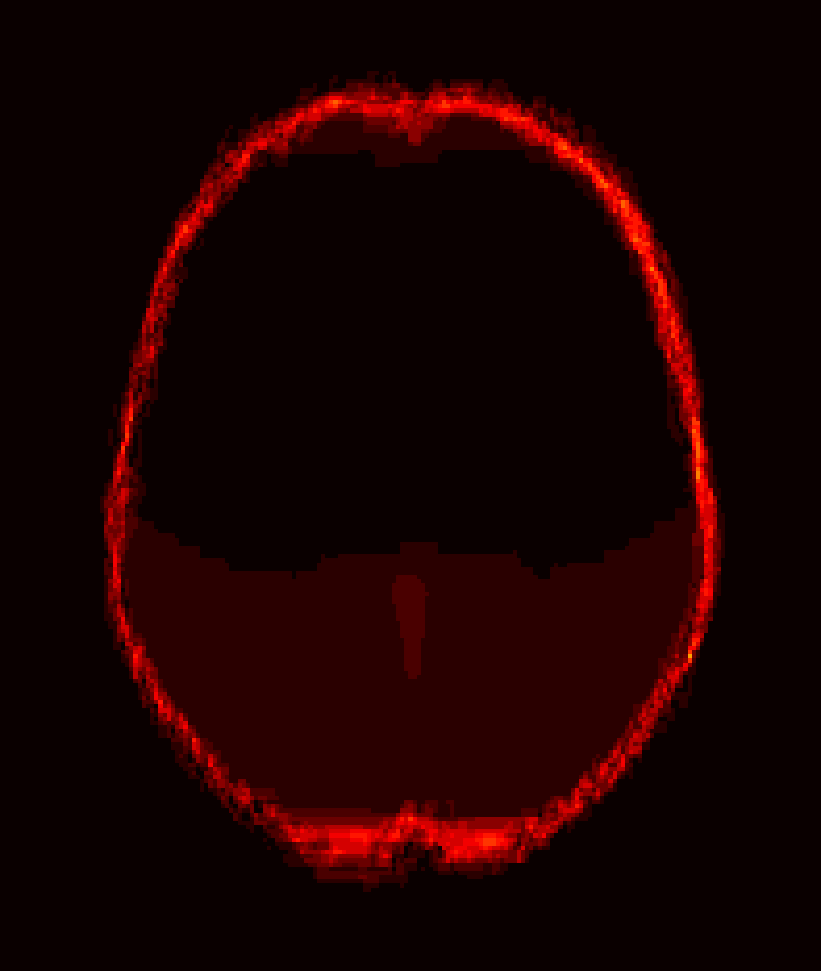} &&
		\includegraphics[scale=0.1]{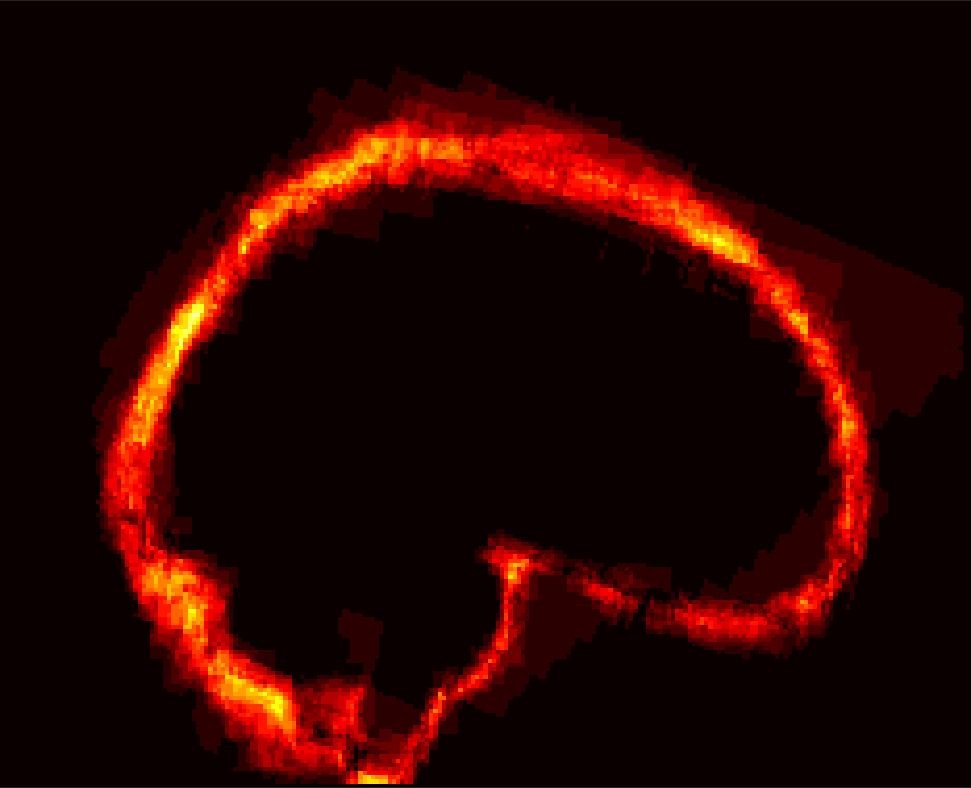} &
		\includegraphics[scale=0.1]{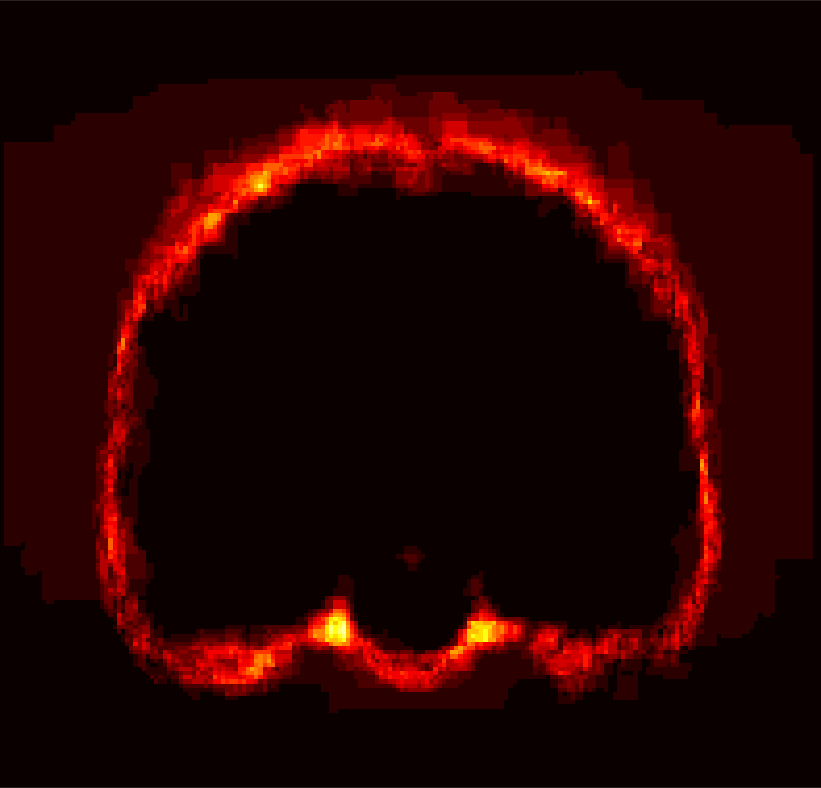} &
		\includegraphics[scale=0.08128]{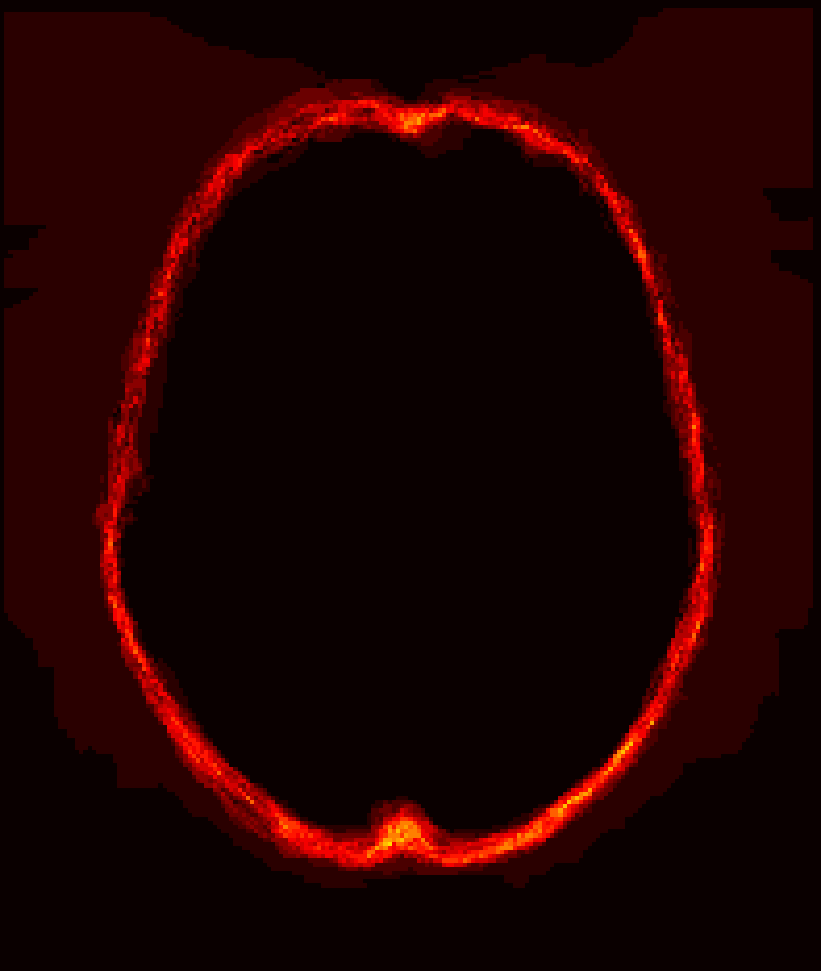} &\\
		\textbf{MASS} &
		\includegraphics[scale=0.1]{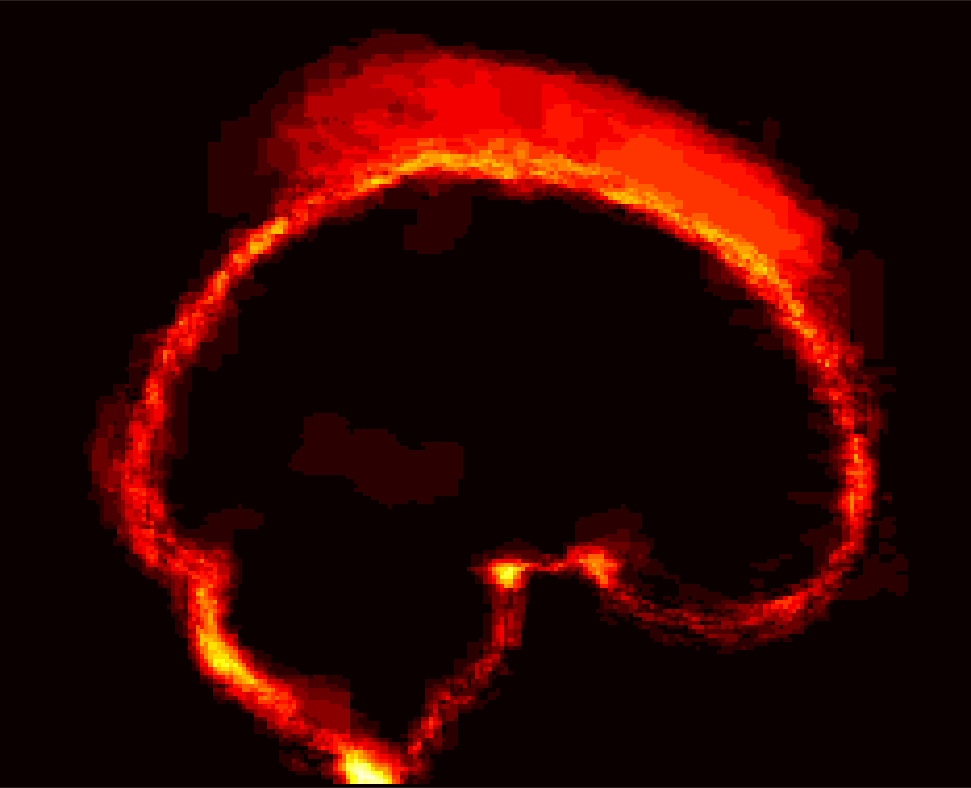} &
		\includegraphics[scale=0.1]{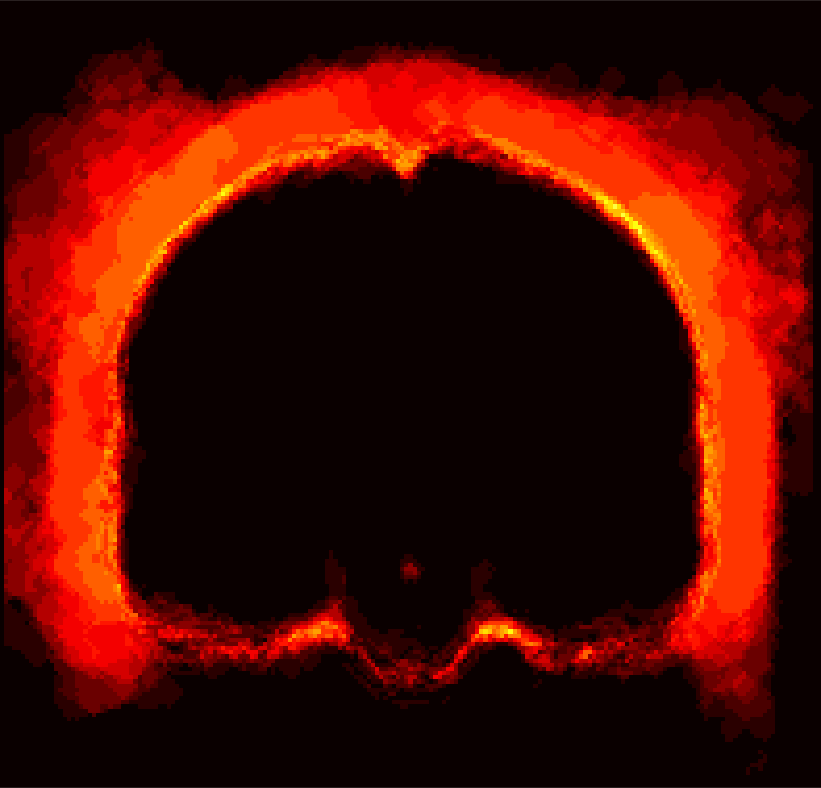} &
		\includegraphics[scale=0.08128]{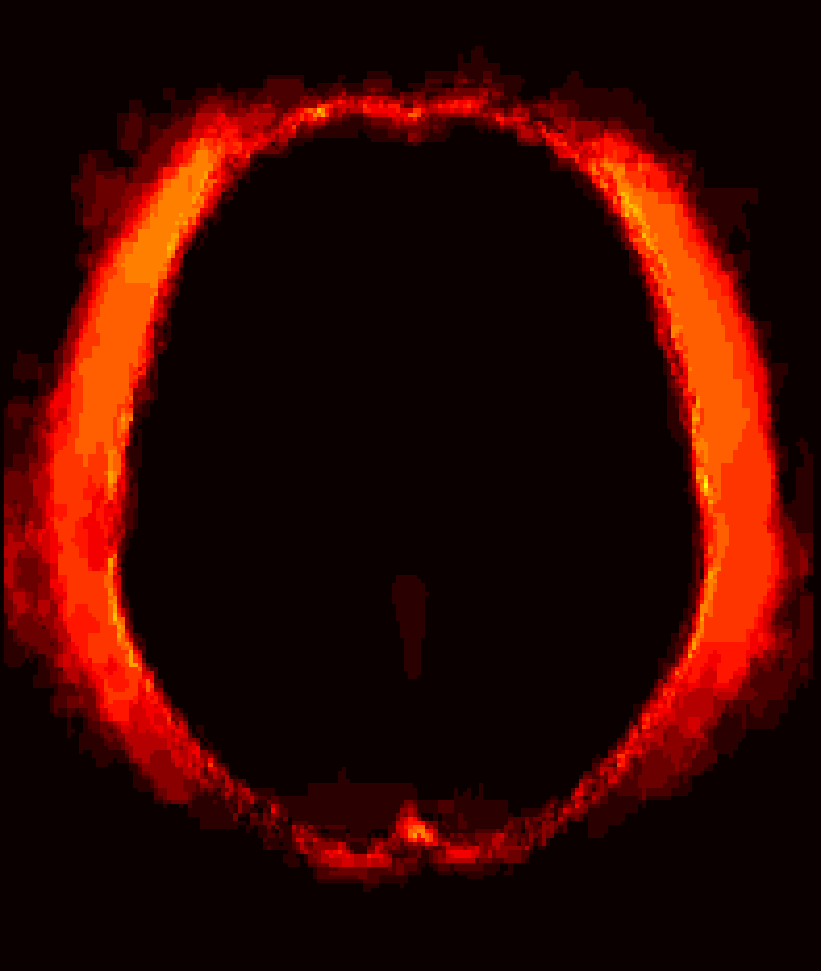} &&
		\includegraphics[scale=0.1]{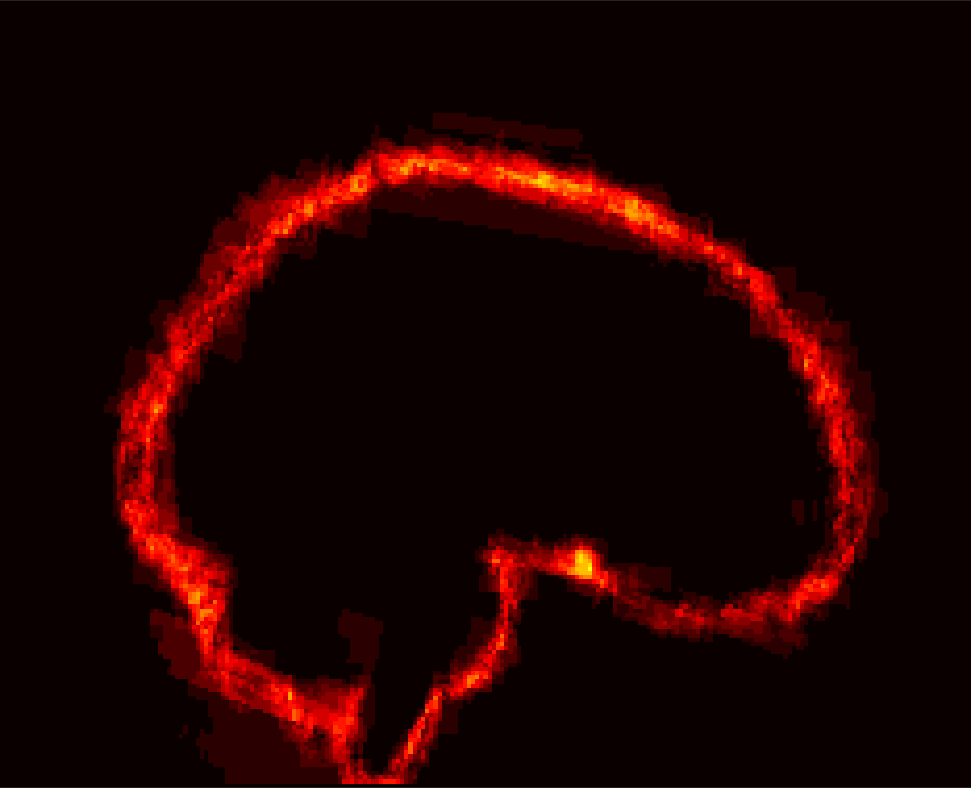} &
		\includegraphics[scale=0.1]{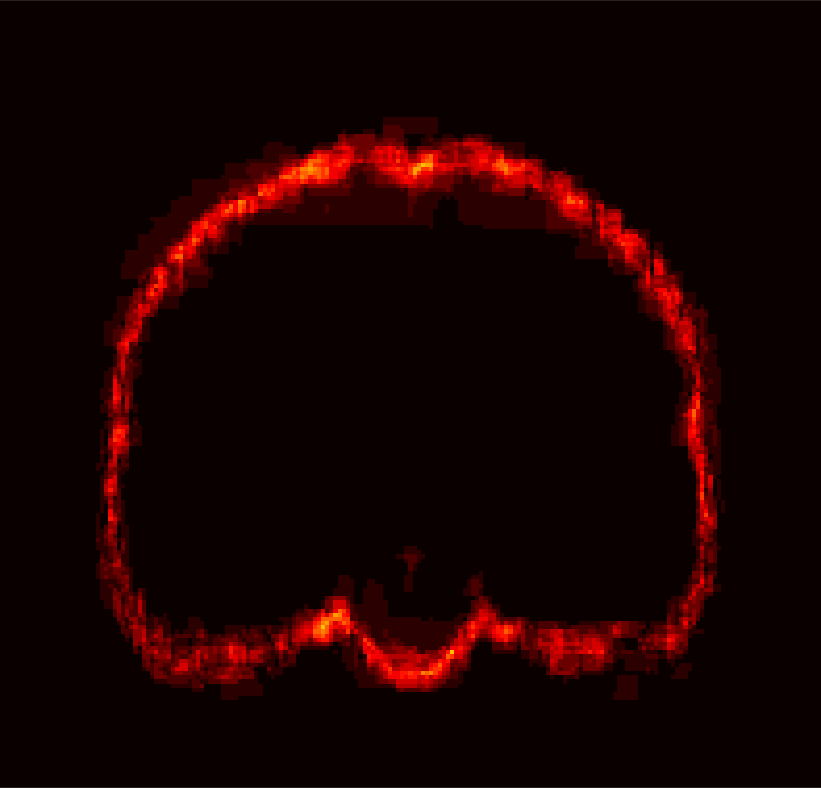} &
		\includegraphics[scale=0.08128]{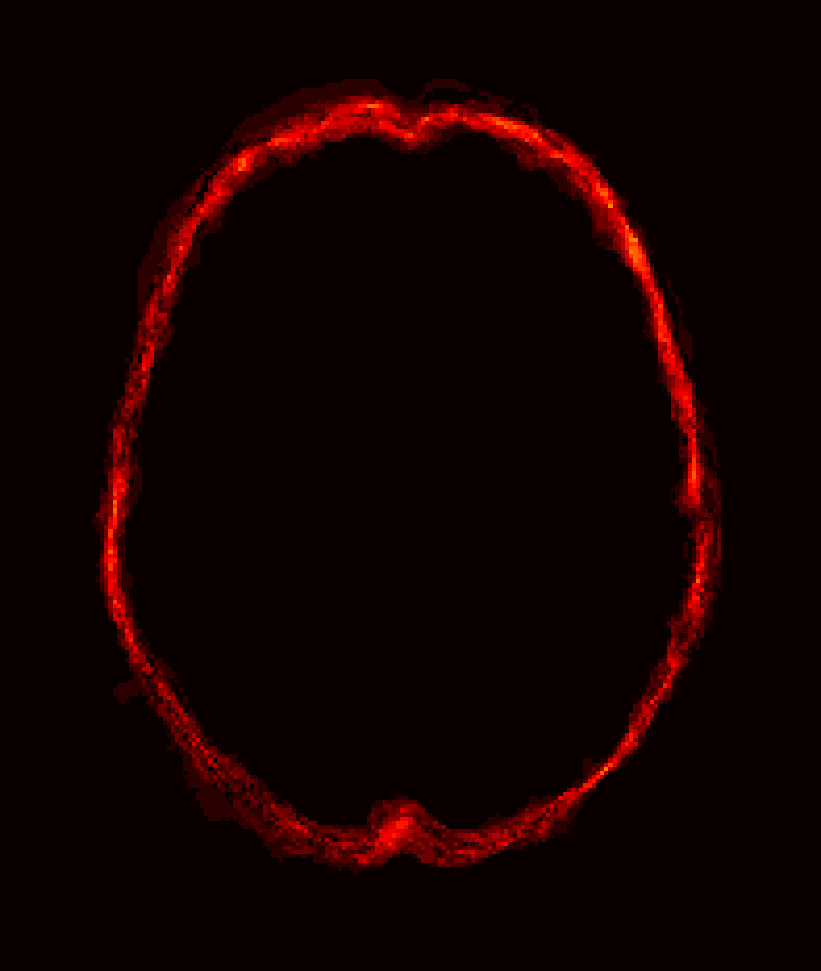} &\\
		\textbf{BET} &
		\includegraphics[scale=0.1]{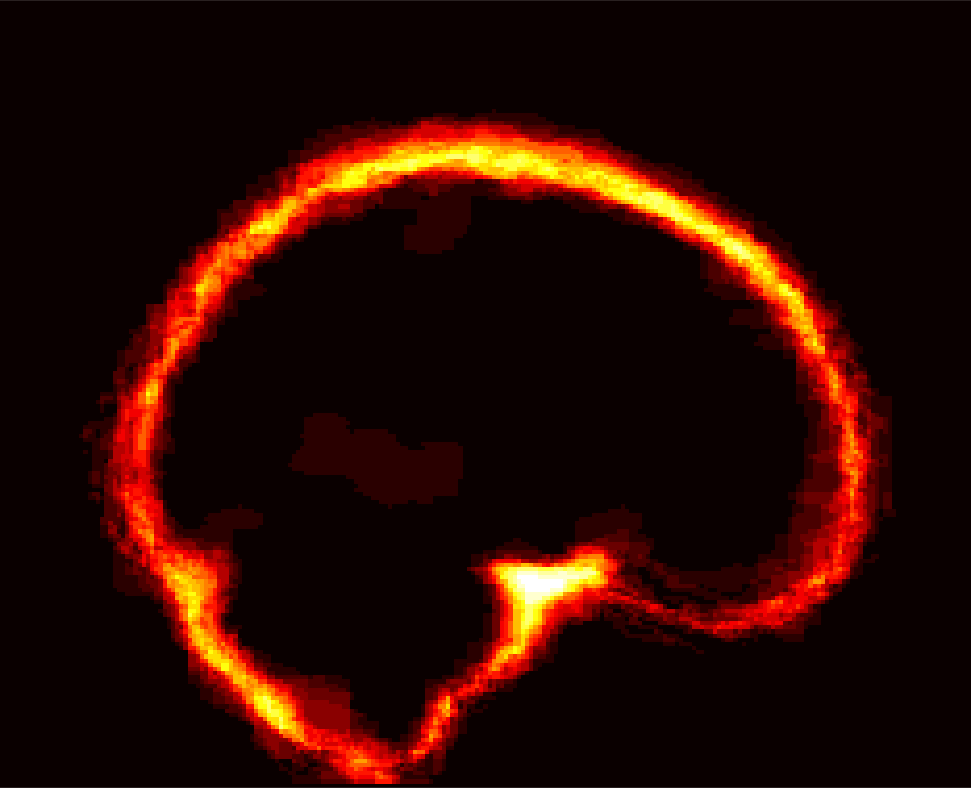} &
		\includegraphics[scale=0.1]{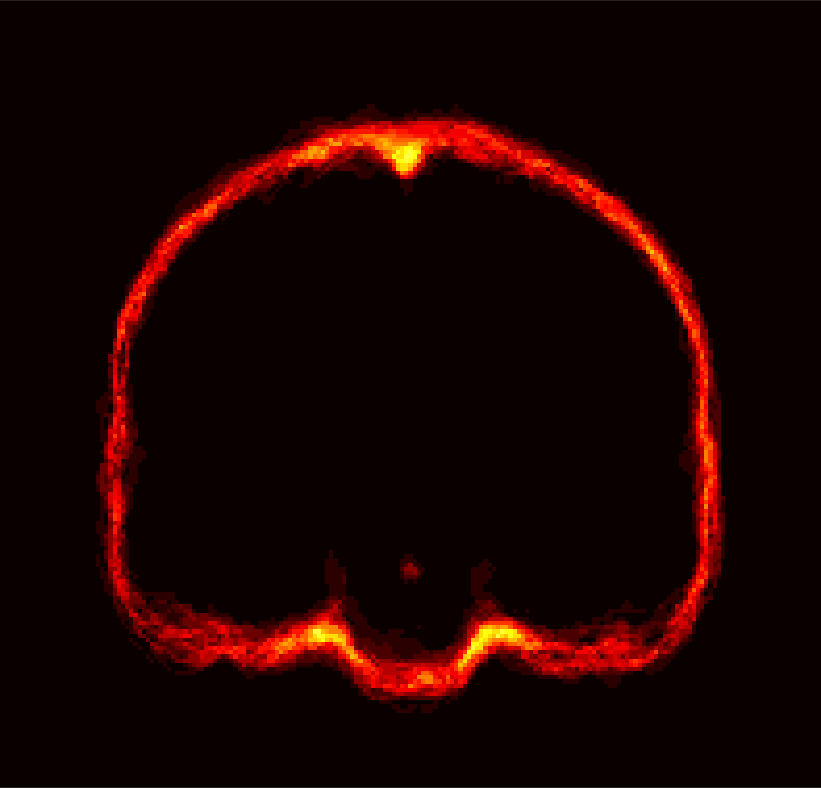} &
		\includegraphics[scale=0.08128]{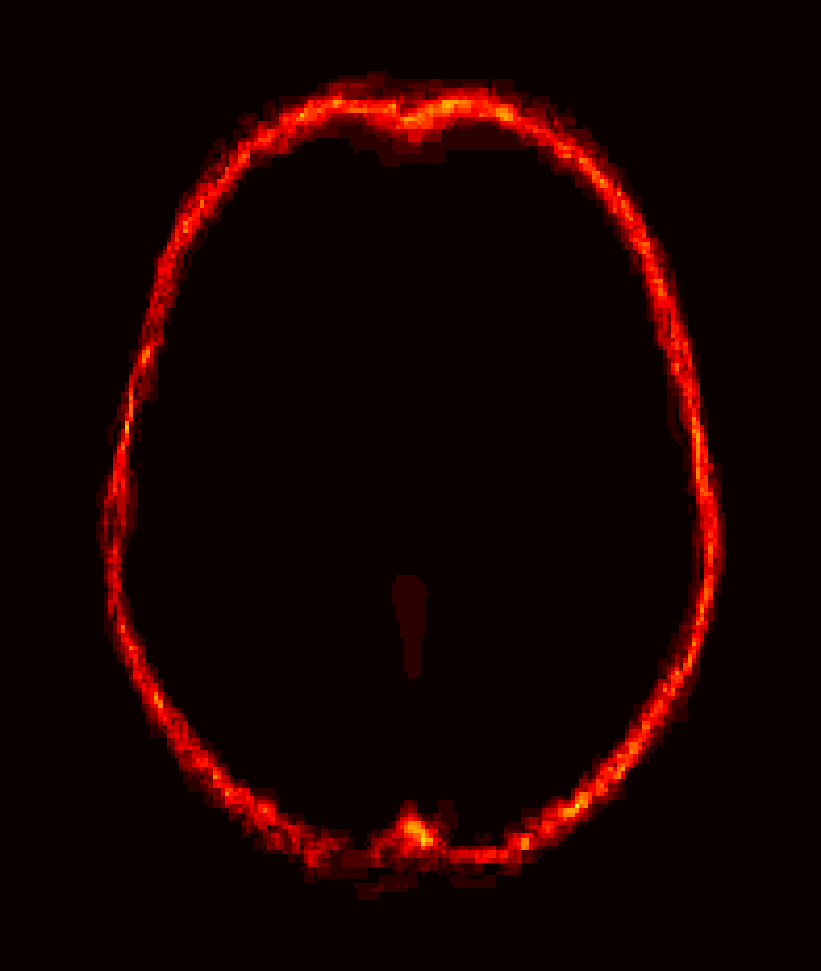} &&
		\includegraphics[scale=0.1]{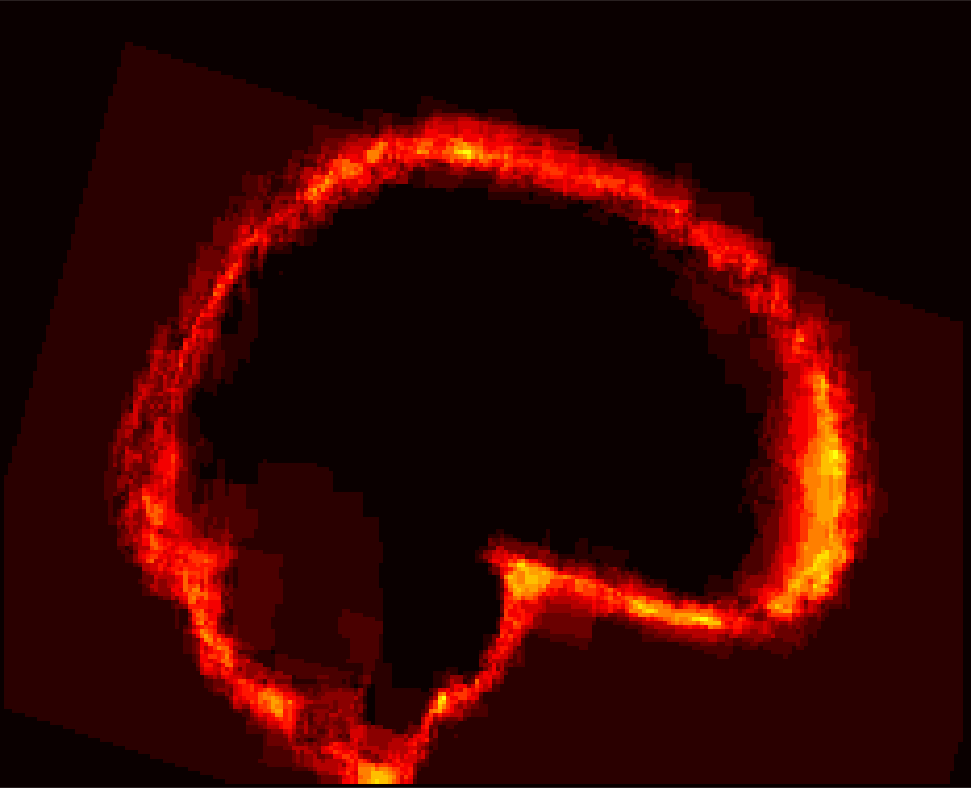} &
		\includegraphics[scale=0.1]{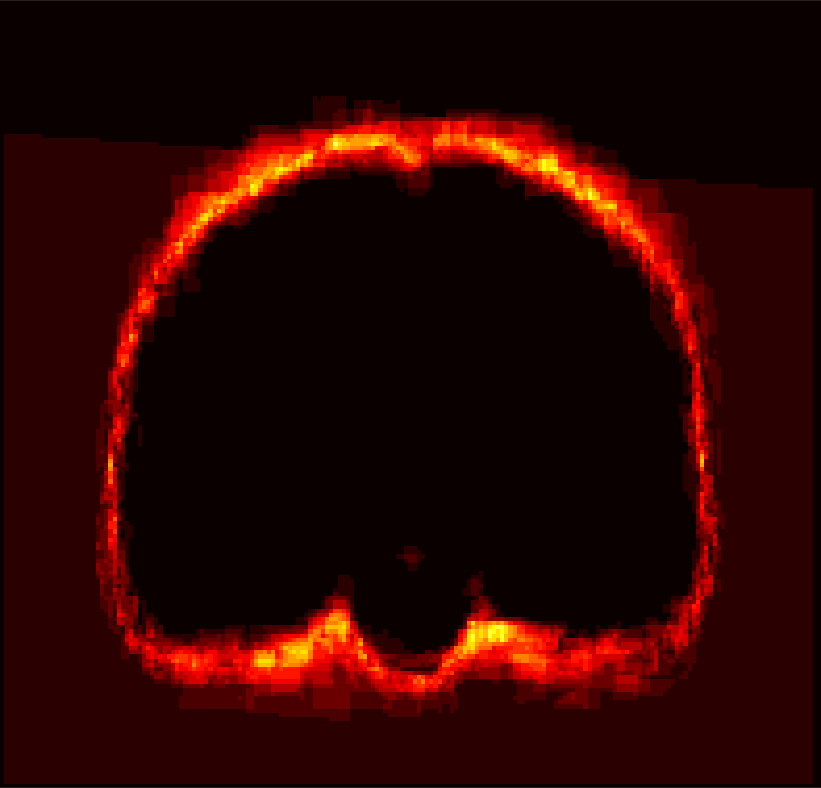} &
		\includegraphics[scale=0.08128]{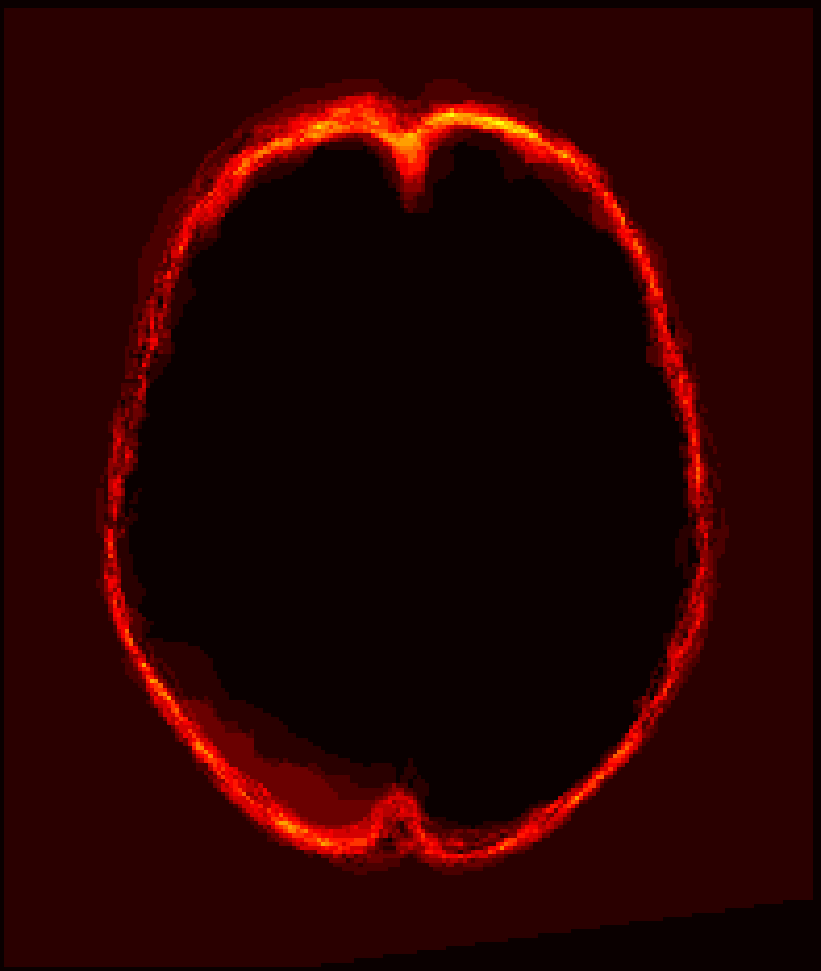} &\\
		\textbf{BSE} &
		\includegraphics[scale=0.1]{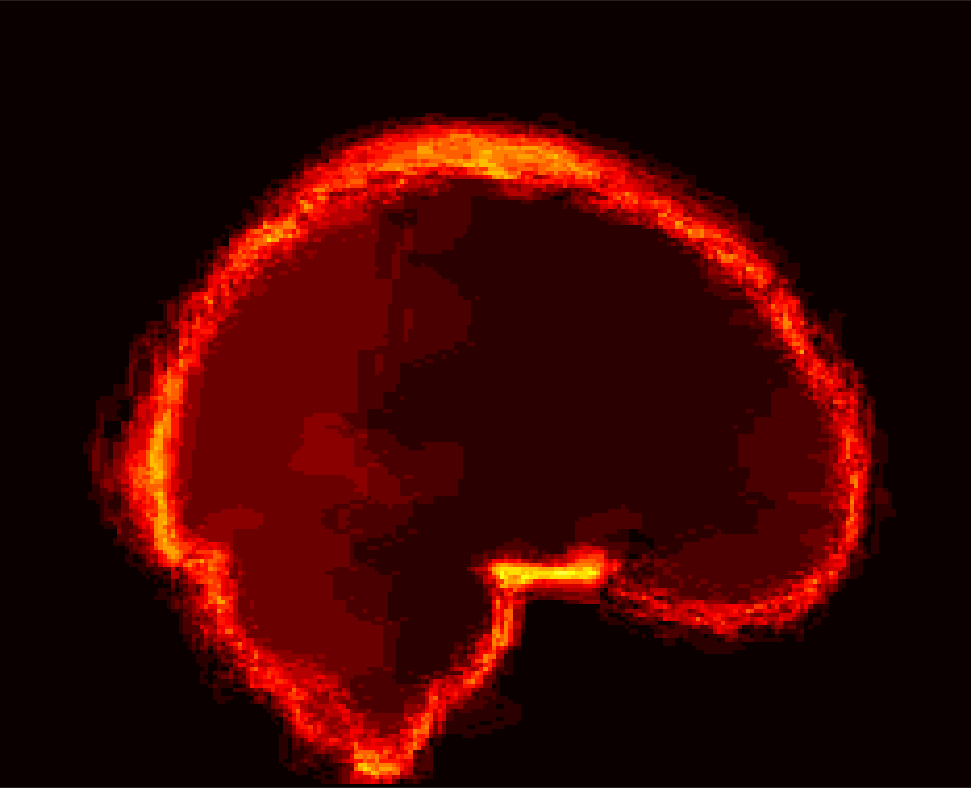} &
		\includegraphics[scale=0.1]{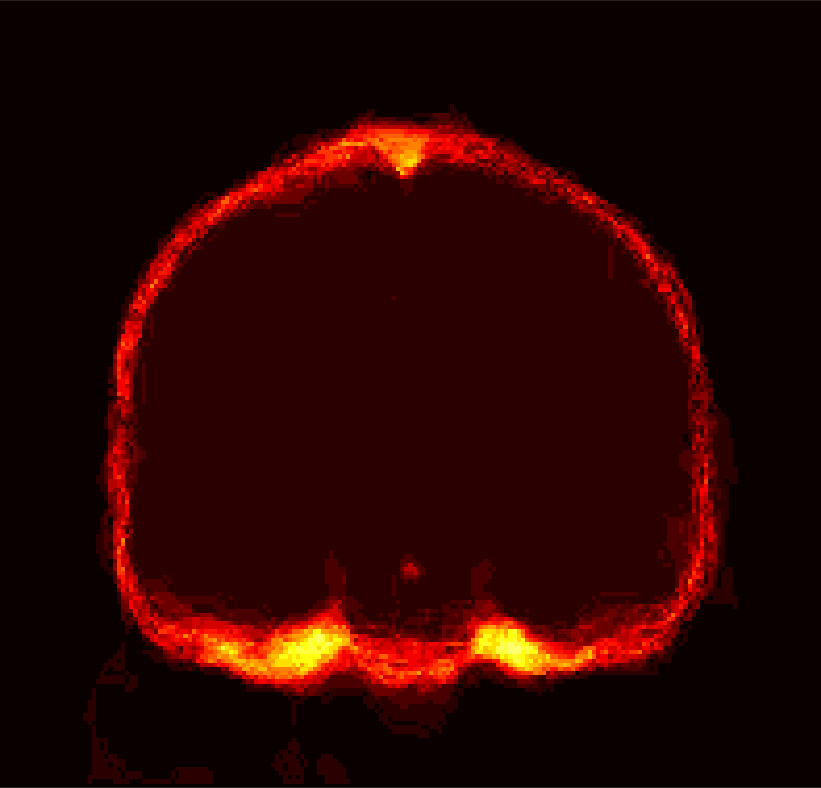} &
		\includegraphics[scale=0.08128]{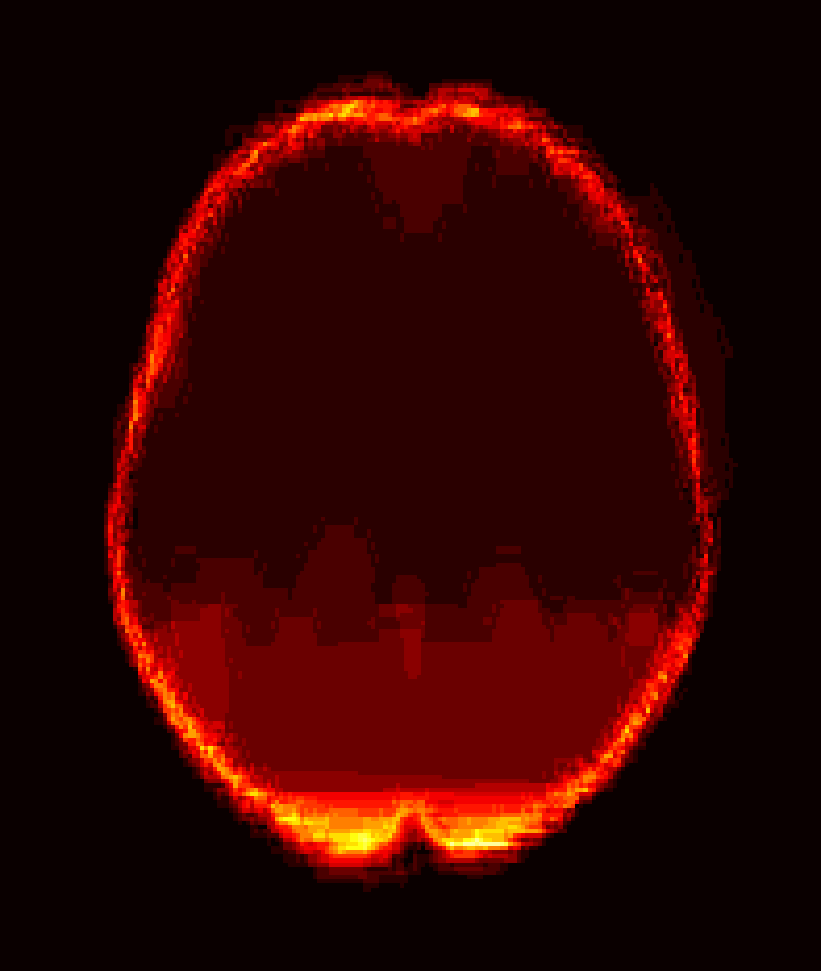} &&
		\includegraphics[scale=0.1]{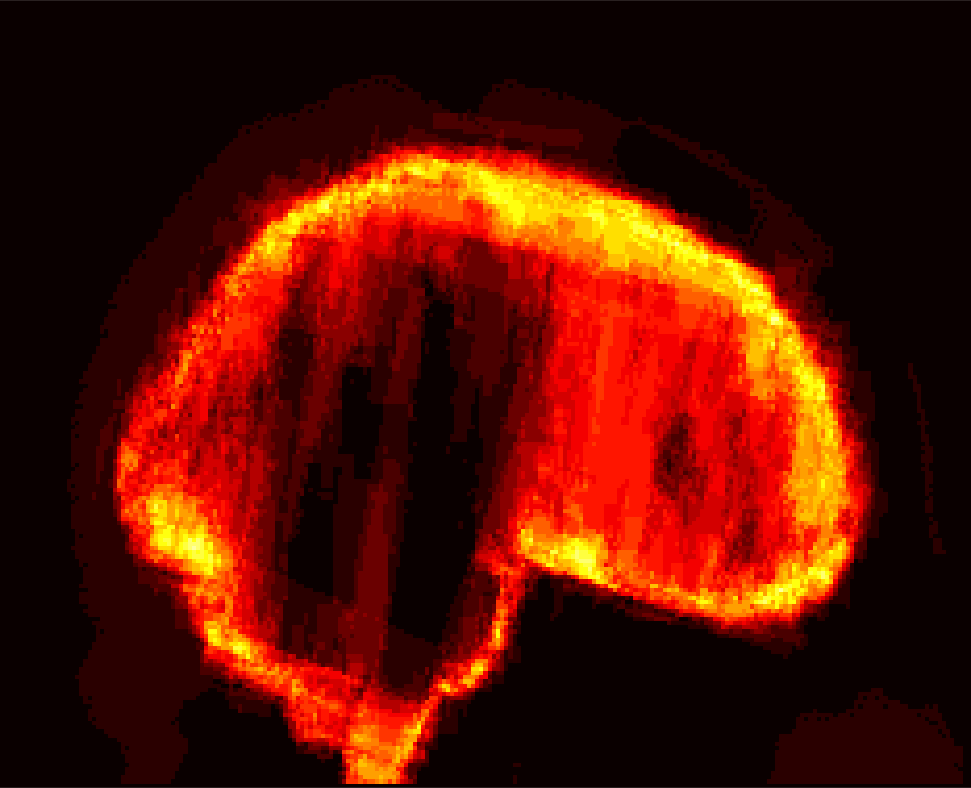} &
		\includegraphics[scale=0.1]{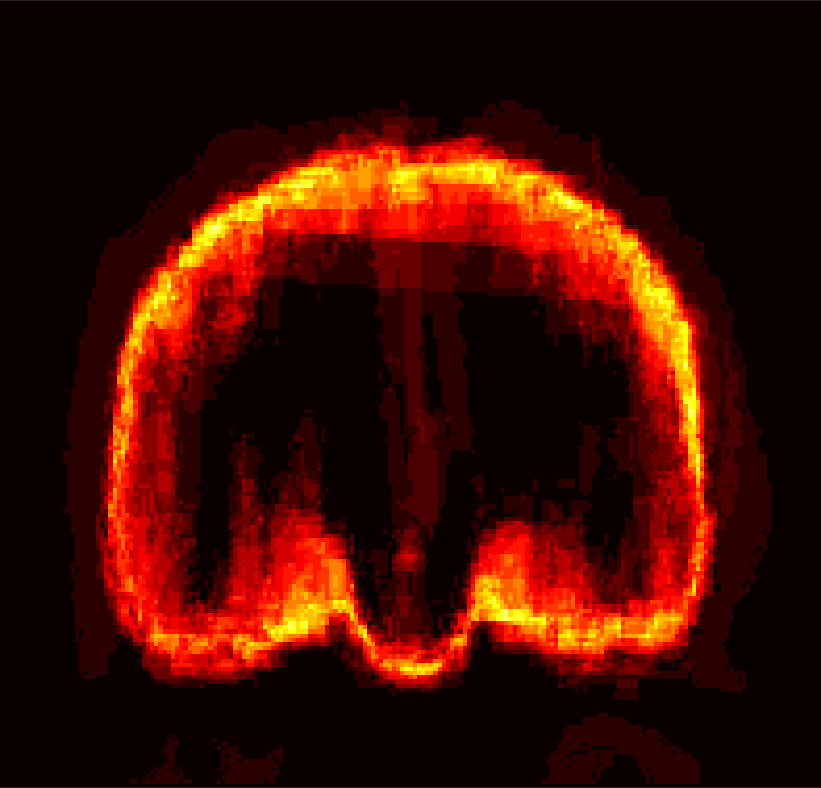} &
		\includegraphics[scale=0.08128]{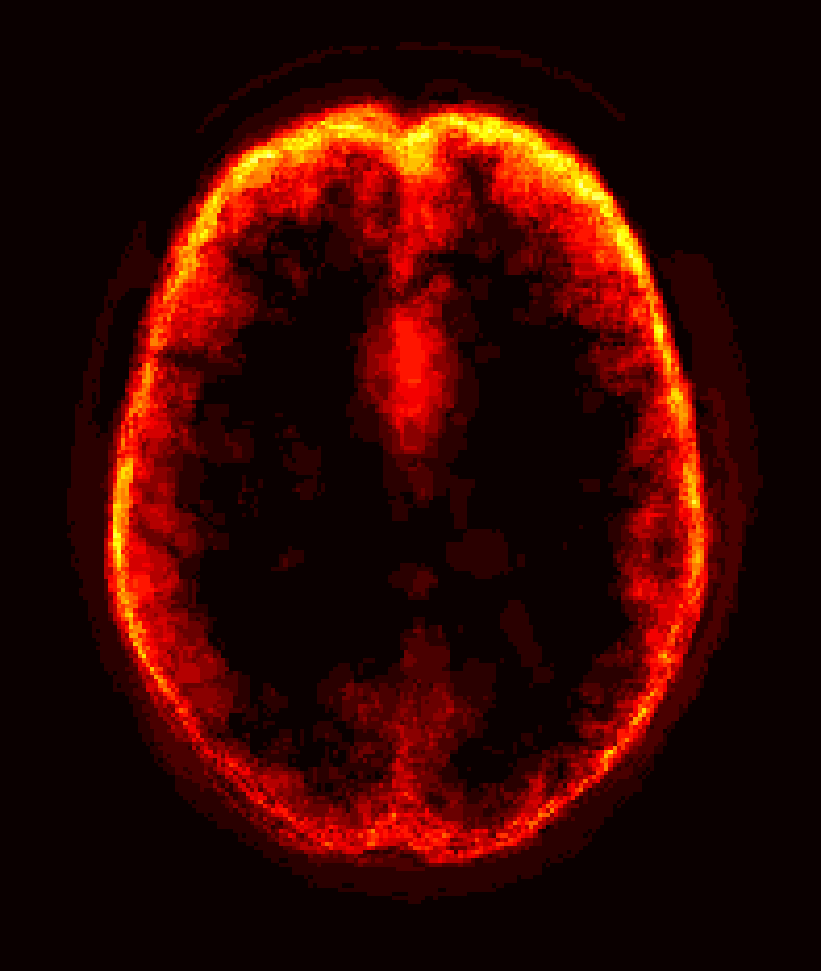} &\\
		\textbf{CNN} &
		\includegraphics[scale=0.1]{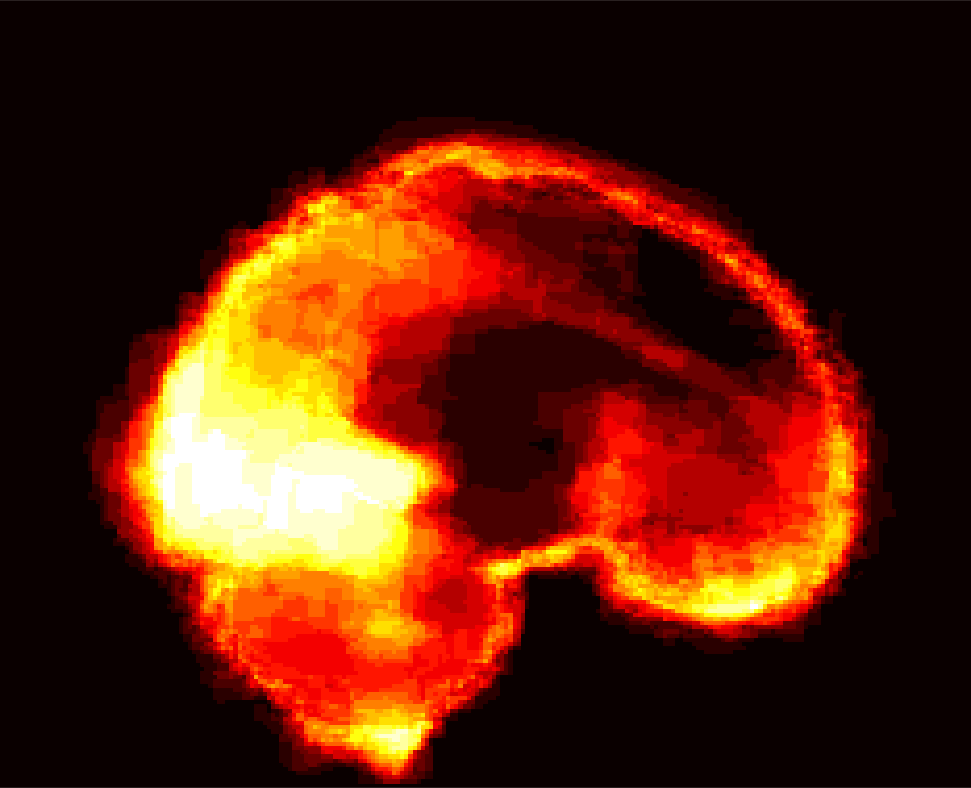} &
		\includegraphics[scale=0.1]{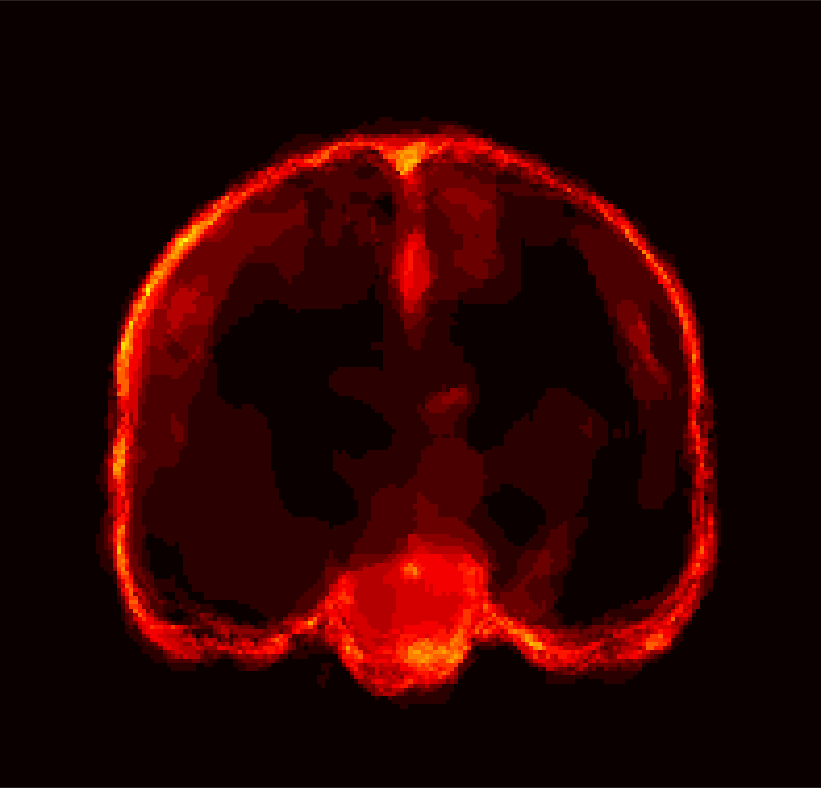} &
		\includegraphics[scale=0.08128]{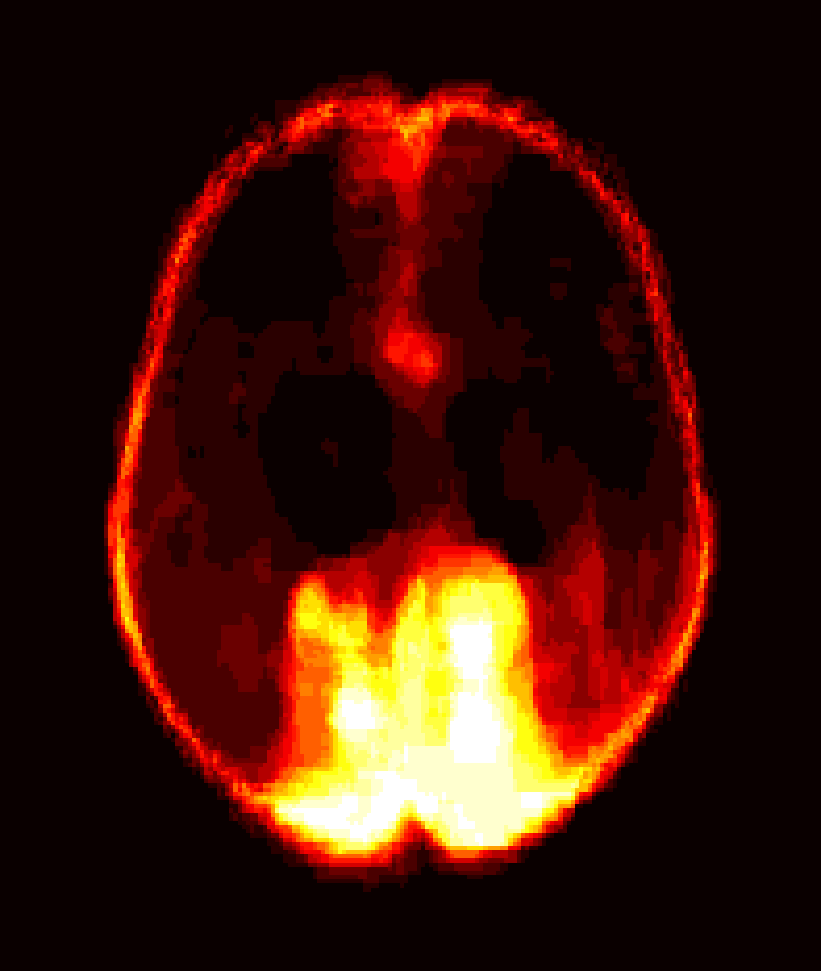} &&
		\includegraphics[scale=0.1]{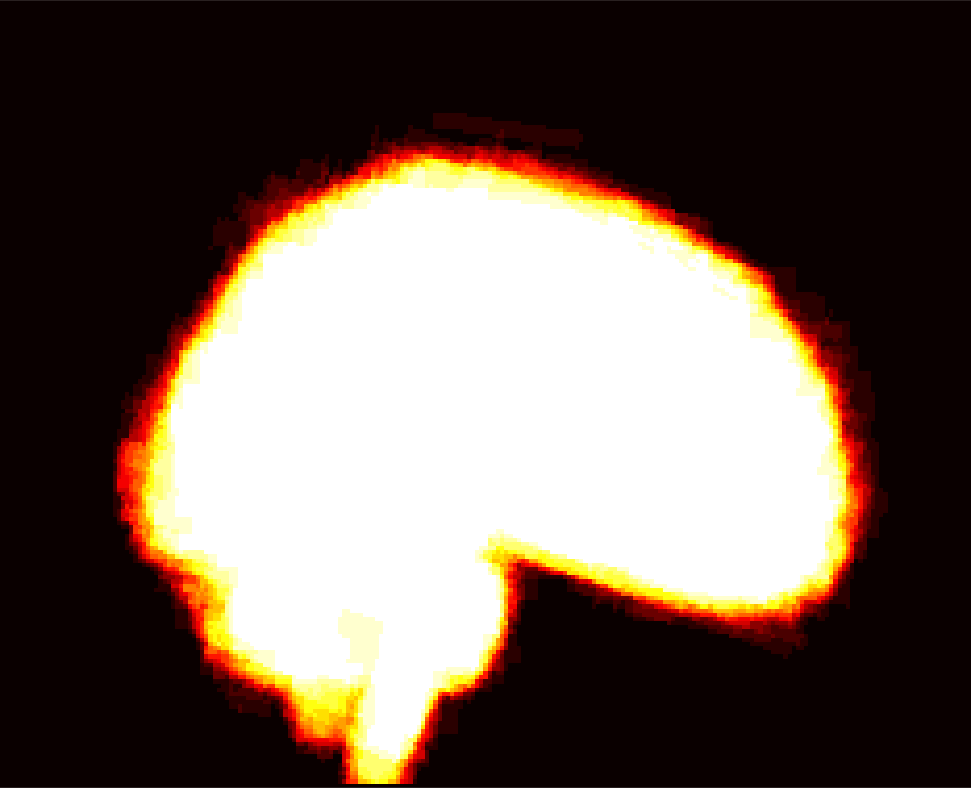} &
		\includegraphics[scale=0.1]{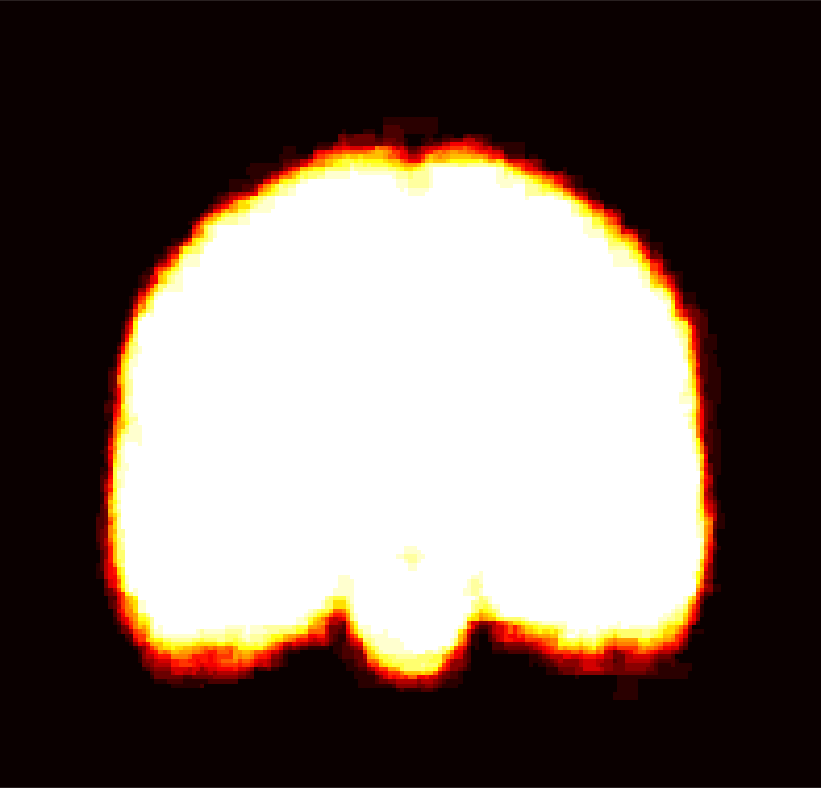} &
		\includegraphics[scale=0.08128]{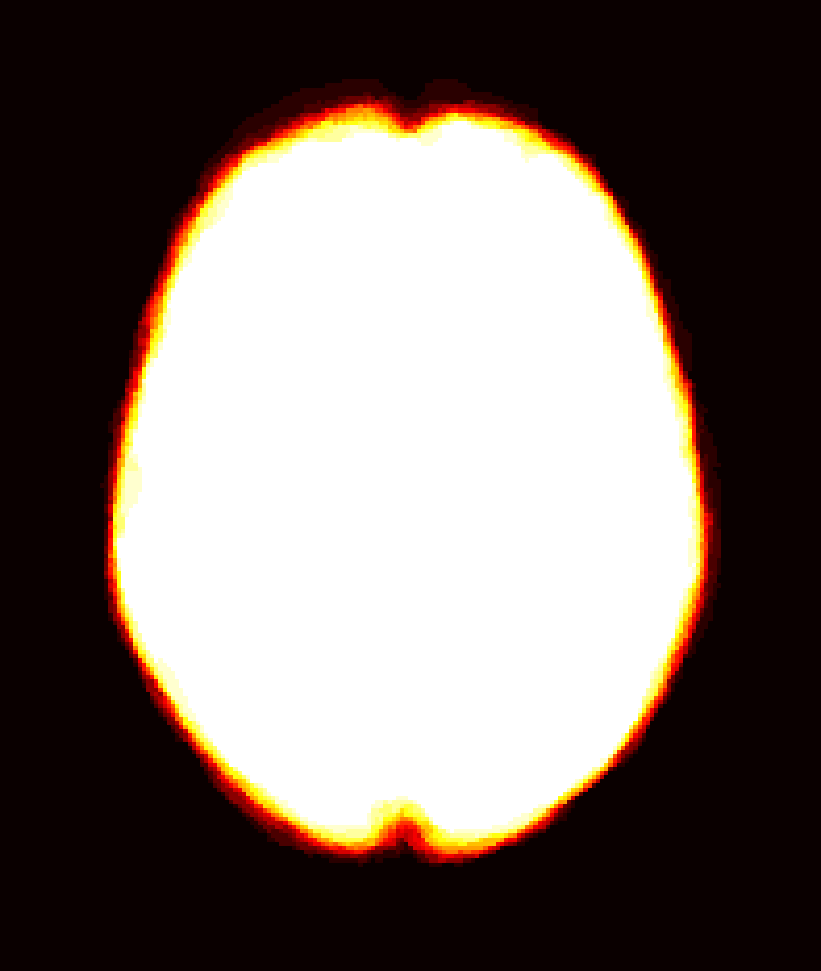}&\\[0.5ex]
		& \multicolumn{3}{c}{\textbf{LPBA40}} && \multicolumn{3}{c}{\textbf{TBI}}
		 \\[0.5ex]
		\textbf{PCA} &
		\includegraphics[scale=0.1]{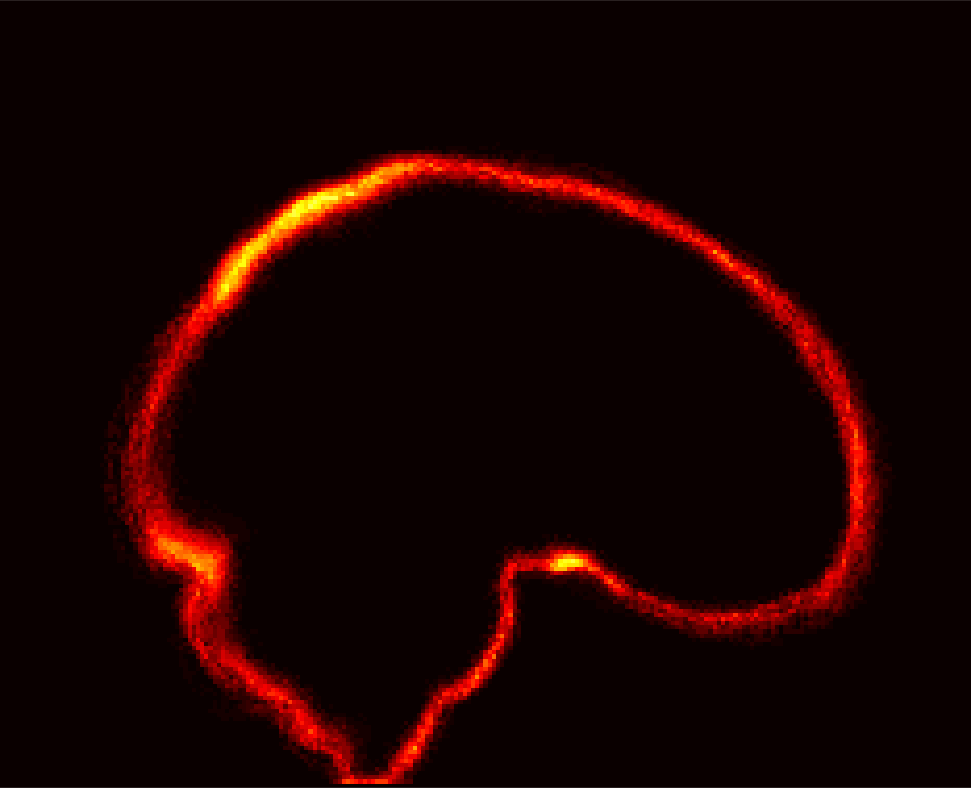} &
		\includegraphics[scale=0.1]{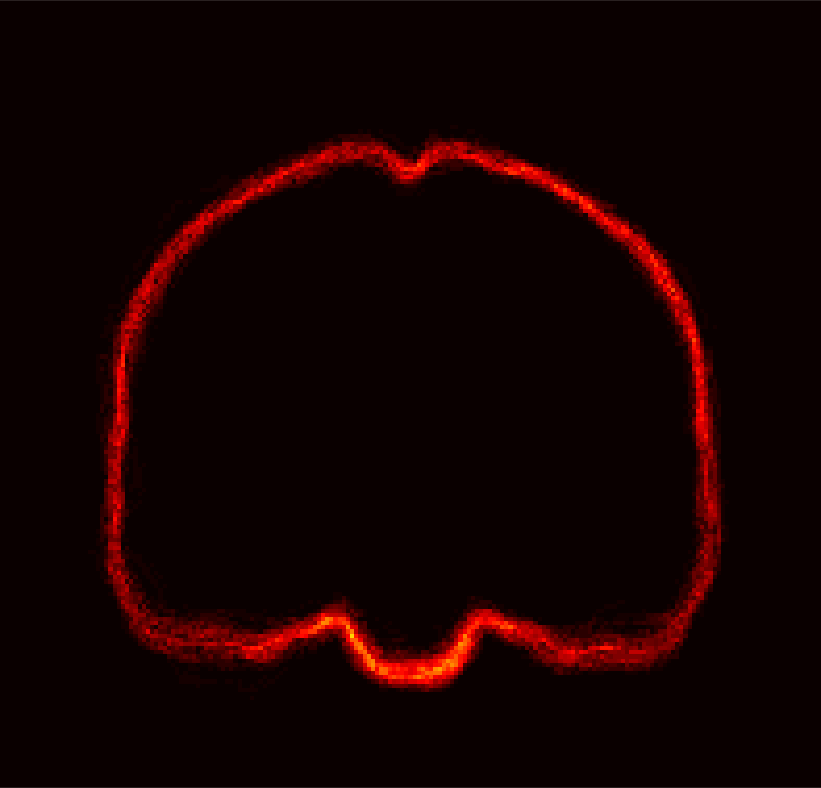} &
		\includegraphics[scale=0.08128]{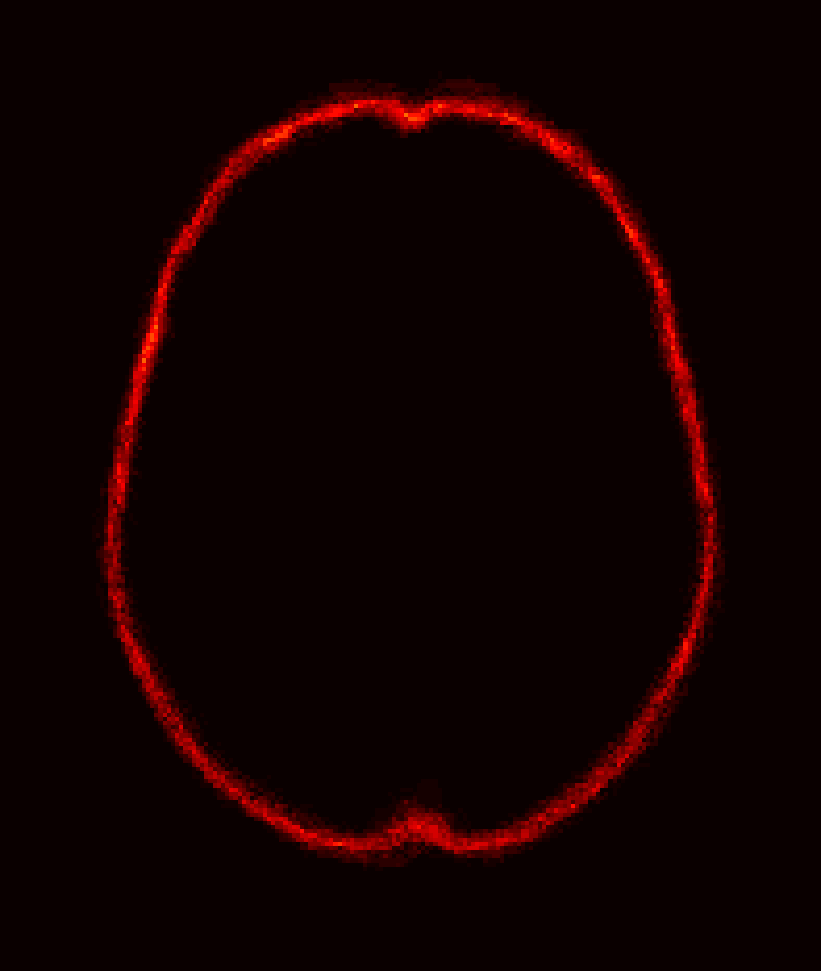} &&
		\includegraphics[scale=0.1]{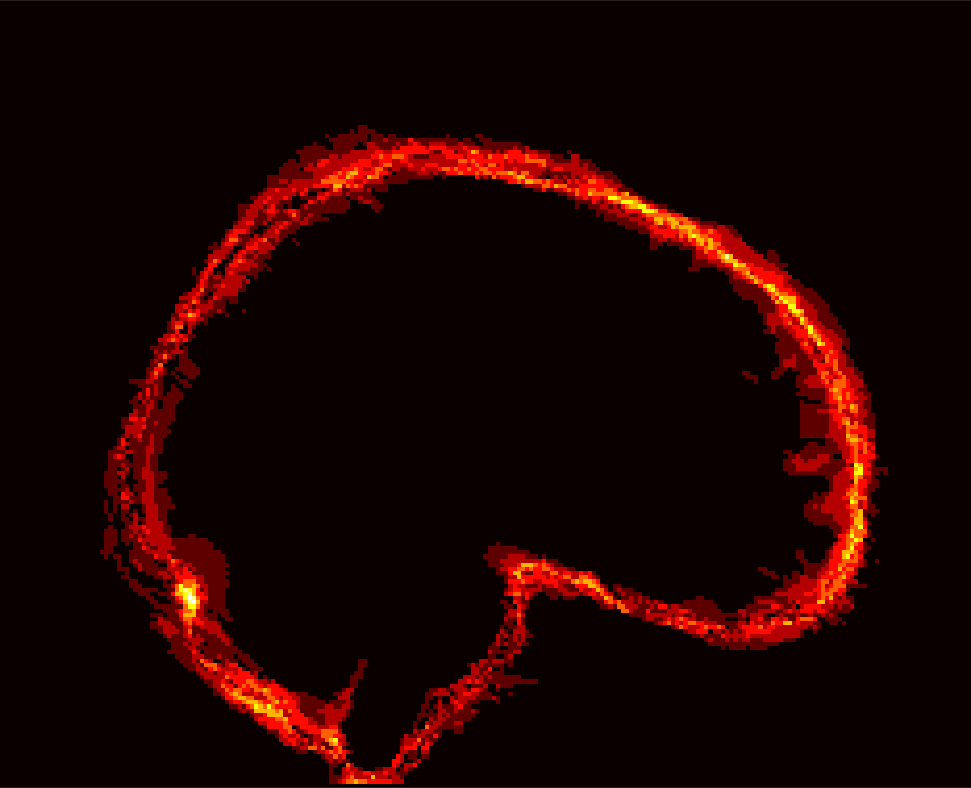} &
		\includegraphics[scale=0.1]{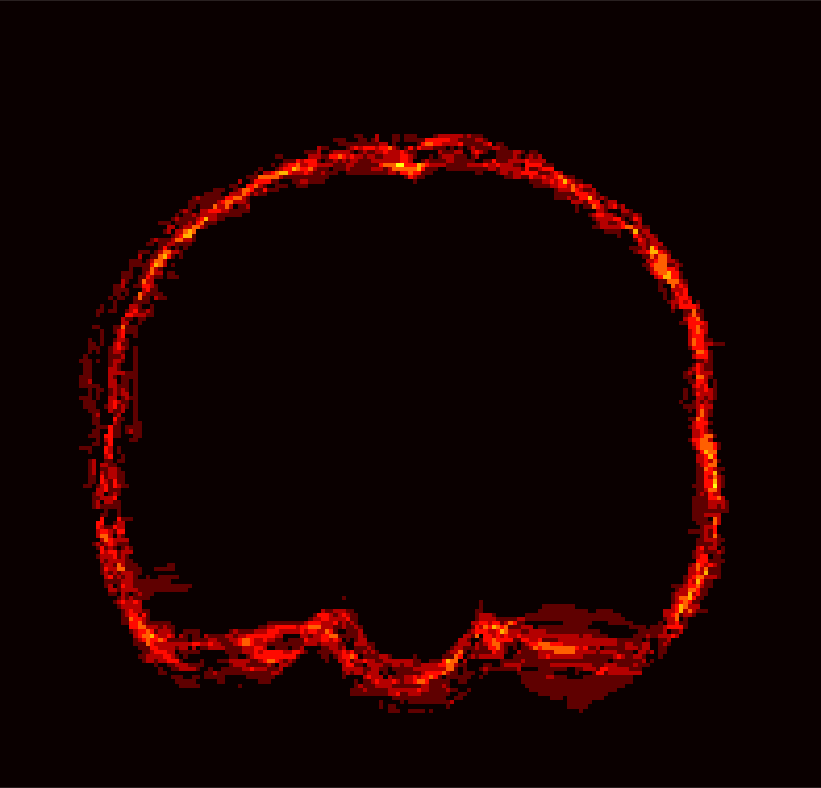} &
		\includegraphics[scale=0.08128]{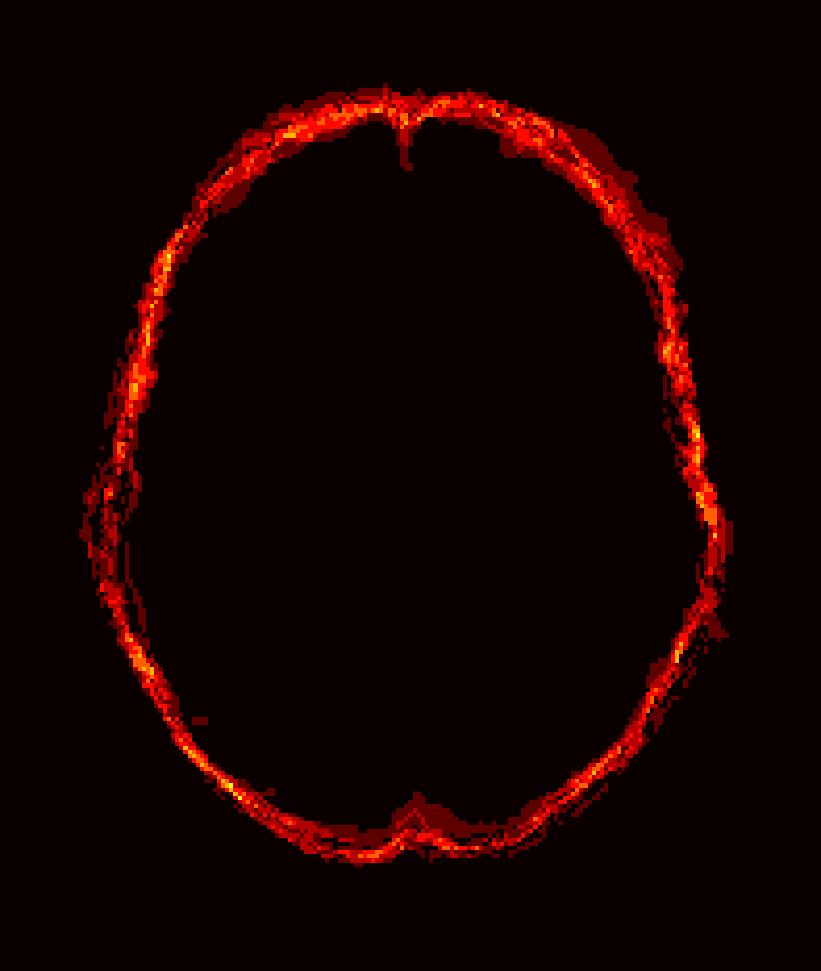}&
		\\
		\textbf{ROBEX} &
		\includegraphics[scale=0.1]{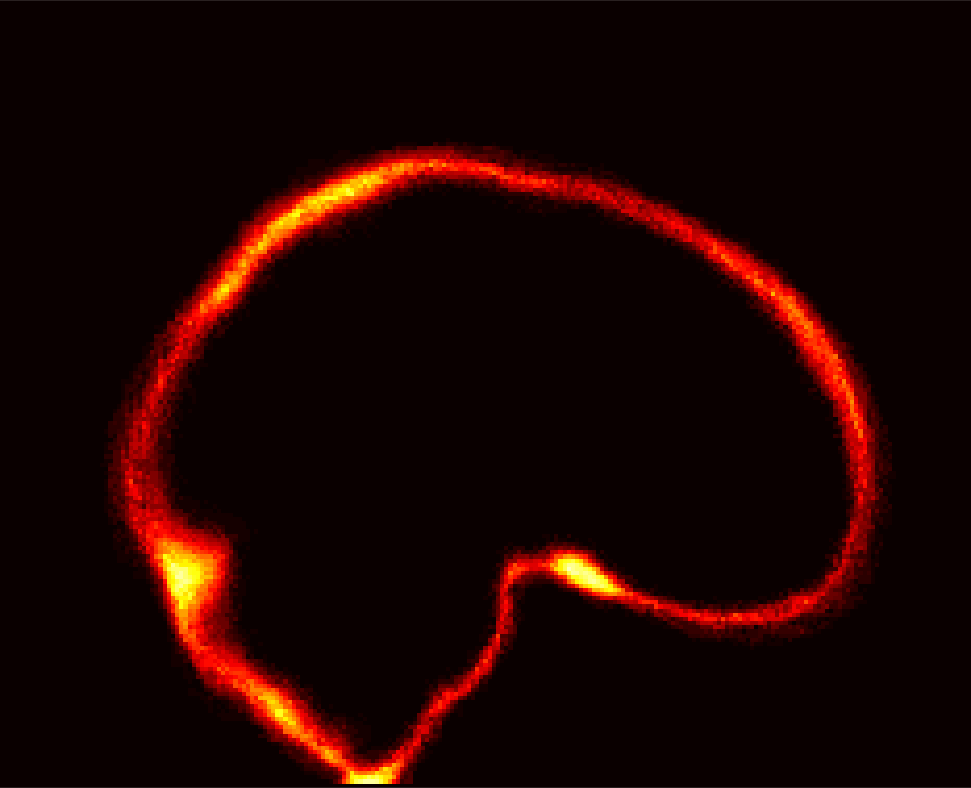} &
		\includegraphics[scale=0.1]{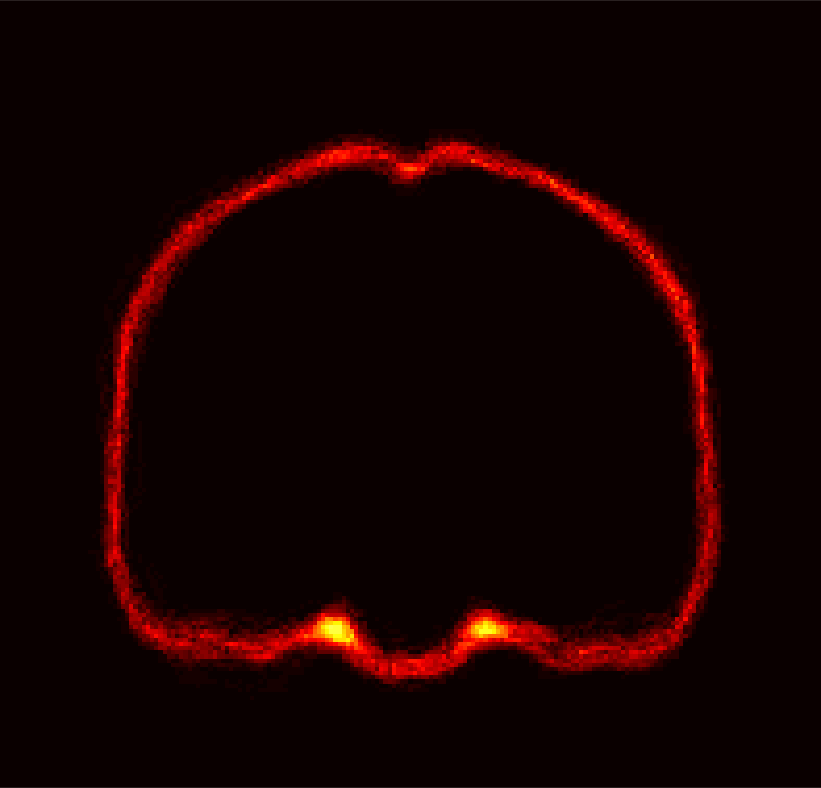} &
		\includegraphics[scale=0.08128]{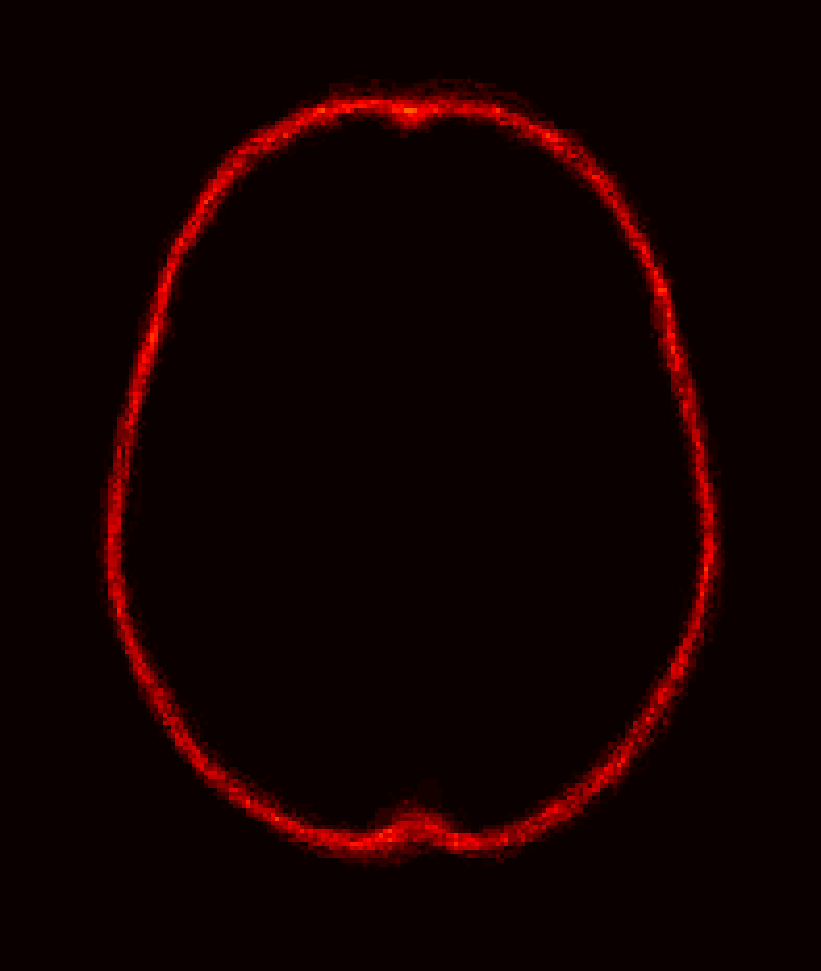} &&
		\includegraphics[scale=0.1]{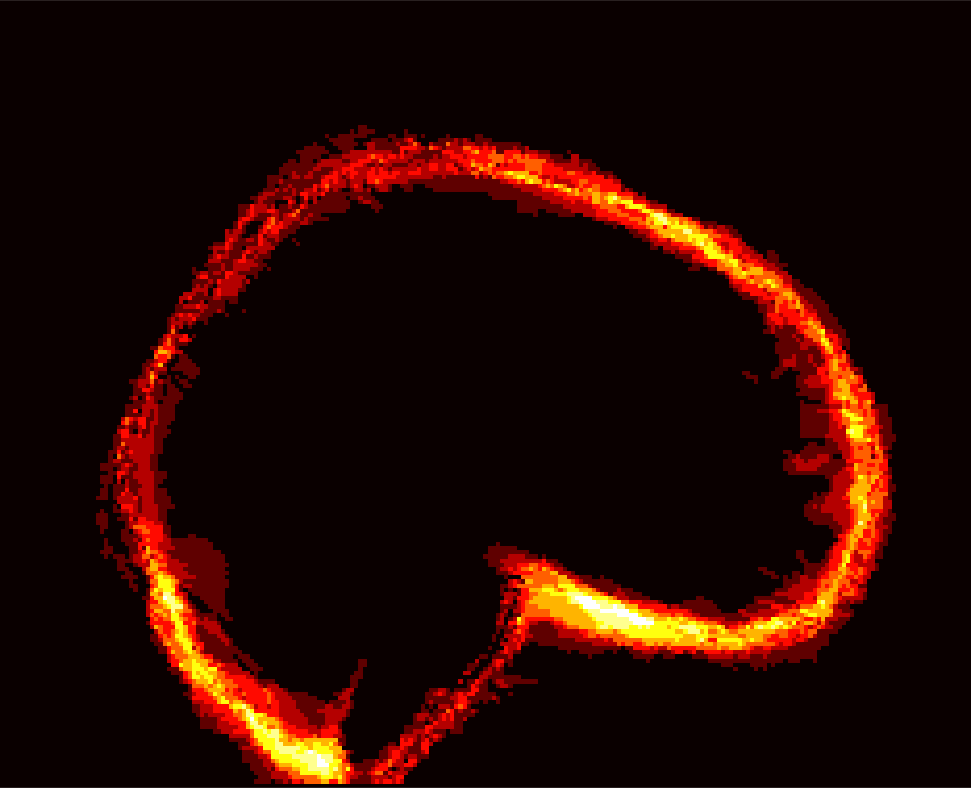} &
		\includegraphics[scale=0.1]{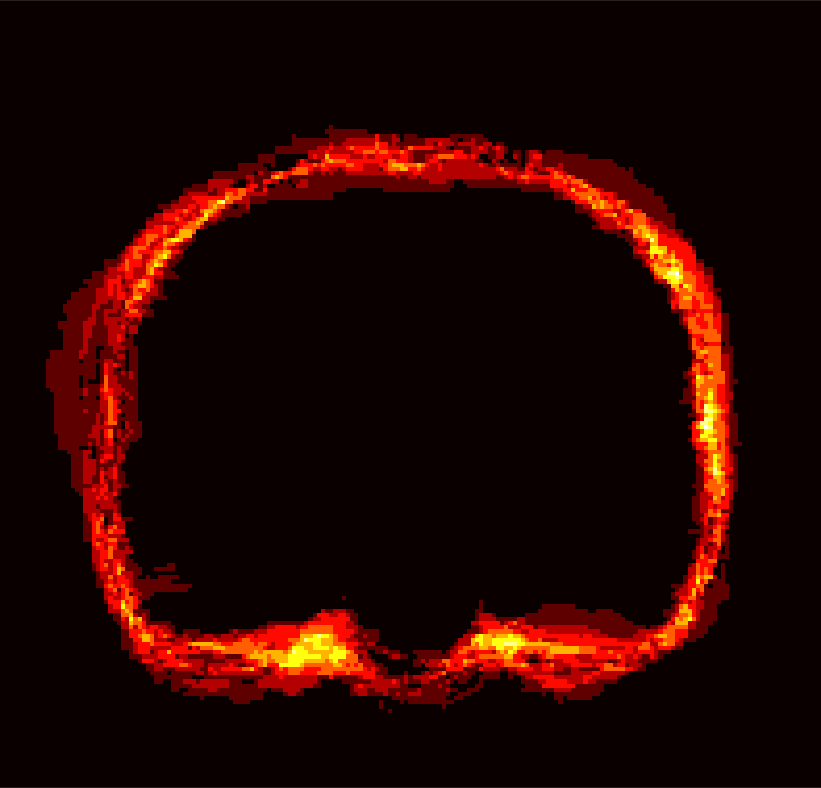} &
		\includegraphics[scale=0.08128]{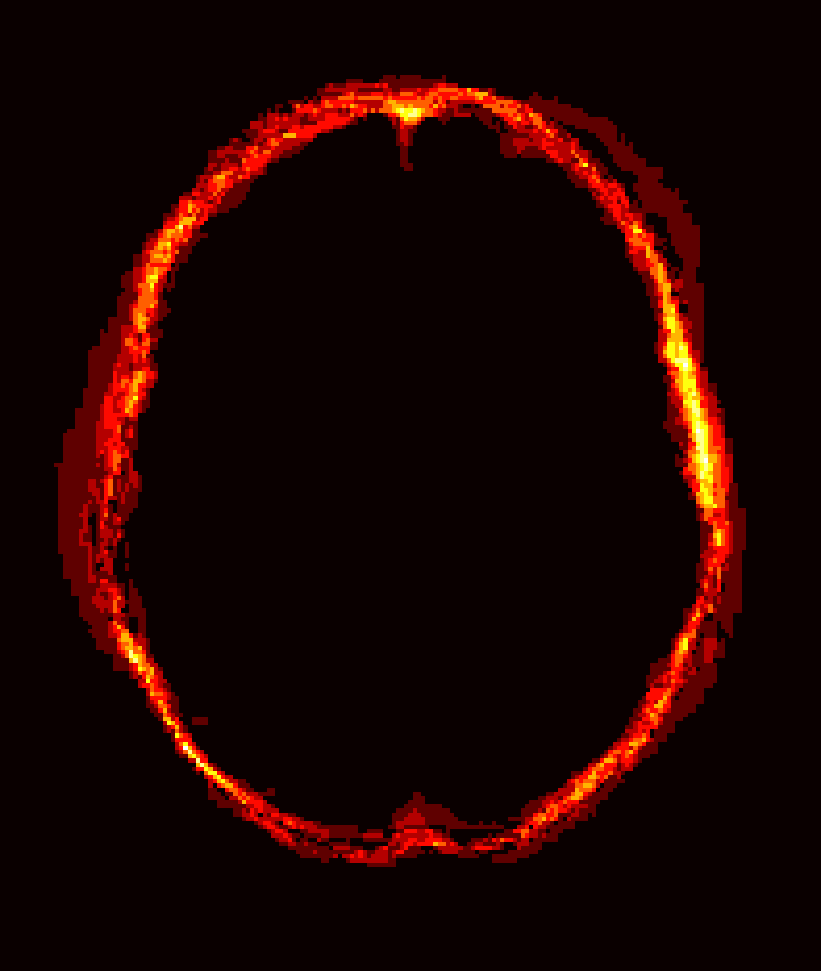} &\\
		\textbf{BEaST} &
		\includegraphics[scale=0.1]{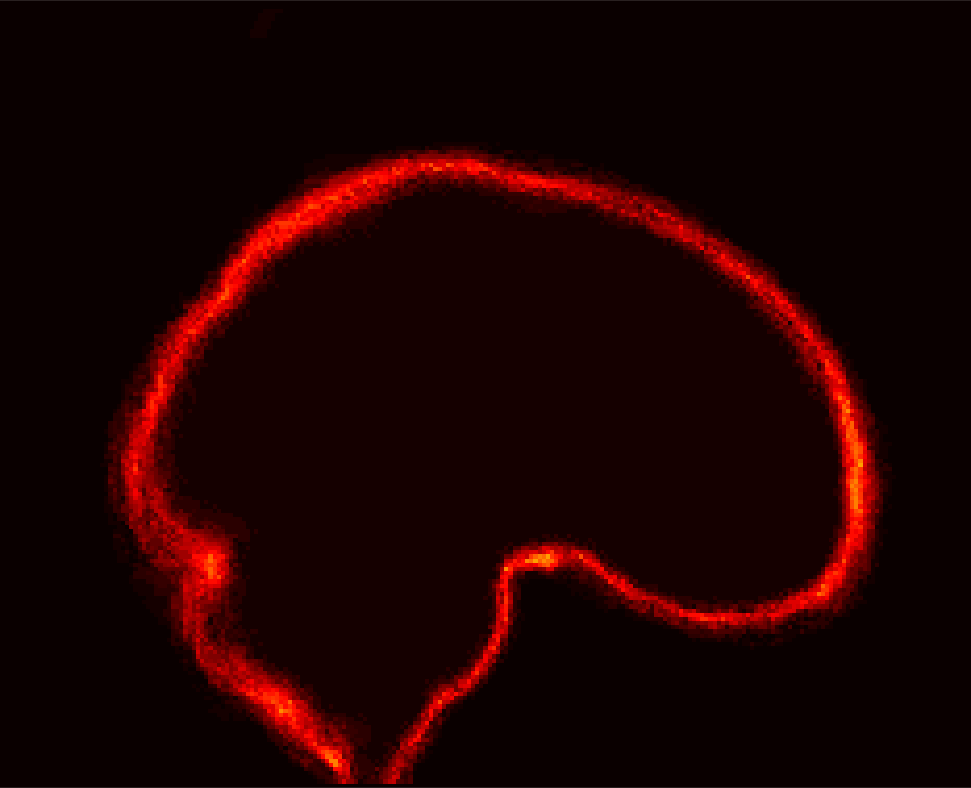} &
		\includegraphics[scale=0.1]{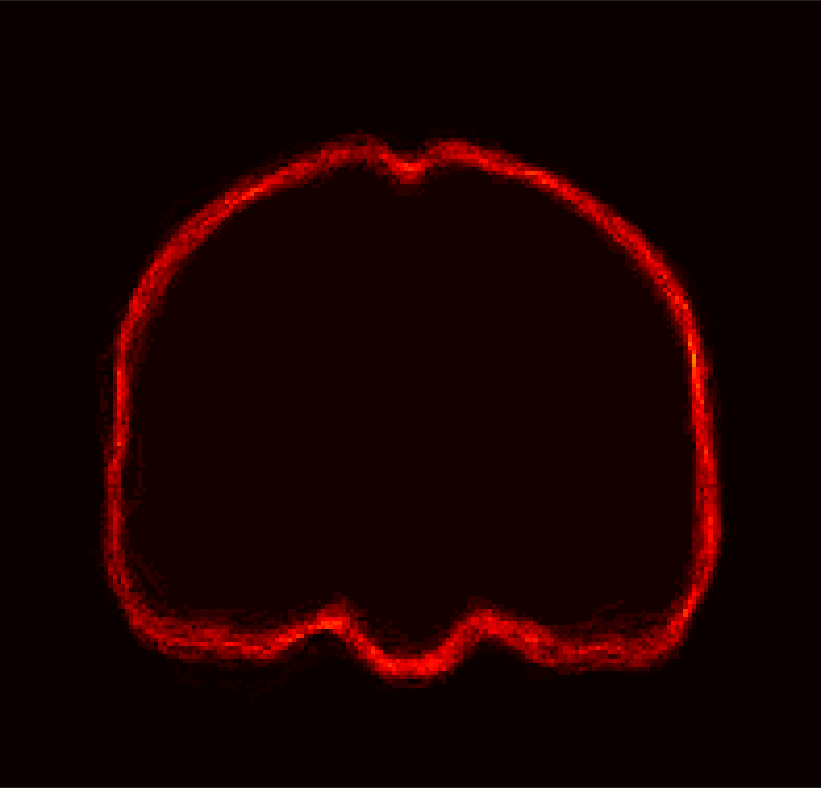} &
		\includegraphics[scale=0.08128]{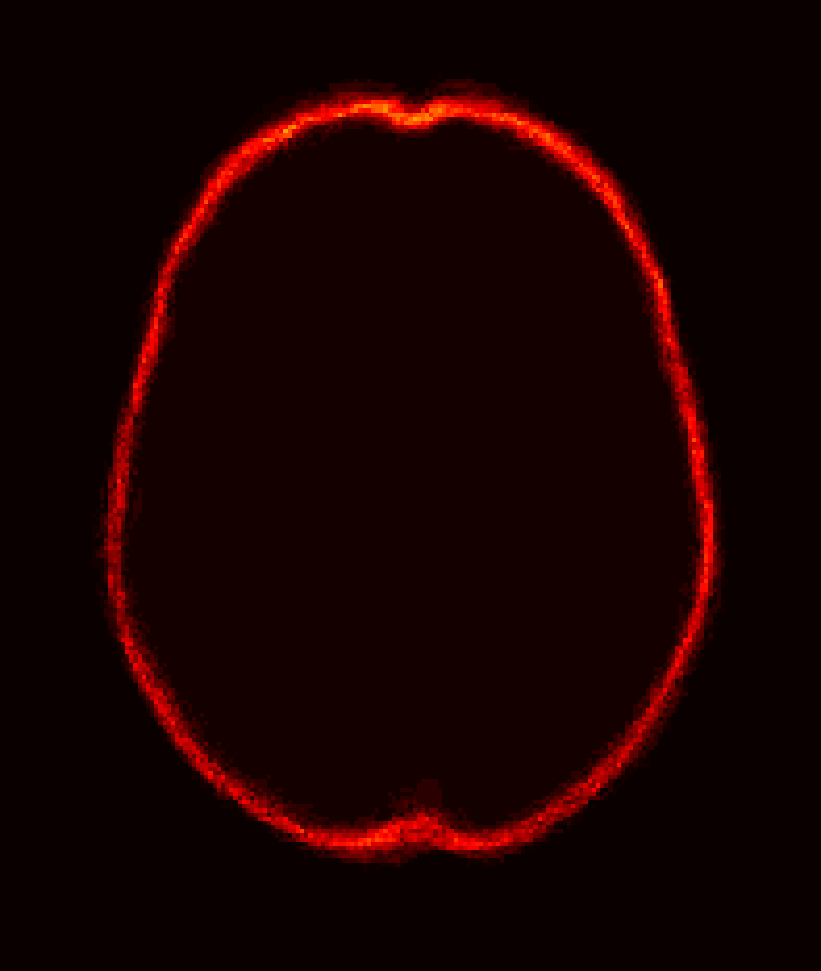} &&
		\includegraphics[scale=0.1]{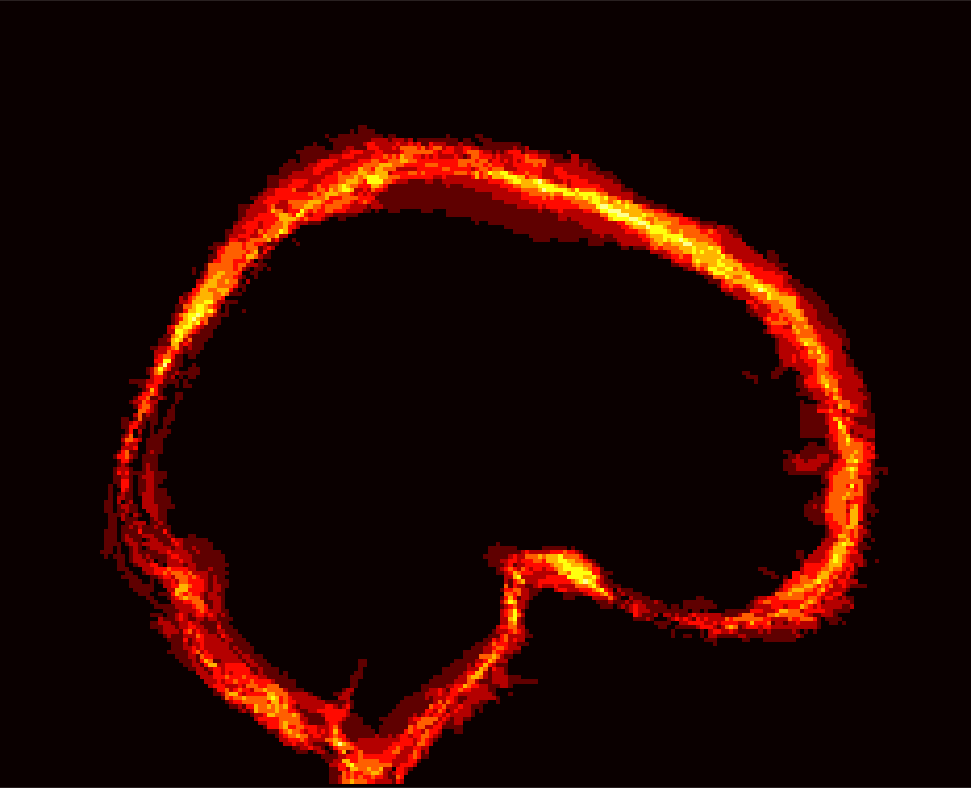} &
		\includegraphics[scale=0.1]{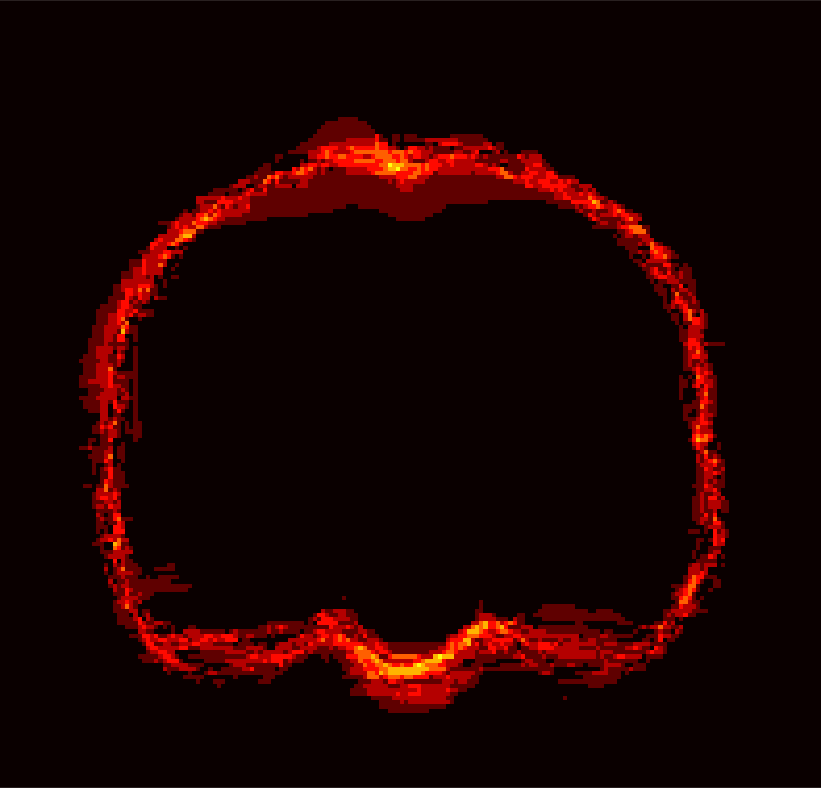} &
		\includegraphics[scale=0.08128]{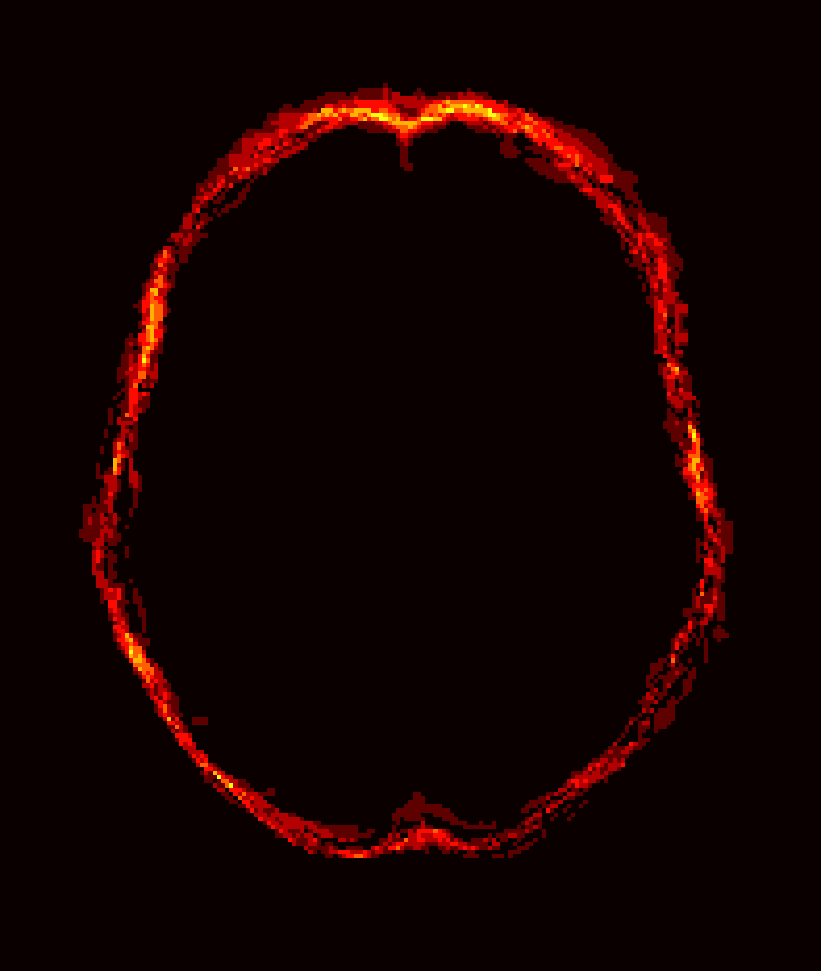} &\\
		\textbf{MASS} &
		\includegraphics[scale=0.1]{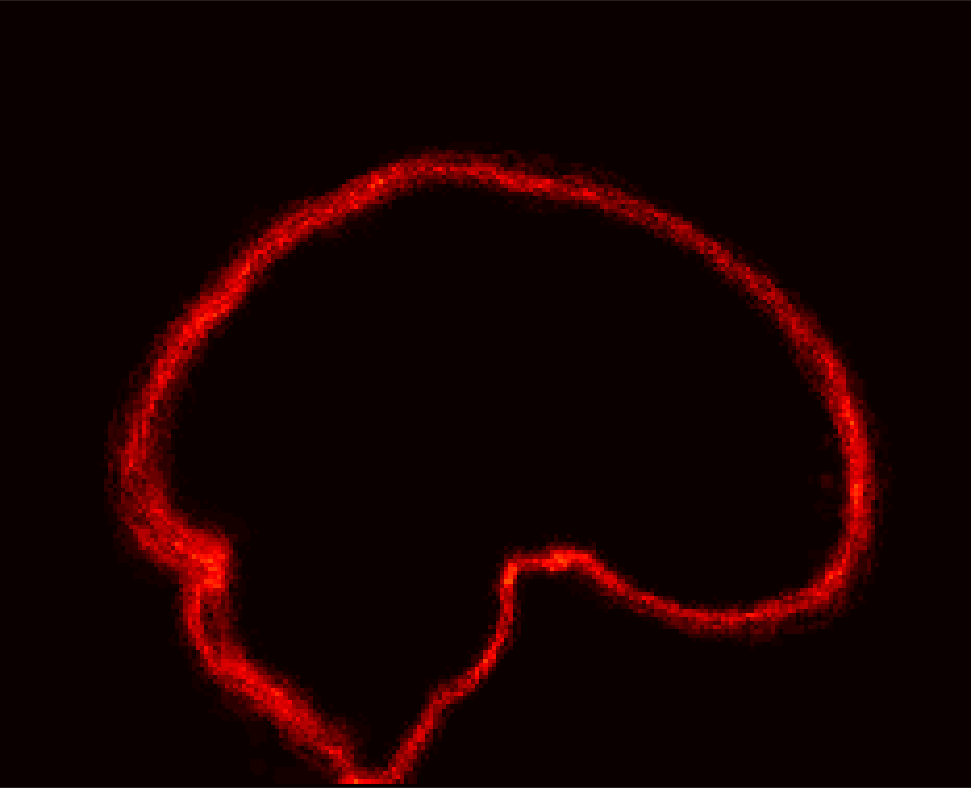} &
		\includegraphics[scale=0.1]{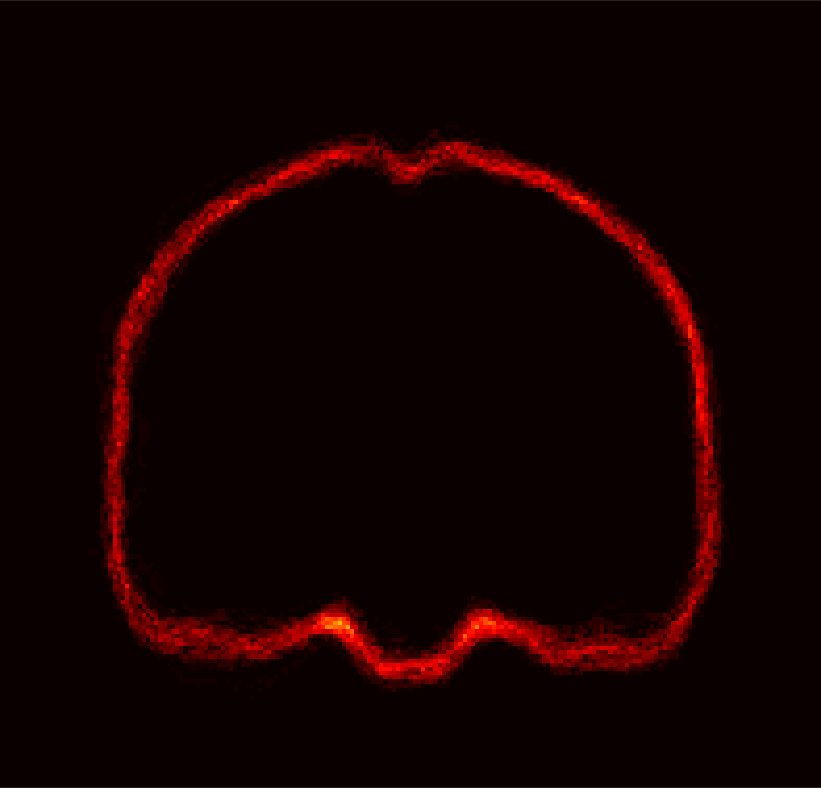} &
		\includegraphics[scale=0.08128]{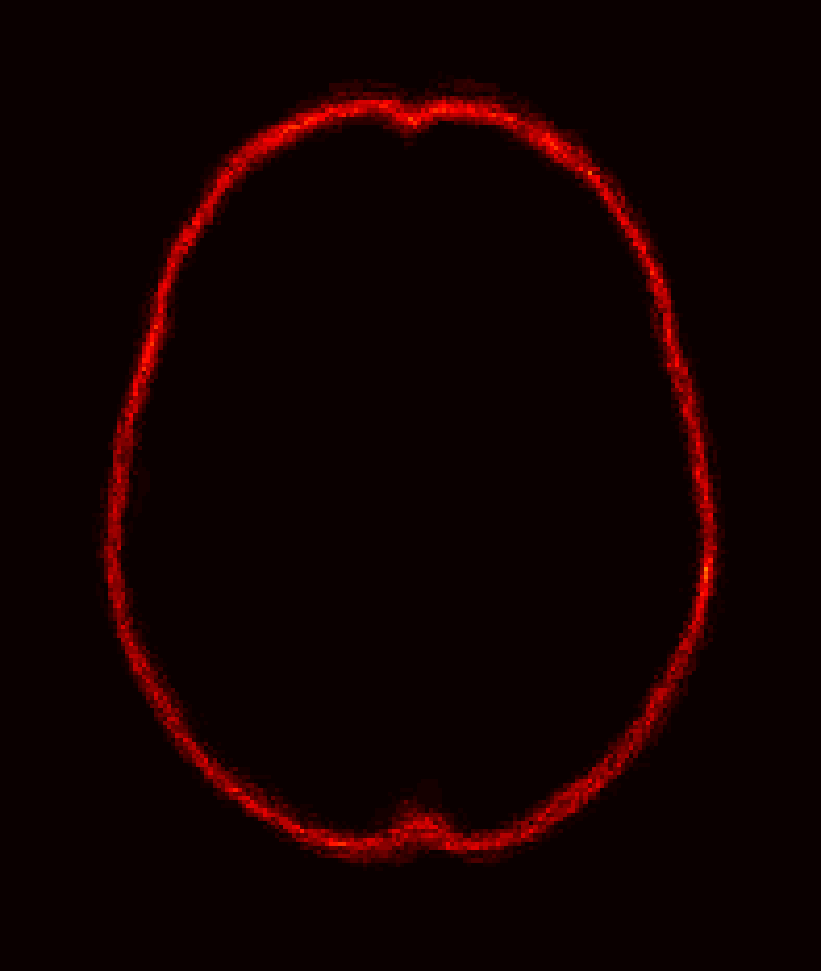} &&
		\includegraphics[scale=0.1]{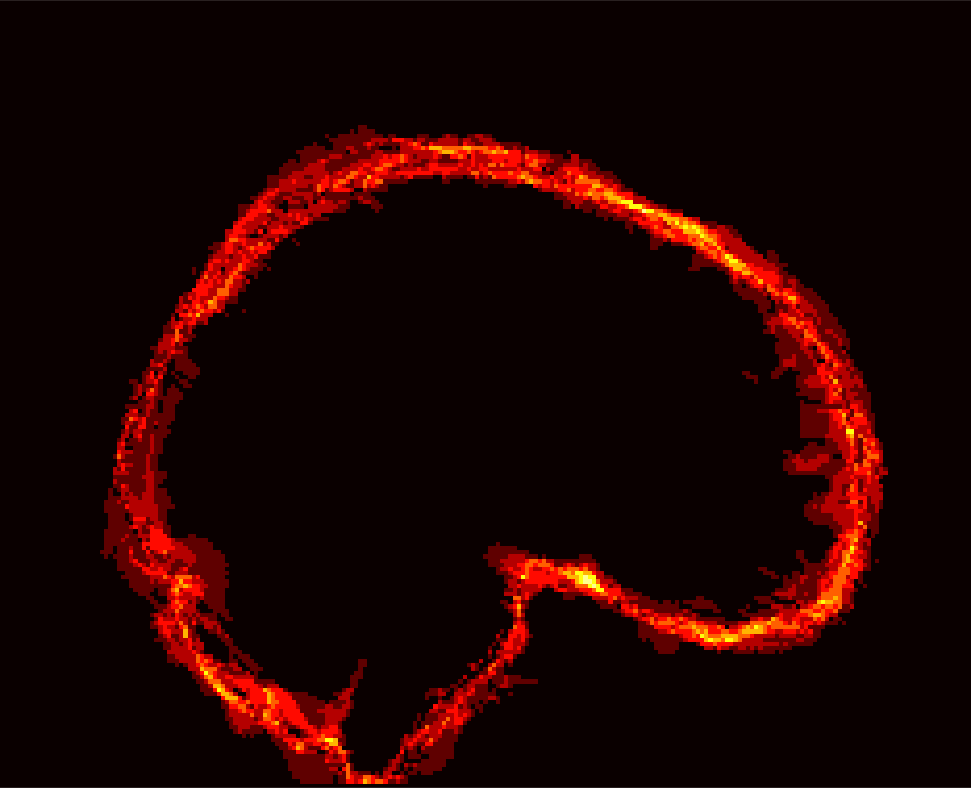} &
		\includegraphics[scale=0.1]{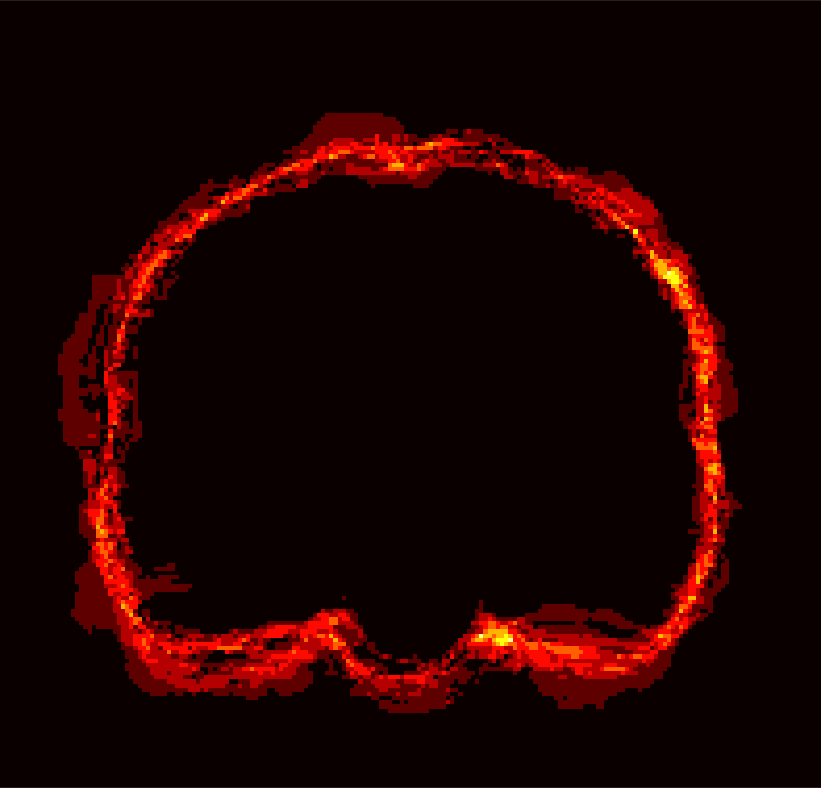} &
		\includegraphics[scale=0.08128]{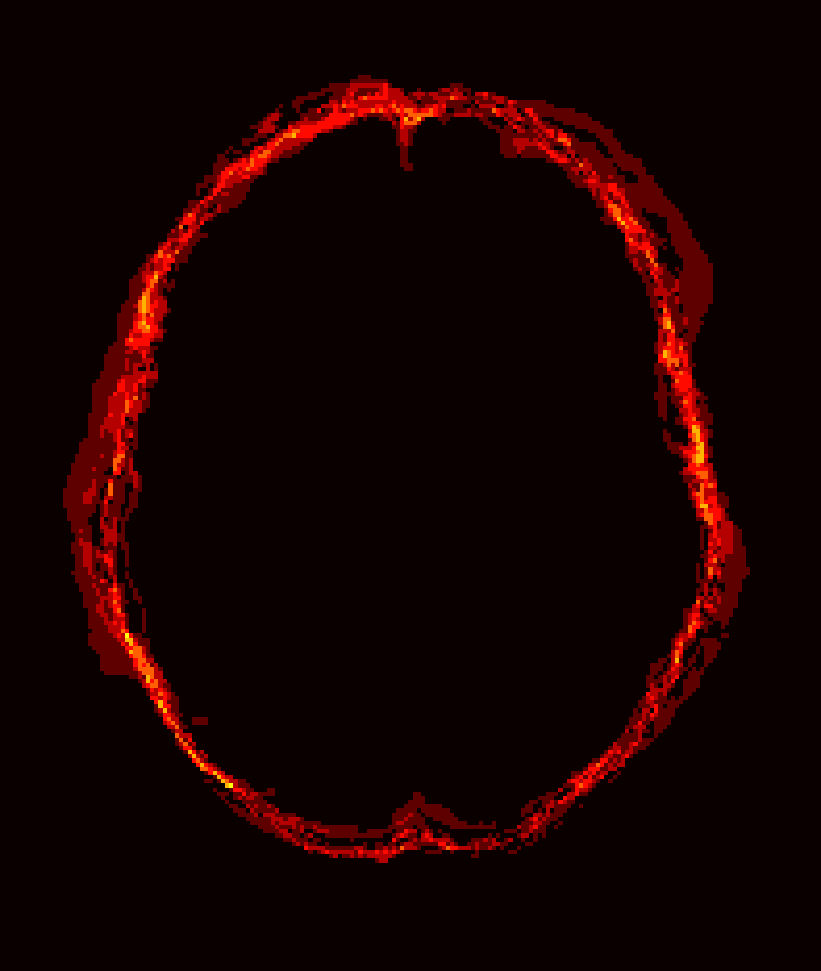} &\\		
		\textbf{BET} &
		\includegraphics[scale=0.1]{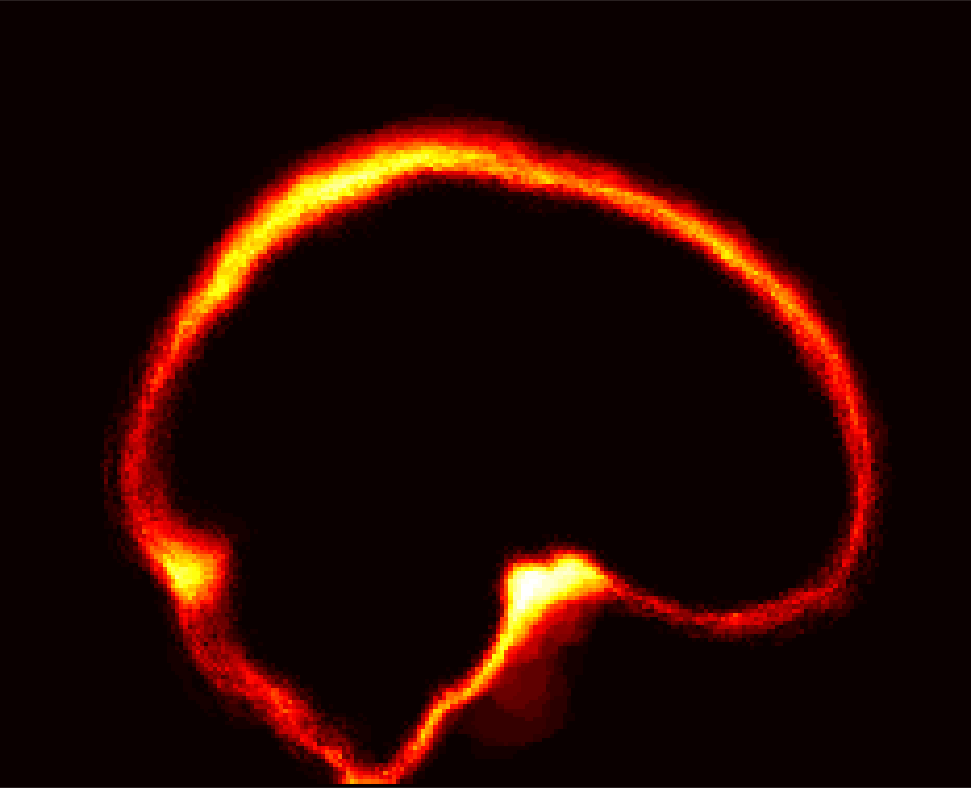} &
		\includegraphics[scale=0.1]{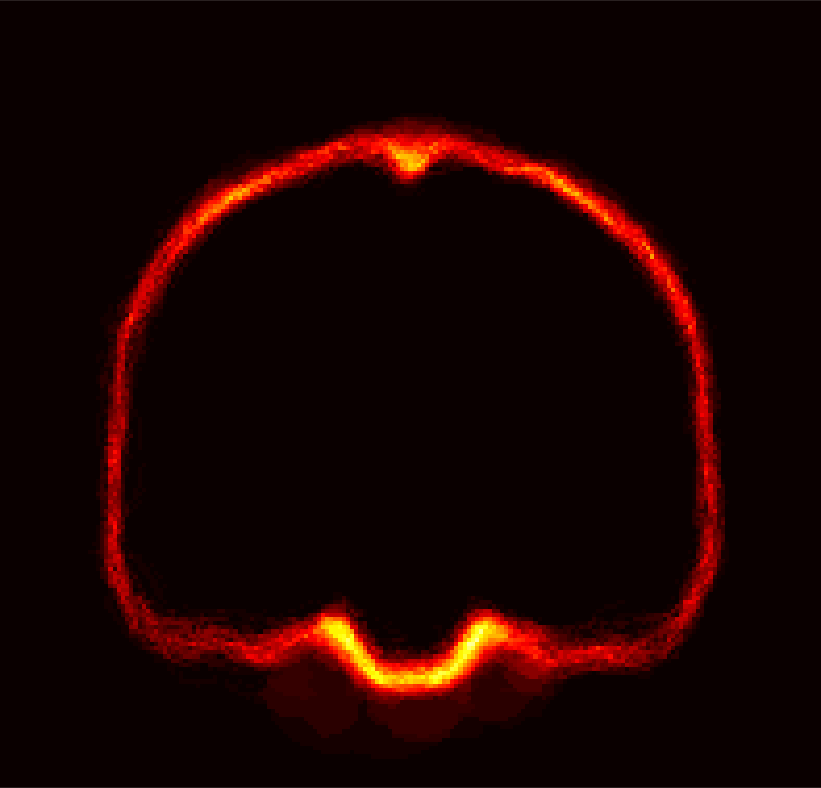} &
		\includegraphics[scale=0.08128]{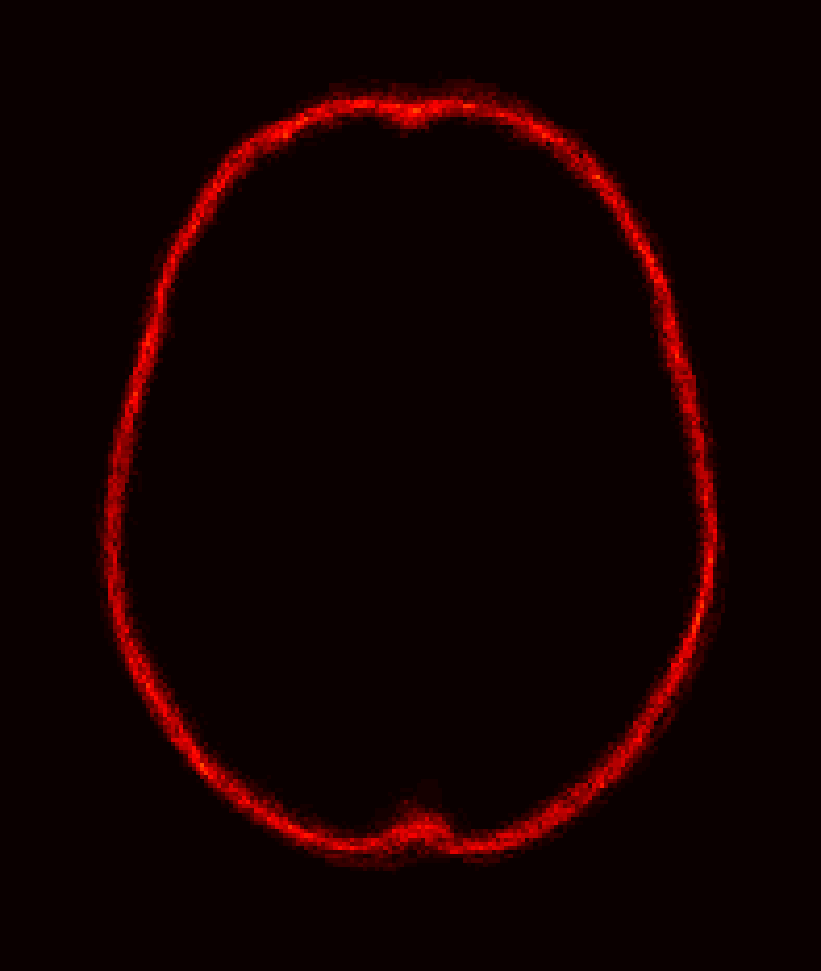} &&
		\includegraphics[scale=0.1]{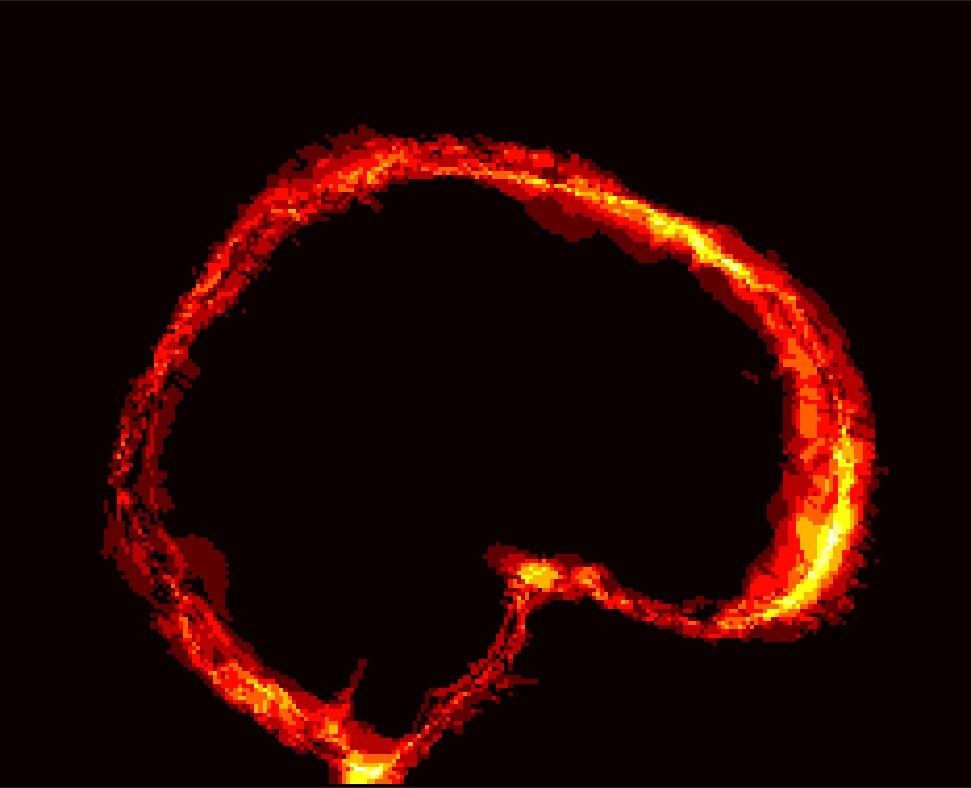} &
		\includegraphics[scale=0.1]{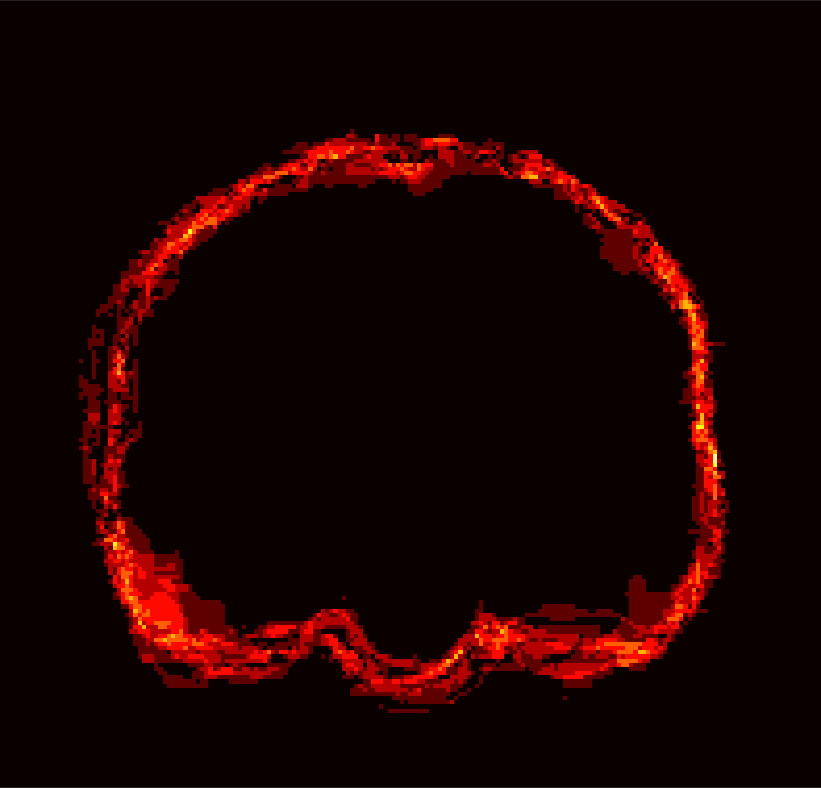} &
		\includegraphics[scale=0.08128]{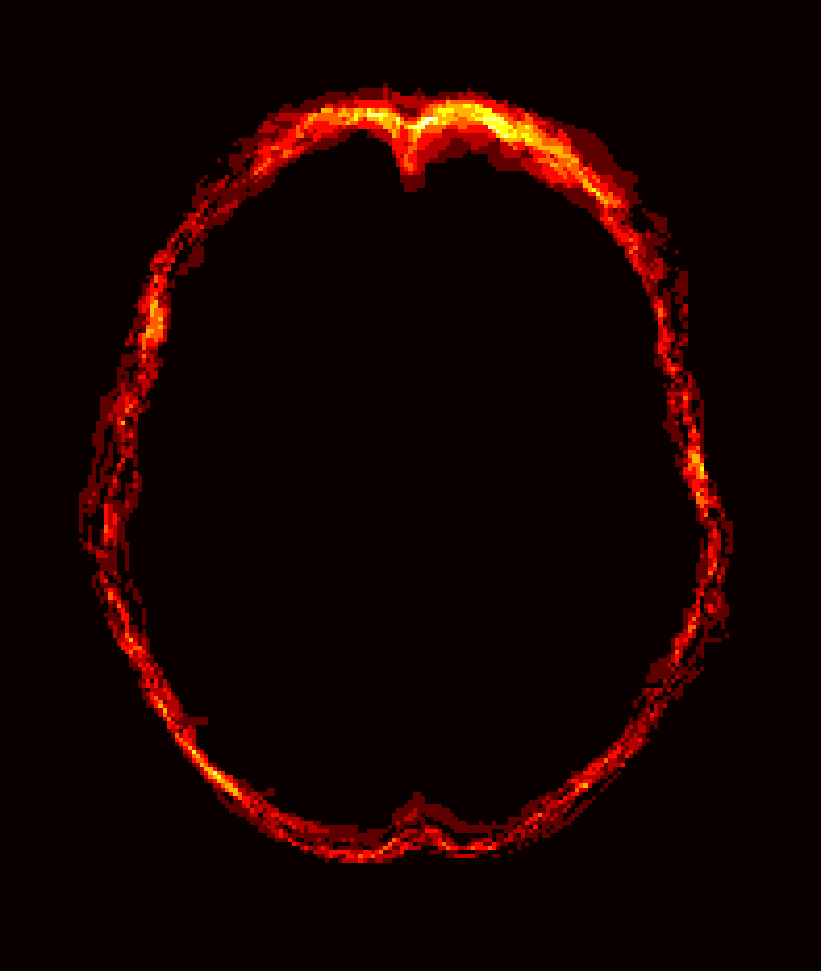} &\\
		\textbf{BSE} &
		\includegraphics[scale=0.1]{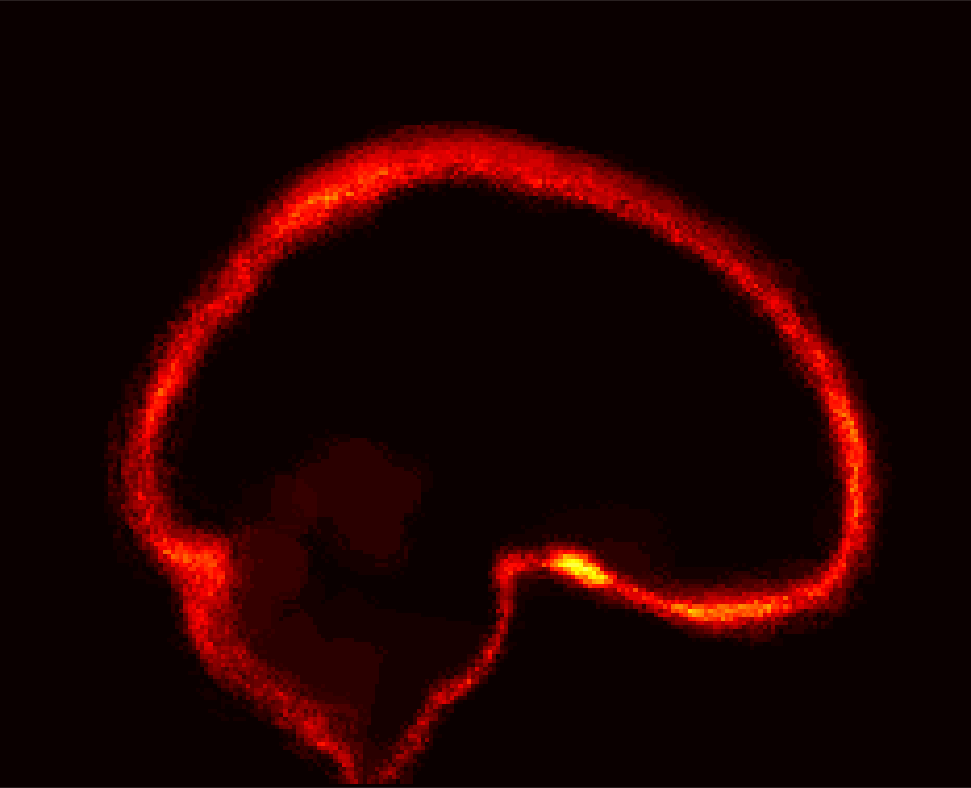} &
		\includegraphics[scale=0.1]{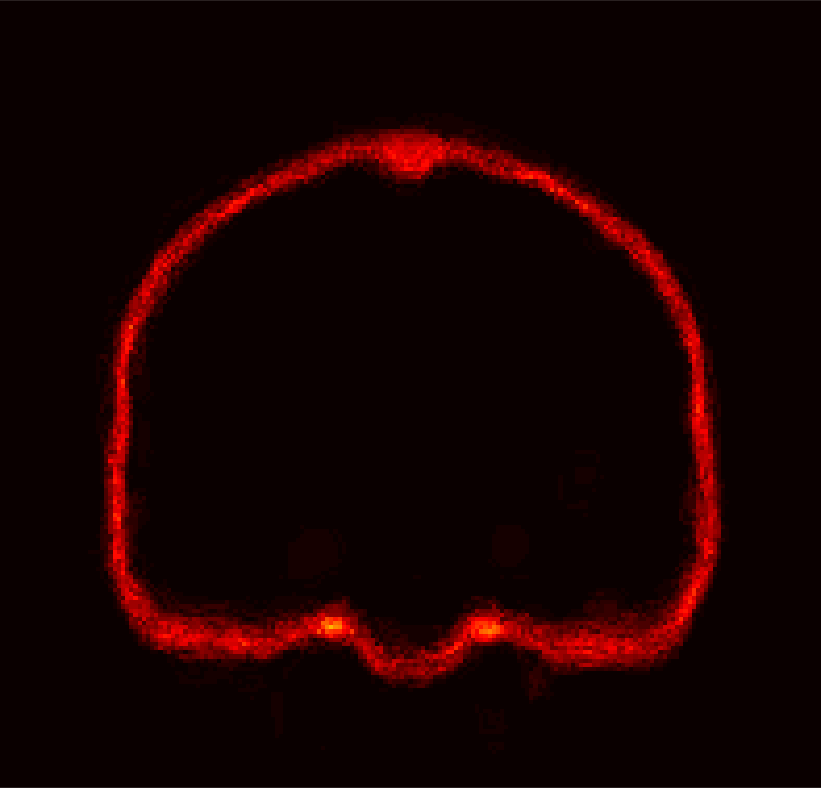} &
		\includegraphics[scale=0.08128]{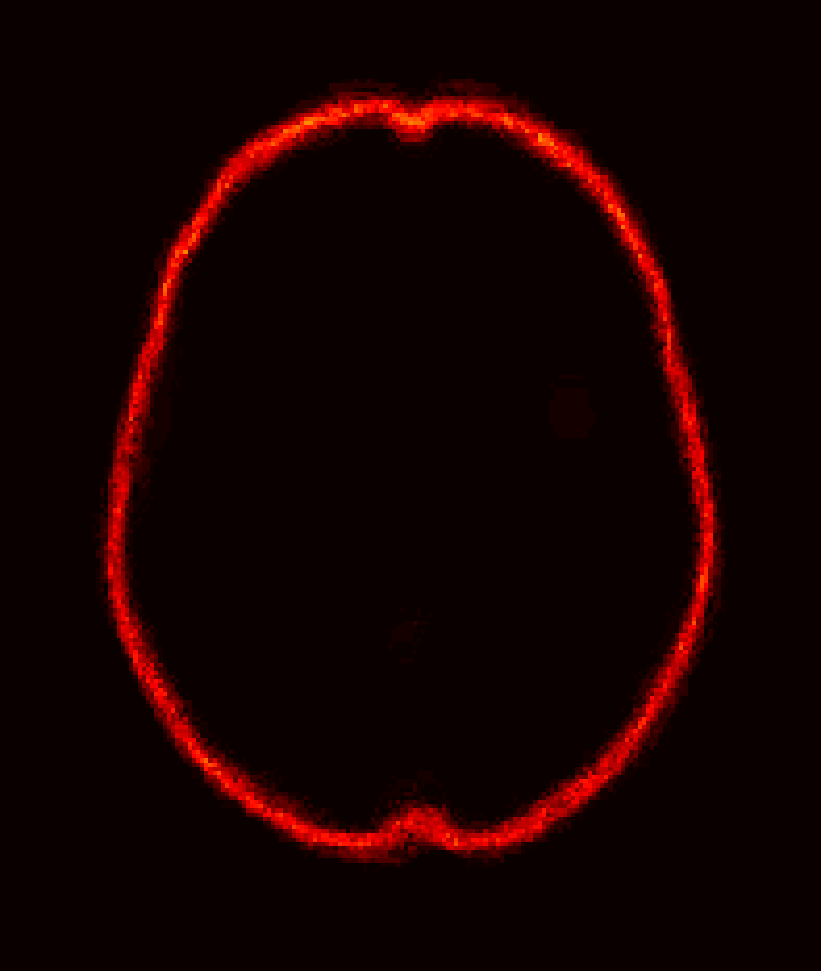} &&
		\includegraphics[scale=0.1]{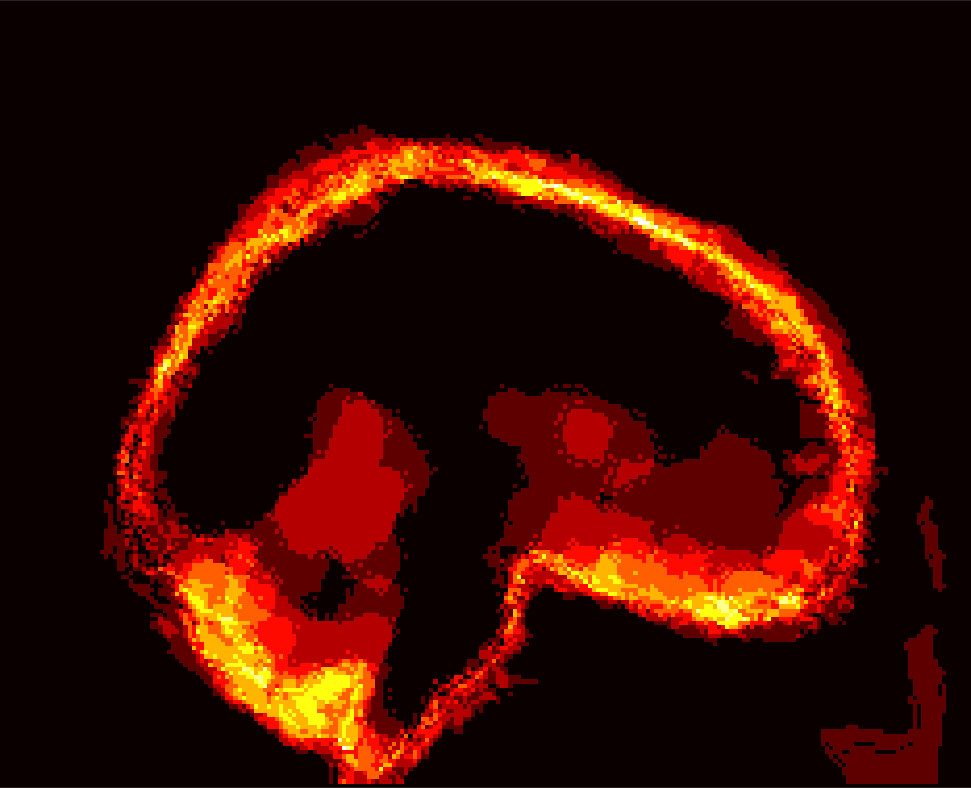} &
		\includegraphics[scale=0.1]{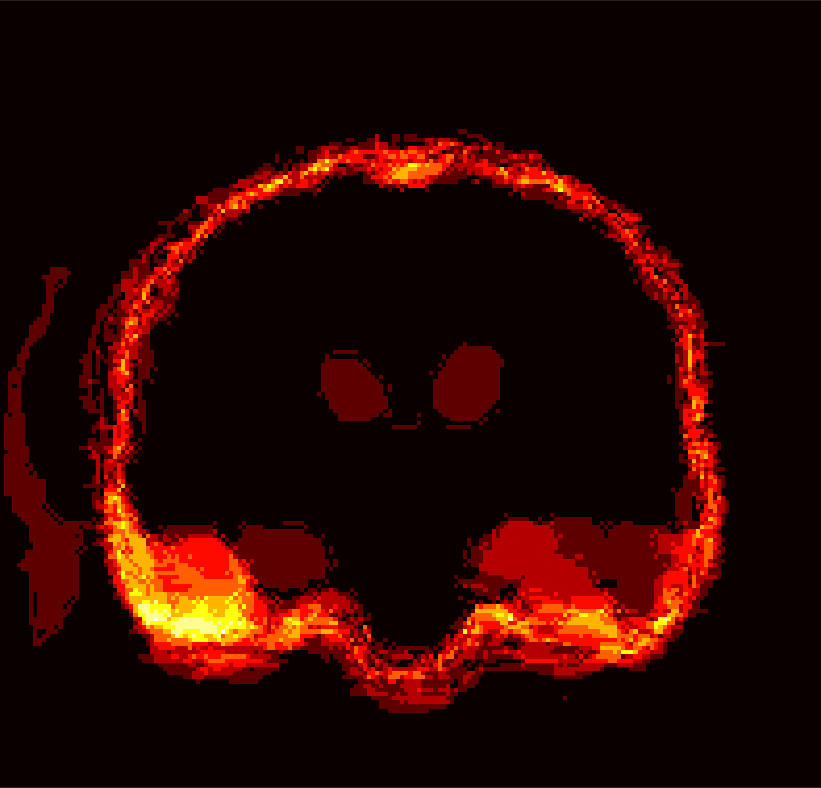} &
		\includegraphics[scale=0.08128]{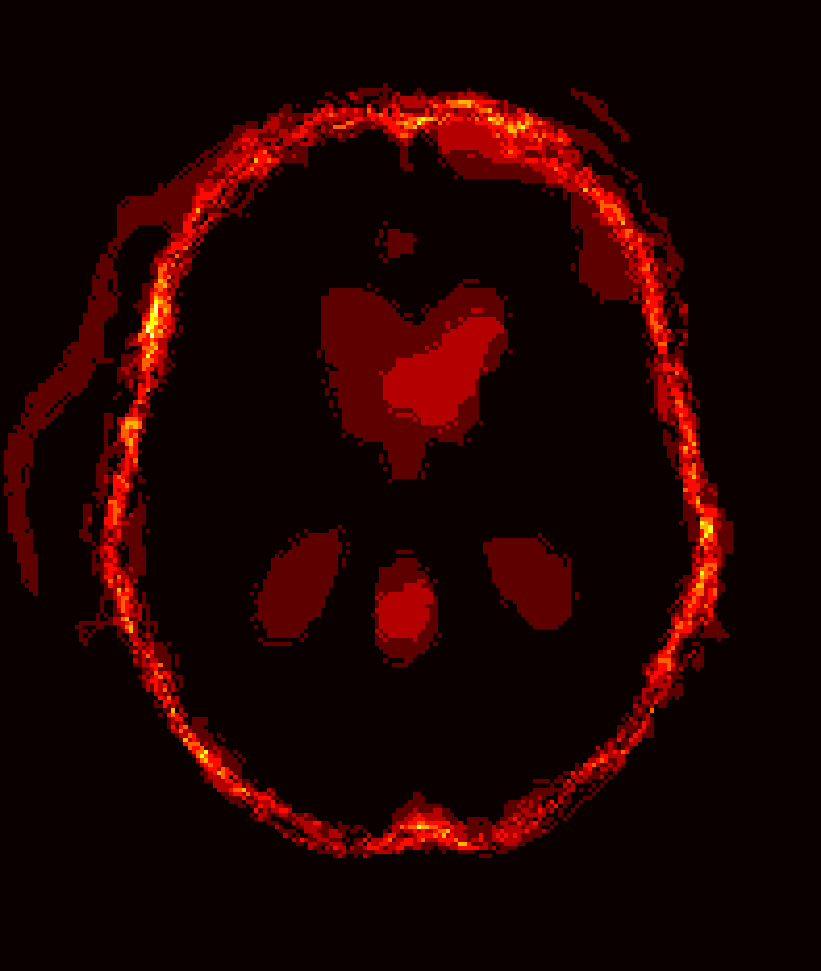} &\\
		\textbf{CNN} &
		\includegraphics[scale=0.1]{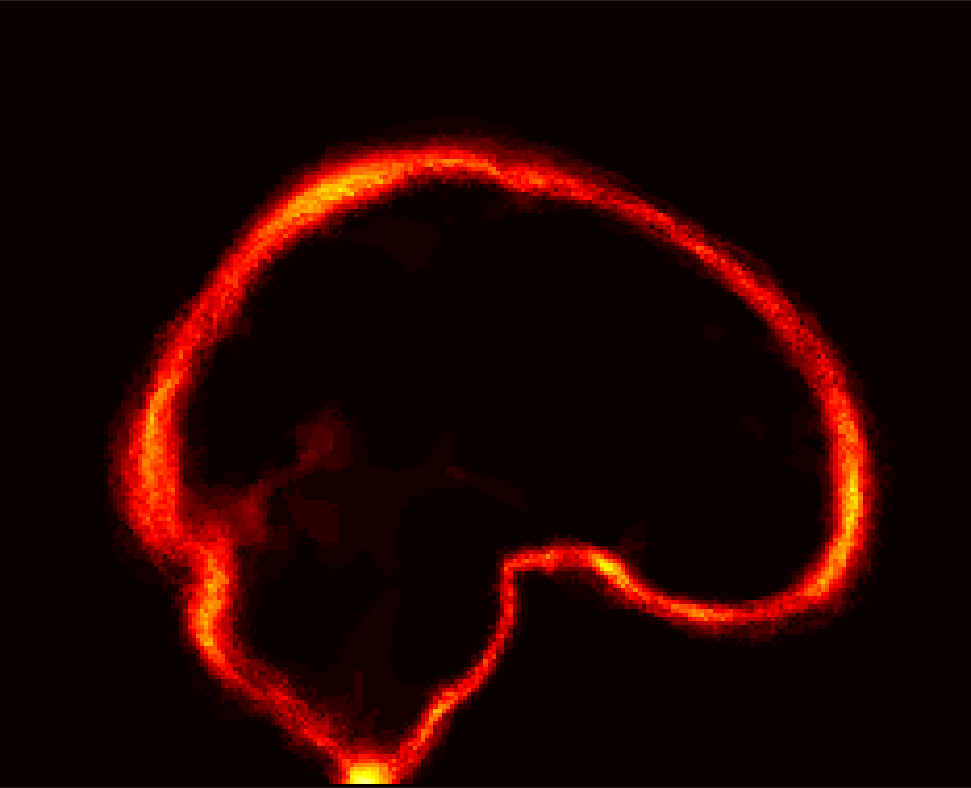} &
		\includegraphics[scale=0.1]{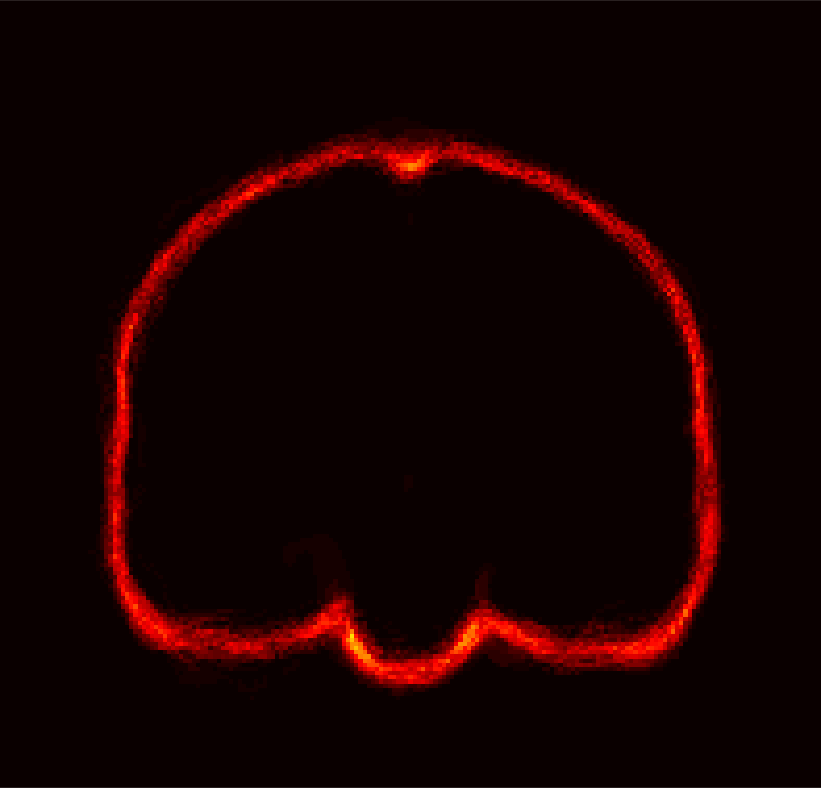} &
		\includegraphics[scale=0.08128]{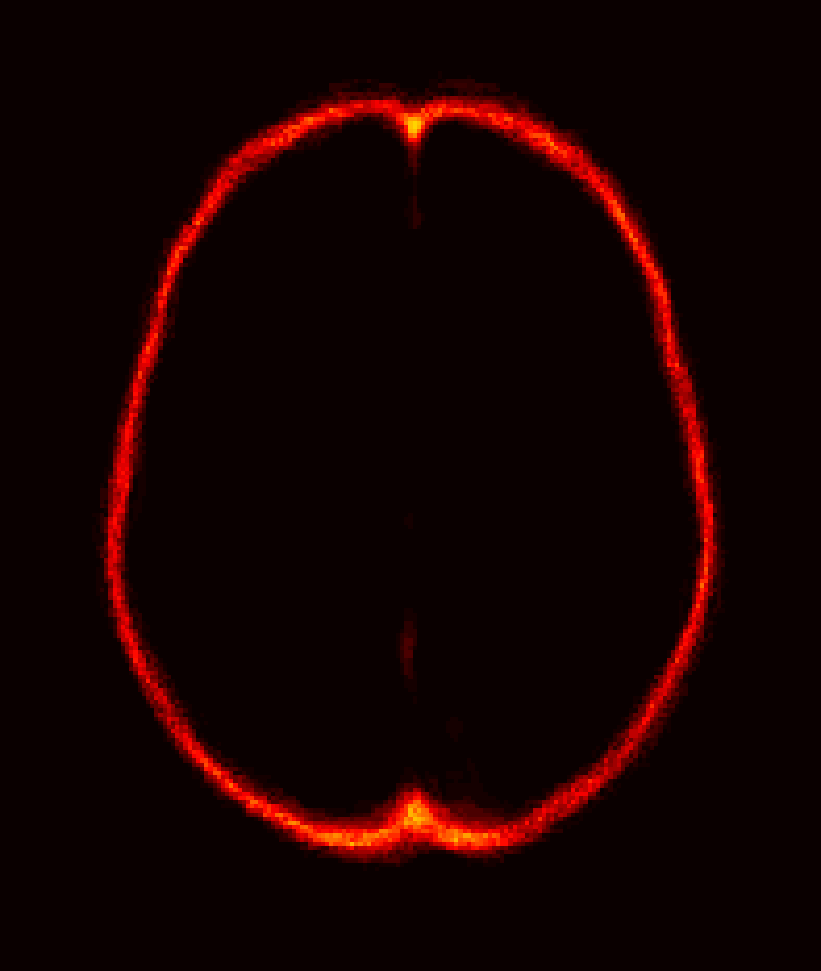} &&
		\includegraphics[scale=0.1]{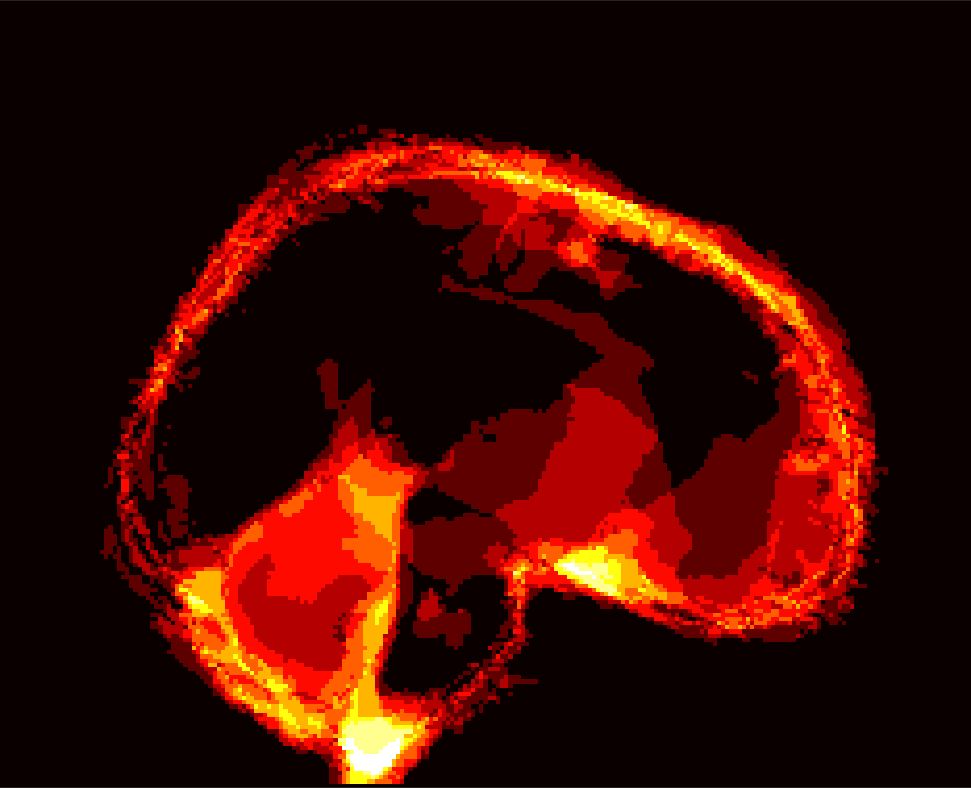} &
		\includegraphics[scale=0.1]{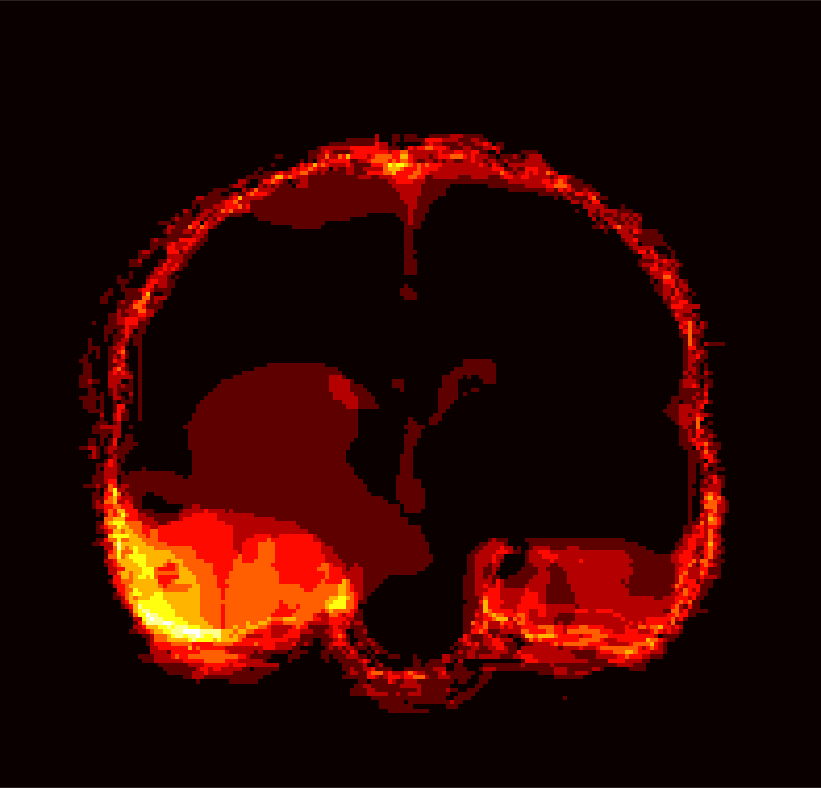} &
		\includegraphics[scale=0.08128]{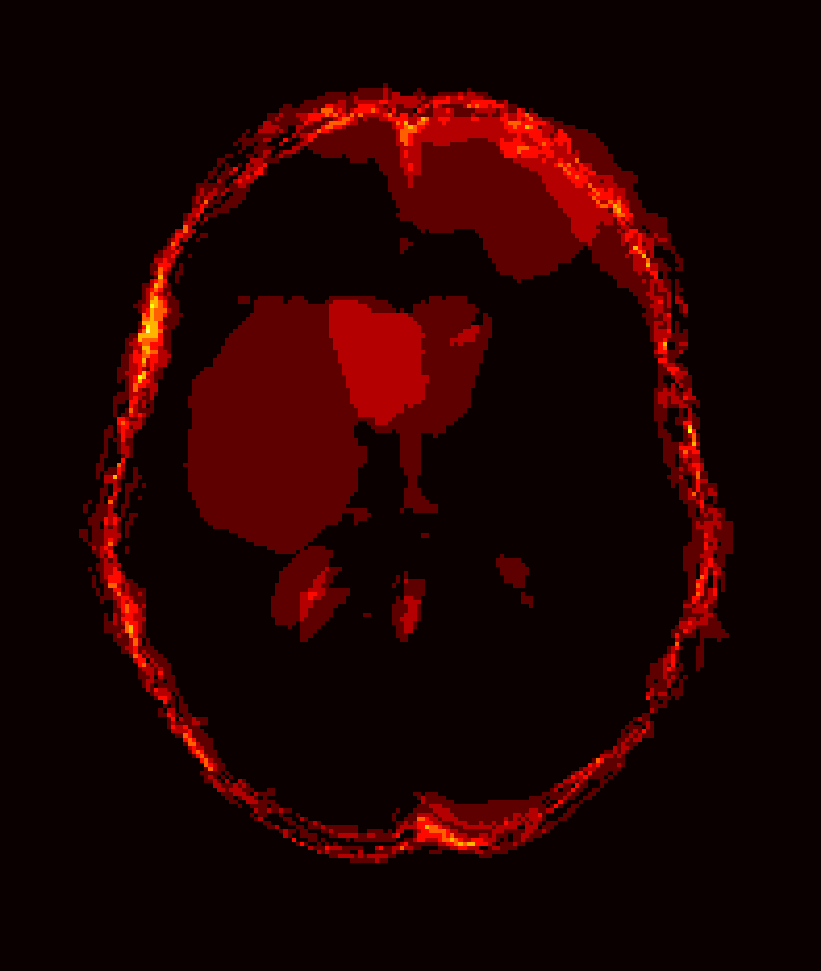}&
	\end{tabular}
	\captionof{figure}{Examples of 3D volumes of average errors for the normal IBSR and LPBA40 datasets, as well as for the pathological BRATS and TBI datasets. \xhsp{For IBSR/BRATS, we show results for BEaST*.} Images and their brain masks are first affinely aligned to the atlas. At each location we then calculate the proportion of segmentation errors among all the segmented cases of a dataset (both over- and under-segmentation errors). Lower values are better (a value of 0 indicates perfect results over all images) and higher values indicate poorer performance (a value of 1 indicates failure on all cases). Clearly, BSE and CNN struggle with the BRATS dataset whereas our PCA method shows good performance across all datasets.}
	\label{fig:heat_map}
\end{table*}

\subsection{Runtime and memory consumption}
\chg{Decomposition is implemented on the GPU. Each decomposition takes between 3 to 5 minutes. Currently, the registration steps are the most time-consuming parts of the overall algorithm. We use \texttt{NiftyReg} on the CPU for registrations. Each affine registration step takes less than 3 minutes and the B-spline step takes 5 minutes. However, in the current version of \texttt{NiftyReg} a B-spline registration can take up to 15 minutes when cost function masking is used. Overall our brain extraction approach takes around 1 hour to 1.5 hours for each case, including the pre-processing step. 
	
Storing the PCA basis requires the most memory. Each 197$\times$232$\times$189 3D image (stored as double) consumes about 66MB of memory. Hence it requires less than 7 GB to store the 100 PCA basis images, in addition to the atlases and masks. As our model only uses 50 PCA bases, stored in $B$, and requires two variable copies during runtime, our overall algorithm requires less than 7 GB of memory and hence can easily be run on modern GPUs.}

\section{Discussion}
\label{section:discussion}

We presented a PCA-based model specifically designed for brain extraction from pathological images. The model decomposes an image into three parts. Non-brain tissue outside of the brain is captured by a sparse term, normal brain tissue is reconstructed as a quasi-normal image close to a normal PCA space, and brain pathologies are captured by a total-variation term. The quasi-normal image allows for registration to an atlas space, which in turn allows registering the original image to atlas space and hence to perform brain extraction. Although our approach is designed for reliable brain extraction from strongly pathological images, it also performs well for brain extraction from normal images, or from images with subtle pathologies. 

\chg{This is in contrast to most of the existing methods, which assume normal images or only slight pathologies. These algorithms are either not designed for pathological data (BET, BSE, BEaST) or use normal data for training (e.g., ROBEX and CNN). Consequently, as we have demonstrated, these methods may work suboptimally or occasionally fail when presented with pathological data. While our PCA model is built on OASIS data which contains abnormal images (from patients with Alzheimer's disease), OASIS data does not exhibit strong pathologies as, for example, seen in the BRATS and the TBI datasets. However, as our algorithm is specifically modeling pathologies on top of a statistical model of {\it normal} tissue appearance, it can tolerate pathological data better and, in particular, does not require pathology-specific training.}

\chg{In fact, one of the main advantages of our method is that we can use a {\it fixed} set of parameters (without additional tuning or dataset-specific brain templates) across a wide variety of datasets.} \xhsp{This can, for example, be beneficial for small-scale studies, where obtaining dataset-specific templates may not be warranted, or for more clinically oriented studies, where image appearance may be less controlled.} We validated our brain extraction method using four different datasets (two of them with strong pathologies: brain tumors and traumatic brain injuries). On all four datasets our approach either performs best or is among the best methods. Hence, our approach can achieve good brain extraction results on a variety of different datasets.

\chg{There are a number of ways in which our method could be improved. For example, our decomposition approach is a compromise between model realism and model simplicity to allow for efficient computational solutions. However, it may be interesting to explore more realistic modeling assumptions to improve its quality. While the total variation term succeeds at capturing the vast majority of large tumor masses and would likely work well for capturing volumes of resected tissue, the texture of pathological regions will not be appropriately captured and will remain in the quasi-normal image. To obtain a more faithful quasi-normal image reconstruction would require more sophisticated modeling of the pathology. A possible option could be to train a form of auto-encoder (i.e., a non-linear generalization of PCA) to remove the pathology as in our prior work~\cite{yang2016registration}. A natural approach could also be to perform this in the setting of a general adversarial network~\cite{goodfellow2014generative} (GAN) to truly produce normal-looking quasi-normal images. As tumor images, for example, frequently exhibit mass effects, training and formulating such a model could be highly interesting as one could attempt to model the expected mass effect as part of the GAN architecture.}

The way we integrate our PCA model into the decomposition could also be improved. Specifically, for computational simplicity we only use the eigenspace created by a chosen number of PCA modes, but we do not use the strength of these eigenmodes. This is a simple, yet reasonable strategy, to form a low-dimensional subspace capturing normal tissue appearance as long as a pathology remains reasonably orthogonal to this subspace and hence would get assigned to the total variation part of the decomposition. 
	
We effectively constructed a form of robust PCA decomposition, which prefers outliers that jointly form regions of low total variation. Instead of modeling the decomposition in this way, it could be interesting to explore an LRS model which uses a partially-precomputed $L$ matrix and gets adapted for a given single image. Such a strategy may allow more efficient computations of the LRS decomposition, but would require keeping the entire training dataset in memory (instead of only a basis of reduced dimension). Such an approach could likely also be extended to a form of low-rank-total variation decomposition if desired. 

Regarding our PCA decomposition, it would be natural to use a reconstruction that makes use of a form of Mahalanobis distance~\cite{mahalanobis1936generalised}. This would then emphasize the eigendirections that explain most of the variance in the training data. Note, however, that our model is relatively insensitive to the number of chosen PCA modes. In fact, while different numbers of chosen PCA modes may affect how well the quasi-normal image is reconstructed, the number of PCA modes has only slight effects on the brain extraction results.

\chg{Tumors or general pathologies may also affect some of the pre-processing steps. For example, we perform histogram matching over the entire initial brain mask which includes the pathology. In practice, we visually assessed that such a histogram matching strategy produced reasonable intensity normalizations. However, this step could be improved, for example, by coupling it or alternating it with the decomposition in such a way that regions that likely correspond to pathologies are excluded from the histogram computations for histogram matching.}

\chg{While our model's simplicity allowed it to work well across a wide variety of datasets, this generality likely implies suboptimality. For example, a likely reason why the CNN approach performs poorly on some of the datasets is because these datasets do not correspond well to the data the CNN was trained on. Dataset-specific fine-tuning of the model would likely help improve the CNN performance. Similarly, approaches, including our own, relying on some form of registration and a model of what a well-extracted brain looks like would likely also benefit from a dataset-specific atlas (including a dataset-specific PCA basis in our case) or dataset-specific registration templates. Such dataset-specific templates can, for example, easily be used within MASS and improve performance slightly. In practice, large-scale studies may warrant the additional effort of obtaining dataset-specific manually segmented brain masks for training. However, in many cases such manual segmentation may be too labor-intensive. In this latter case our proposed approach is particularly attractive as it is only moderately affected by differing image appearances and works well with a generic model for brain extraction.}

\chg{Runtime of the algorithm is currently still in the order of an hour. It could be substantially reduced by using a faster registration method. For example, it may be possible to use one of the recently proposed deep learning approaches for fast registration~\cite{yang2017quicksilver,gutierrez2017guiding}. Furthermore, to speed-up the decompositions one could explore numerical algorithms with faster convergence or reformulations of the decomposition itself, as discussed above.}

\chg{Exploring formulations for different image sequences or modalities (or combination of modalities) would be interesting future work as well. It would also be interesting to explore if the generated quasi-normal image and the identified pathology could be used to help assess longitudinal image changes, for example for comparing the chronic and the acute phases of TBI.}

Our software is freely available as open source code at \url{https://github.com/uncbiag/pstrip}.

\section*{Acknowledgements}

Research reported in this publication was supported by the National Institutes of Health (NIH) and the National Science Foundation (NSF) under award numbers NIH R41 NS091792, NSF ECCS-1148870, and EECS-1711776. The content is solely the responsibility of the authors and does not necessarily represent the official views of the NIH or the NSF.

\appendix
\section{\texttt{NiftyReg} settings}
This section introduces the settings for \texttt{NiftyReg} used in this paper. We mainly use the affine registration \texttt{reg\_aladin} and the B-spline registration \texttt{reg\_f3d}.
\paragraph{Affine Registration:}For affine registration, we use \texttt{reg\_aladin} in \texttt{NiftyReg}. The options for affine registration are \texttt{-ref, -flo, -aff, -res}, which stand for reference image, floating image, affine transform output, warped result image, respectively. If the symmetric version is disabled, we add "\texttt{-noSym}". If center of gravity is used for the initial transformation, we add "\texttt{-cog}".
\paragraph{B-spline Registration:}For B-spline registration, we use \texttt{reg\_f3d} in \texttt{NiftyReg}. In addition to the options as shown in affine (except for \texttt{reg\_f3d} we use \texttt{-cpp} for output transform), we also use options \texttt{-sx 10, --lncc 40, -pad 0}, which include local normalized cross-correlation with standard deviation of the Gaussian kernel of 40, grid spacing of 10 mm along all axes, and padding 0.
\section{Methods settings}
This section introduces the settings that are used for all methods.
\paragraph{PCA} We use $\lambda=0.1$ for the sparse penalty and $\gamma=0.5$ for the total variation penalty.
\paragraph{ROBEX/CNN} ROBEX and CNN do not require parameter tuning. Therefore, we use the default settings, and for ROBEX we add a seed value of \texttt{1} for all datasets.
\paragraph{BET} We use the parameter settings suggested in the literature \cite{iglesias2011robust}\cite{kleesiek2016deep} for the IBSR and LPBA40 datasets. For the BRATS and TBI datasets, we choose the option \texttt{"-B"} for BET, which corrects the bias field and ``cleans-up'' the neck.
\paragraph{BSE} We use the parameter settings suggested in the literature \cite{iglesias2011robust}\cite{kleesiek2016deep} for the IBSR and LPBA40 datasets. For the BRATS and TBI datasets, we use the default settings.
\chg{\paragraph{BEaST} \xhsp{We use the ICBM and ADNI BEaST libraries to run all our experiments.} We first normalize the images to the \texttt{icbm152\_model\_09c} template in the BEaST folder. Then, we run BEaST with options \texttt{"-fill"}, \texttt{"-median"} and with configuration file \texttt{"default.2mm.conf"}. \xhsp{The spatial normalization step does not work reliably on the original IBSR and BRATS data. Thus, for these two datasets, we first apply the same affine transform as in our PCA pre-processing and then perform BEaST on the affine aligned images.}
\paragraph{MASS} We use the default parameters for MASS and use the 15 anonymized templates, provided with the MASS software package.}
\bibliography{biblio}

\end{document}